\documentclass[journal]{IEEEtai}
\usepackage[colorlinks,urlcolor=blue,linkcolor=blue,citecolor=blue]{hyperref}
\usepackage{makecell}
\usepackage{amssymb}
\usepackage{amsmath}
\usepackage{times}
\usepackage{graphicx}
\usepackage{subfigure}
\usepackage{multirow}
\usepackage{multicol}
\usepackage{amsmath}
\usepackage{bm}
\usepackage{threeparttable}
\usepackage{dcolumn}
\usepackage{booktabs}
\usepackage{amsthm}
\usepackage{latexsym}
\usepackage{rotating}
\usepackage{textcomp}
\usepackage{colortbl}
\usepackage{enumerate}
\usepackage{cite}
\usepackage{lscape}
\usepackage{color}
\usepackage{eqparbox,array}
\usepackage{algorithm}
\usepackage[switch]{lineno}
\usepackage{algorithm}  
\usepackage{algorithmic} 
\usepackage{multirow} 
\usepackage{amsmath} 
\usepackage{xcolor}
\usepackage{float} 

\newcommand{\etal}{\mbox{\emph{et al.}}}

\begin{document}
	\title{CausalBN-Bench:~A Comprehensive Benchmark for Causal Learning Capability of LLMs} 

	\author{
	Yu~Zhou,~\IEEEmembership{Graduate Student Member,~IEEE} Xingyu~Wu,~\IEEEmembership{Member,~IEEE}  Jibin~Wu,~\IEEEmembership{Member,~IEEE} Liang~Feng,~\IEEEmembership{Senior Member,~IEEE}  Kay~Chen~Tan,~\IEEEmembership{Fellow,~IEEE}
	\thanks{This work was supported in part by the Research Grants Council of the Hong Kong SAR (Grant Nos. C5052-23G, PolyU15215623, PolyU15229824, PolyU25216423, PolyU15217424, and SRFS2526-5S04), the National Natural Science Foundation of China (Grant No. 62306259), and The Hong Kong Polytechnic University (Project IDs P0056503, P0058445, P0060651, P0061084, and P0051130). \textit{(Corresponding author:~Xingyu Wu)}}.
	\thanks{
		Yu~Zhou, Xingyu~Wu, Jibin~Wu and Kay~Chen~Tan are with the Department of Data Science and Artificial Intelligence, The Hong Kong Polytechnic University, Hong Kong SAR 999077, China (e-mail: zy-yu.zhou@connect.polyu.hk, \{xingy.wu, jibin.wu, kctan\}@polyu.edu.hk)
	}
	\thanks{
		Liang Feng is with the College of Computer Science, Chongqing University, Chongqing 400044, China (e-mail: liangf@cqu.edu.cn).
	} 
}

\markboth{Journal of IEEE Transactions on Artificial Intelligence, Vol. XX, No. XX, Month 202X}
{Zhou~\MakeLowercase{\textit{et al.}}: CausalBN-Bench:~A Comprehensive Benchmark for Causal Learning Capability of LLMs}
\maketitle

\begin{abstract} 

The ability to understand causality significantly impacts the competence of large language models (LLMs) in output explanation and counterfactual reasoning, as causality reveals the underlying data distribution. However, the lack of a comprehensive benchmark currently limits the evaluation of LLMs' causal learning capabilities. To fill this gap, this paper develops CausalBN-Bench based on data from the causal research community, enabling comparative evaluations of LLMs against traditional causal learning algorithms. To provide a comprehensive investigation, we offer three tasks of varying difficulties, including correlation, causal skeleton, and causality identification. Evaluations of 19 leading LLMs reveal that, while closed-source LLMs show potential for simple causal relationships, they significantly lag behind traditional algorithms on larger-scale networks ($>50$ nodes). Specifically, LLMs struggle with collider structures but excel at chain structures, especially at long-chain causality analogous to Chain-of-Thought techniques. This supports the current prompt approaches while suggesting directions to enhance LLMs' causal reasoning capability. Furthermore, CausalBN-Bench incorporates background knowledge and training data into prompts to thoroughly unlock LLMs' text-comprehension ability during evaluation, whose findings indicate that, LLM understand causality through semantic associations with distinct entities, rather than directly from contextual information or numerical distributions.

\end{abstract}

\begin{IEEEImpStatement}

Existing benchmarks primarily assess the causal learning capability of LLMs via isolated variable question-answer pairs, typically ignoring the influence of additional variables. However, real-world causal mechanisms are inherently multivariate and often form complex networks. To address this gap, we propose CausalBN-Bench, a reproducible benchmark built from real-world datasets in the causal learning community with vetted ground-truth graphs that integrate training data and background knowledge.
CausalBN-Bench defines three tasks, correlation estimation, causal skeleton recovery, and causality identification, to holistically assess LLMs’ causal learning capabilities. It also provides four prompt formats, including variable names alone, variable names with background knowledge, variable names with training data, and variable names with both, thereby probing models’ ability to exploit prior information and to comprehend long text in causal reasoning. Overall, CausalBN-Bench makes the attainable performance and failure modes of LLMs visible across scales and complexities.
\end{IEEEImpStatement}


\begin{table}[tbp]
\centering

\caption{
	Previous causality identification evaluations for LLMs.}
\scalebox{0.9}{
	
	\begin{tabular}{cccc}
		\toprule
		\textbf{
			Previous
		} & \multirow{1}[3]{*}{\textbf{No. of Nodes}} & \multirow{1}[3]{*}{\textbf{Prompt format}} & \textbf{No. of Evaluated} \\
		\textbf{Studies} &  &  & \textbf{LLMs} \\
		\midrule
		\multirow{2}{*}{\cite{cai2023knowledge}} & \multirow{2}{*}{4$\sim$12} & Variable name        & \multirow{2}{*}{5} \\
		&                             & Observed data &                     \\
		\midrule
		\cite{zhiheng2022can} & 2$\sim$10 & Variable name & 1 \\
		\midrule
		\cite{long2023causal} & 8$\sim$37  & Variable name & 2 \\
		\midrule
		\cite{long2023can} & 3$\sim$4   & Variable name & 1 \\
		\midrule
		\cite{pawlowski2023answering} & 5$\sim$40  & Variable name & 2 \\
		\midrule
		\cite{tang2023towards} & NA & Variable name & 2 \\
		\midrule
		\cite{zhang2023understanding} & 2$\sim$20 & Variable name & 2 \\
		\midrule
		\cite{zevcevic2023causal} & 2$\sim$10  & Variable name & 4 \\
		\midrule
		\multirow{1}[3]{*}{\cite{jin2023can}} & \multirow{1}[3]{*}{2$\sim$6}   & \multirow{1}[3]{*}{Variable name} &3 series  \\
		&   &  & (9 LLMs) \\
		\midrule
		\cite{tu2023causal}  & 2$\sim$10  & Variable name & 3 \\
		\midrule
		\cite{zhang2024causal} & 11    & Variable name &  1  \\
		\midrule
		\cite{jiralerspong2024efficient} & 8$\sim$20  & Variable name &  1 \\
		\midrule
		\cite{antonucci2023zero}  & 2$\sim$20 & Variable name & 1 \\
		\midrule
		\cite{gao2023chatgpt}  & NA & Variable name & 4 \\
		\midrule
		\cite{jin2023cladder}  & 3$\sim$4 & Variable name & 8 \\
		\midrule
		\multirow{2}{*}{\cite{11141551}} & \multirow{2}{*}{2$\sim$25} & Variable name        & \multirow{2}{*}{9} \\
		&                             & Background Knowledge &                     \\
		\midrule
		\multirow{2}{*}{\cite{liu2024llms}} & \multirow{2}{*}{NA} & Variable name        & \multirow{2}{*}{1} \\
		&                             & Background Knowledge &                     \\
		\midrule
		\cite{tu2024carl}  & 10$\sim$51 & Variable name & 5 \\
		\midrule
		\cite{wang2024causalbench}  & NA & Variable name & 10 \\
		\midrule
		\cite{chen2024causal}  & {9} & {Variable name} & {28} \\
		\midrule
		\multirow{2}{*}{\cite{babakov2025causalgraphbench}} & \multirow{2}{*}{{5$\sim$222}} & {Variable name}        & \multirow{2}{*}{{3}} \\
		&                             & {Background knowledge} &                     \\
		\midrule
		\multirow{2}{*}{\cite{sheth2025causalgraph2llm}} & \multirow{2}{*}{{20$\sim$37}} & {Variable name}        & \multirow{2}{*}{{6}} \\
		&                             & {Background knowledge} &                     \\
		\midrule		
		\multirow{2}{*}{\cite{li2025can}} & \multirow{2}{*}{{37}} & {Variable name}        & \multirow{2}{*}{1} \\
		&                             & {Background knowledge} &                     \\
		\midrule
		
		\multirow{2}{*}{\cite{chen2023mitigating}} & \multirow{2}{*}{{48}} & {Variable name}        & \multirow{2}{*}{{1}} \\
		&                             & {Background knowledge} &                     \\
		\midrule
		\multirow{1}[3]{*}{CausalBN-Bench} 		 &\multirow{1}[3]{*}{2$\sim$109} &\multirow{1}[3]{*}{Four types \textsuperscript{*}}   &6 series  \\ 
		& &    & (19 LLMs) \\
		\bottomrule				
	\end{tabular}\label{tab:ref_summary}
}	
\newline
\newline
{*} `Four types' refers to variable name, variable name +  training data, variable name + background knowledge, and variable name +  training data + background knowledge.
\end{table}	
\section{Introduction}
\label{sec:intro}
Recently, large language models (LLMs) have garnered explosive attention across academia and industry \cite{10638808,wu2024evolutionary}.
Compared to traditional specialized models, LLMs exhibit versatility and exceptional performance on a wide range of tasks \cite{ijcai2024p579,huang2024exploring,zhou2024hm3}. However, as LLMs are trained on extensive corpora with billions of parameters, they sometimes struggle to accurately determine which learned knowledge to prioritize or ignore in different situations \cite{chang2023survey}. 
Consequently, LLMs tend to suffer from domain shift, where performance declines on data that differs from the training set, and long-tail bias, where less frequent examples are often inadequately handled~\cite{liu2024large}. 
{Hence, in many cases, LLMs not only need to understand or generate text based on observable patterns in the data, such as word frequency or common sentence structures, but also need to recognize the inherent causality of these patterns, such as the intent behind word usage or the contextual relationships between phrases.}
These requirements demand that LLMs not only comprehend surface-level correlations, but also perceive the causal relationships underlying text and data. By understanding causality, the performance of LLMs can be improved across various metrics, such as fairness, robustness, and explainability~\cite{feder2022causal}. Moreover, by considering reasonable alternatives in unobserved scenarios, known as counterfactuals, LLMs can better approximate human-like identification and comprehension~\cite{cai2023knowledge}.

To identify causality, causal learning has been proposed to recover a causal graph from the data \cite{nadkarni2001bayesian}, typically using directed acyclic graphs (DAGs) to visualize causal relationships, with nodes representing the variables and edges indicating the cause-effect direction\footnote{Due to its basis in causal graphs, the performance of causal learning is often influenced by the scale of the data (i.e., the number of nodes) \cite{cheng2022evaluation} and the density of the causal network (i.e., network sparsity) \cite{Fang2024On}.} \cite{vowels2022d}.
{Recently, increasing research interest \cite{zhiheng2022can,long2023causal,long2023can,pawlowski2023answering,tang2023towards,zhang2023understanding,zevcevic2023causal,jin2023can,tu2023causal,zhang2024causal,jiralerspong2024efficient,antonucci2023zero,tu2024carl,wang2024causalbench,chen2024causal,babakov2025causalgraphbench,sheth2025causalgraph2llm,li2025can,chen2023mitigating} has been attached to investigating the capabilities of causal learning in LLMs, which are summarized in Table \ref{tab:ref_summary}. A comprehensive related works are provided in Appendix~I}.
As shown in Fig. \ref{fig:evaluation_process}, these studies generally construct prompts based on variable names and a description of the evaluation task, i.e., identification of pairwise causal relationships. The prompt is input to the evaluated LLMs to obtain a causality identification result, which is compared with the ground truth of the causal graph from the data, over which metrics such as F1 score are calculated to measure the causal learning capability of the LLMs.
These works have made valuable contributions to the preliminary exploration of LLMs' causal learning abilities. However, there are still some common shortcomings. 
Specifically,

\begin{figure}[tb] 
\center{\includegraphics[width=\linewidth]  {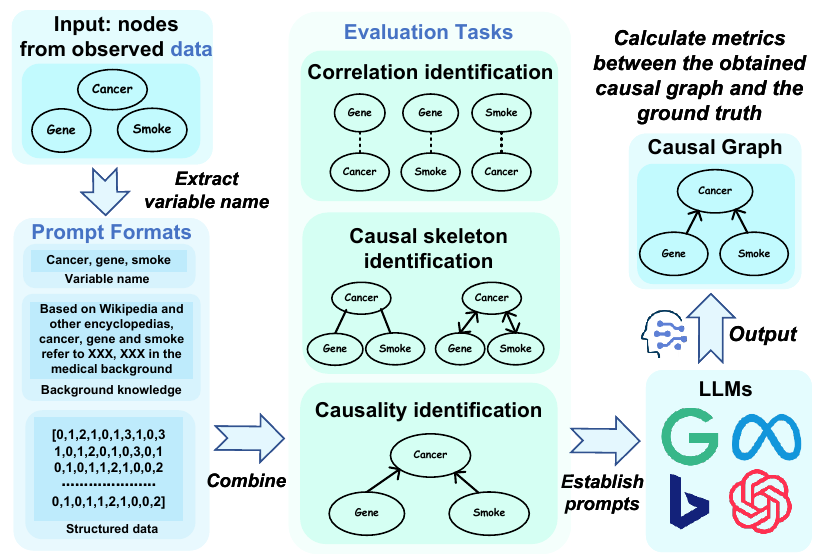}}\caption{Illustration of the overall evaluation process on CausalBN-Bench. The general evaluation framework is similar to existing evaluations, and the key differences between CausalBN-Bench and previous evaluations lie in each sub-process, where CausalBN-Bench possesses more standardized and comprehensive evaluation approaches. For instance, CausalBN-Bench features diverse prompt formats and evaluation tasks compared to the existing evaluations.
} 
\label{fig:evaluation_process} 
\end{figure}

\begin{itemize}
\item The causal networks used in these works either originate from private datasets \cite{pawlowski2023answering, tu2023causal} or only comprise a limited number of nodes, as shown in Col.~2 of Table \ref{tab:ref_summary}.

\item  The evaluation tasks lack diversity.
Most works~\cite{zhang2023understanding, zevcevic2023causal,jin2023can} focus on pairwise causality identification for evaluation while neglecting simpler correlation identification and more complex causal network identification tasks, which could better demonstrate LLMs' ability to grasp causal relationships at different scales and levels of difficulty.

\item  The prompt formats lack rich semantic information. Typically, only variable names are adopted for evaluation, as shown in Col.~3 of Table \ref{tab:ref_summary}, and the prior knowledge integration and long-text comprehension capabilities of LLMs \cite{tang2023towards,zhang2024causal,antonucci2023zero} are not fully exploited.

\item The diversity of the examined LLMs is limited. As shown in Col.~4 of Table \ref{tab:ref_summary}, only a small number of LLMs are evaluated, potentially undermining the generalizability of the assessments. 	
\end{itemize}

\begin{figure*}[htb] 
\center{\includegraphics[width=0.98\linewidth]  {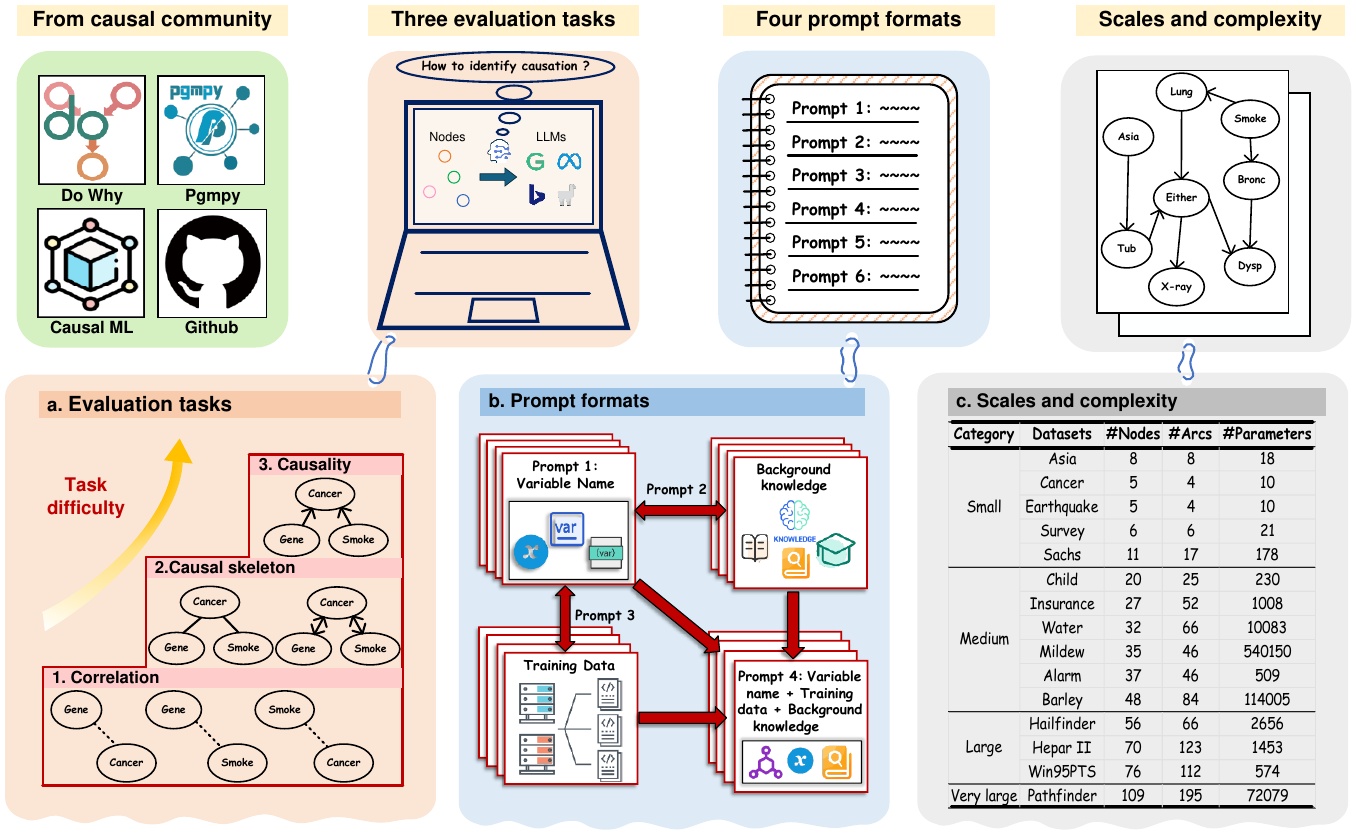}}
\caption{Illustration of CausalBN-Bench. CausalBN-Bench has four advantages, including diverse datasets from the causal learning community, three evaluation tasks of varying depths and difficulties, four prompt formats with rich information, and the demonstration of the upper limit of LLM capabilities across various scales and complexities. Specifically, in (a), CausalBN-Bench offers three tasks of different difficulties, i.e. correlation, causal skeleton identification, and causality, respectively, to holistically assess the causal learning capabilities of existing LLMs (e.g., gene, smoke and cancer). In (b), the variable name is the most prevalent prompt format in existing works; background knowledge is the domain knowledge for each variable in its field sourcing from Wikipedia and other encyclopedic websites, and training data refers to a matrix where columns denote nodes and rows denote observed samples (i.e., cases). In this paper, training data represents 500 observed samples for each variable. In CausalBN-Bench, four prompt formats are designed, including prompt 1 (i.e., variable name), prompt 2 (i.e., variable name + background knowledge), prompt 3 (i.e., variable name + training data), and prompt 4 (i.e., variable name + background knowledge + training data). In (c), CausalBN-Bench covers causal learning tasks of various scales, ranging from 5 to 109 nodes, evaluates various types of causal structures and discusses different densities in causal learning networks.} 
\label{fig:framework} 
\end{figure*}

To address these deficiencies, we create CausalBN-Bench, a comprehensive benchmark for evaluating the causal learning capacity of LLMs by curating datasets from Bnlearn\footnote{Bnlearn is a causal learning community bringing together widely used benchmarks in the field. Its website is https://www.bnlearn.com/.}
. From each dataset, we extract variable names and design four prompt formats, as shown in Col. 3 of Table \ref{tab:ref_summary}. We then define three core evaluation tasks: correlation, causal skeleton, and causality identification.
Compared with existing benchmarks that are ad hoc or pairwise, CausalBN-Bench is community-grounded and inherits vetted ground-truth graphs. This design ensures accurate relations and a difficulty level that mirrors observational discovery: it spans networks of varied scale and density, covers heterogeneous structures, and naturally captures the challenges of recovering skeletons and directions under realistic conditions. These properties are hard to emulate in pairwise settings and provide a common, graph-level testbed that bridges LLM-based causal learning and classical causal discovery across scales and structures.
In order to gain deeper insights into the overall causal learning capabilities of LLMs, we evaluate three closed-source LLMs, i.e. GPT3.5-Turbo,  GPT4 \cite{achiam2023gpt}, and GPT4-Turbo, along with five series of open-source LLMs: BERT series \cite{kenton2019bert}, LLAMA series \cite{touvron2023llama} (i.e., 7B, 13B and 33B), OPT series \cite{zhang2022opt} (i.e., 1.3B, 2.7B, 6.7B and 66B), Falcon series \cite{falcon40b} (i.e., 7B and 40B), and InternLM series \cite{team2023internlm} (i.e., 7B and 20B). As shown in Fig. \ref{fig:framework}, 
the resultant CausalBN-Bench possesses at least four advantages: 
\begin{itemize}
\item {\bf Diverse scales of datasets from the causal learning community:}
CausalBN-Bench is an extension of the causal learning research community’s efforts, designed to offer a robust evaluation framework. It incorporates 15 commonly used real-world causal learning datasets of diverse scales, enabling rigorous and quantitative measurement of LLMs' causal learning capacities with extensive evaluation results in the causal community as a reference.

\item {\bf Evaluation tasks of varying depths and difficulties: }
CausalBN-Bench offers three tasks of different difficulties, i.e. correlation, causal skeleton, and causality identification respectively, to holistically assess the causal learning capabilities of existing LLMs. 
Additionally, we have provided an example on causal structures similar to a particular prompt technique, which identifies the long cause-effect chain to evaluate causal learning abilities on multi-step reasoning like the popular Chain-of-Thought (CoT) prompt technique.

\item {\bf Diverse prompts with rich information: }
CausalBN-Bench offers four distinct prompt formats, encompassing variable name and its combinations with background knowledge and training data respectively and the combination of the three. With these diverse prompts, CausalBN-Bench can well assess LLMs' causal learning capacities through looking into their abilities to utilize prior information and comprehend long-text in understanding causal relations.

\item {\bf Demonstration of the upper limit of LLMs' causal learning capability across various scales and complexities:} CausalBN-Bench evaluates causal relations of varying scales and complexities. CausalBN-Bench covers causal learning datasets with scales ranging from 5 to 109 nodes, far exceeding what current evaluation works have explored. Meanwhile, it evaluates various types of causal structures and discusses different densities in causal learning networks.	
\end{itemize}

With extensive evaluations on CausalBN-Bench, our empirical study reveals that:

\begin{itemize}
\item Overall, the performance of LLMs in causal learning is still inferior to that of humankind.
Specifically, closed-source LLMs significantly outperform open-source ones but underperform compared with classic and state-of-the-art (SOTA) causal learning algorithms (Section \ref{sec:com_classical}). 
The best-performing LLMs approach the performance of traditional causal learning algorithms on small scale datasets but are inferior to traditional algorithms 
{at scales above 50 nodes, often achieving less than 50\% of their performance metrics.}
Additionally, in small scale networks, LLMs exhibit superior performance in identifying correlations (Section \ref{sec:correlation})  compared to causality (Section \ref{sec:causality}). However, in large scale networks, there are no statistically significant differences in performance across the three evaluation tasks (Sections \ref{sec:correlation}, \ref{sec:causal_skeleton} and  \ref{sec:causality}), indicating that the performance of LLMs in complex networks is usually not influenced by the difficulty of the evaluation task (the complexity of the underlying distribution to be solved). Instead, LLMs primarily rely on the contextual semantic information rather than recognizing the differences in the underlying distributions.

\item Regarding causal networks, LLMs exhibit varying performance across different depths and difficulties. As the scale of the causal learning dataset increases, both F1 score and accuracy of LLMs decrease (Section \ref{sec:causality}). Regarding identifying different structures of causal networks (Section \ref{sec:causality}), LLMs are more proficient at recognizing chain structures but struggle with collider structures, providing the direction for further enhancement of the LLMs' causal reasoning ability. {Notably, LLMs are capable of effectively completing CoT-analogous long-chain causal structures (Section \ref{sec:CoT}), which offers justification for the current mainstream prompt techniques.} Additionally, LLMs exhibit average in/out-degrees far exceeding those of traditional causal learning algorithms (Section \ref{sec:in_out_degree}), indicating that the DAGs obtained by LLMs have a denser structure, with most of the correct edges but a large number of erroneous edges.

\item Background knowledge influences LLM’s causality identification performance but is not always helpful for large scale datasets (Section \ref{sec:bk} and \ref{sec:bk+sd}).  
Background knowledge consistently boosts causal learning performance for clear and easy-understanding variable names, while the domain knowledge is unhelpful for unclear variable names, {suggesting that LLMs comprehend causal relationships through the semantic associations of clear entities rather than relying on contextual information or numerical distributions}. Meanwhile, training data also influences LLM’s causality identification performance (Section \ref{sec:sd} and \ref{sec:bk+sd}).
However, only the latest closed-source LLMs can do well due to their capacity to identify numerical data (Section \ref{sec:fine-gain}). Additionally, different {sentence structures}  and different meanings of variable names also affect the causality identification performance of   LLMs (Section \ref{sec:prompt}).
\end{itemize}

The remainder of this paper is organized as follows: {Section \ref{sec:problem} introduces the construction of the benchmark, highlighting three key components (i.e. data, evaluation tasks, and prompt formats), and illustrates the advantages of CausalBN-Bench compared to previous evaluations. Based on these three components, the evaluation results on CausalBN-Bench regarding different evaluation tasks and prompt formats are provided in Section \ref{sec:exp_task} and Section \ref{sec:exp_prompt}, respectively. After that, Section \ref{sec:exp_ana} presents the analysis of other influencing factors, such as data,  network and prompt, and provides the performance comparison between LLMs and traditional methods.} 
Section \ref{sec:discussion} concludes the paper and discusses the future trend.

{More details of this paper are provided in the Appendix. Appendix~I reviews the background and related works of CausalBN-Bench. Appendix II provides detailed experimental settings and additional evaluation definitions. Appendices III and IV present the prompt templates for different evaluation tasks and prompt formats. Appendix~V and Appendix~VI provide detailed evaluation results for different identification tasks and prompt formats. Appendix~VII reports additional results on other influencing factors, including sentence paraphrasing, variable refactorization, few-shot in-context learning, and parameter-efficient fine-tuning evaluation. Appendix~VIII contains comprehensive discussions, such as task design, traditional causal learning, practical implications, etc. Finally, Appendix~IX presents directions for future exploration, including evaluations of causal strength and token-generation probability.}

\section{Benchmark Construction}\label{sec:problem}
To construct CausalBN-Bench, we collect the data from the causal community and select all causal learning datasets with nodes below 100 based on preliminary evaluation results\footnote{CausalBN-Bench encompasses all datasets in Bnlearn with fewer than 100 nodes, since causal networks with up to 100 nodes are sufficient to evaluate the upper limits of current LLMs' capabilities. CausalBN-Bench will be maintained and updated over time. As LLMs continue to advance in the future, we will incorporate larger-scale causal networks into CausalBN-Bench to assess their evolving causal learning abilities.}. Then, following the steps of traditional causal learning methods to obtain causal graphs, we design three core evaluation tasks with varying depths and difficulties. Finally, we search for prior knowledge of causal learning from the encyclopedic websites such as Wikipedia, collect training data from the causal learning community, and design various prompt formats and templates to ensure fairness. In the following subsections, we will introduce the construction details of three key components in CausalBN-Bench. 
Moreover, we also provide a comparison of CausalBN-Bench with existing evaluation studies from the three aspects to highlight its advantages.

\subsection{Data Construction}

We include $15$ commonly-used real-world datasets from the causal learning community (i.e., Bnlearn) into CausalBN-Bench, as listed in Fig. \ref{fig:framework}(c).
As the upper limit of LLMs' ability to recognize causal structures is approximately 50 nodes, according to our empirical findings in Section \ref{sec:discussion}, 
we choose datasets of various sizes ranging from 2 to 109 nodes, and categorize them into small, medium, large, and very large scales, based on common practice in the community.
The detailed groupings are small scale (2-15 nodes): Asia, Cancer, Earthquake, Survey and Sachs; medium scale (15-50 nodes): Child, Insurance, Water, Mildew, Alarm and Barley; large scale (50-100 nodes): Hailfinder, Hepar II and Win95PTS; and one very large dataset ($>$100 nodes): Pathfinder.
In each dataset, we set three training sample sizes (i.e., 500, 1000 and 1500) to generate the sample data for CausalBN-Bench, which are used to verify the robustness and efficiency of LLMs.

\subsection{Evaluation Task Construction}\label{subsec:eva_task}	
We establish three evaluation tasks to assess the capabilities of LLMs for understanding causal relations of various depths and complexities, including correlation, causal skeleton, and causality identification tasks. {The detailed design rationale for these evaluation tasks is discussed in Appendix~VIII-A.}

\noindent \textbf{Correlation Identification.}
Correlation is a statistical measure expressing the extent to which two or more variables fluctuate together. A positive correlation indicates that as one variable increases, the other variable tends to increase, whereas a negative correlation suggests that as one variable increases, the other tends to decrease \cite{holland1986statistics}. 
Conversely, causality implies a cause-and-effect relationship between variables, where changes in one variable directly result in changes in the other~\cite{Yu2020Multi}.
Thus, causality implies correlation, but correlation does not necessarily imply causality \cite{pearl2009causality}. 
Correlation is a prerequisite for causation, and exploratory data analysis often begins with identifying correlations between variables, which can lead to hypotheses about causal mechanisms \cite{wu2019accurate}. 
In this work, the standard answers for correlations are derived from the causal diagrams of the datasets. 
According to the theory of d-separation \cite{pearl2009causality}, 
if the correlation between two variables can be recognized from the causal graph, then they are considered to be correlated. {In this paper, we consider \textbf{direct correlation} as a direct statistical dependency between a pair of variables and \textbf{indirect correlation} as a correlation recognized from the causal graph. 
Specifically,
\begin{itemize}
	\item \textbf{Direct correlation:} For a variable pair $\{X_i, X_j\}$, we set $y_{\mathrm{direct}}=1$ if and only if $X_i$ and $X_j$ are adjacent in the ground-truth DAG, meaning that either $X_i \rightarrow X_j$ or $X_j \rightarrow X_i$ holds. This is equivalent to adjacency in the DAG skeleton.		
	\item \textbf{Indirect correlation:} For a variable pair $\{X_i, X_j\}$, we set $y_{\mathrm{indirect}}=1$ if and only if $X_i$ and $X_j$ are d-connected in the ground-truth DAG under the empty conditioning set, and $X_i$ and $X_j$ are not adjacent in the DAG. This definition captures marginal dependence induced through paths via intermediate nodes.
\end{itemize}
Thus, the evaluation task about correlation involves the identification of direct and indirect correlations.}

\noindent \textbf{Causal Skeleton Identification.}
Causal skeleton refers to an undirected graph derived from a DAG by disregarding the directions of its edges. 
The skeleton preserves all variables and their pairwise connections, but omits edge directionality. 
The purpose of causal skeleton identification is to determine which variables are potentially causally related, while the directions of these relations are to be addressed in the subsequent evaluation task. 
Previous studies~\cite{Guo2023Adaptive,wu2023feature} %
often rely on statistical tests to assess conditional independence and dependence among variables, where the causal skeleton is constructed.
In CausalBN-Bench, models are required to identify the interconnections among variables and reconstruct the causal skeleton by evaluating conditional independence and dependence \cite{Wu2013Online,wu2020multi}.
Formally, 
consider three variables (or variable sets) $\mathbf{X}$, $\mathbf{Y}$, and $\mathbf{Z}$ in a causal dataset, where $\mathbf{Z}$ is a condition set. 
If the probability distribution $\mathbb{P}$ satisfies $\mathbb{P}(\mathbf{X}, \mathbf{Y}|\mathbf{Z})= \mathbb{P}(\mathbf{X}|\mathbf{Z})\mathbb{P}(\mathbf{Y}|\mathbf{Z})$, then the variables or variable sets $\mathbf{X}$ and $\mathbf{Y}$ are conditionally independent, denoted as $\mathbf{X} \bot \mathbf{Y}$. In contrast, $\mathbf{X} \not\perp \mathbf{Y}|\mathbf{Z}$ indicates that $\mathbf{X}$ is related to $\mathbf{Y}$ under $\mathbf{Z}$. The causal skeleton identification task aims to evaluate whether LLMs can accurately reason about conditional independence and dependence, thereby testing their capability to capture the underlying structure of causal graphs.

\noindent \textbf{Causality Identification.}
With the obtained causal skeletons, we also build an evaluation task to require LLMs to determine the causal directions between pairs of variables in the undirected graph. 
Causality identification task aims to assess the capacities of LLMs to identify causal relationships by leveraging the extensive prior knowledge. 
We provide two ways of causality identification. 
One is to identify the direction of the edges for the causal skeleton with LLMs. The other is to employ the d-separation method to derive six different descriptive characteristics of causal relationships, including two types of direct causal relationships, i.e. parental and child relationships, and four types of indirect causal relationships, i.e. ancestral, descendant, those with a collider structure, and those with a confounder structure; LLMs are then required to discover the causality based on these characteristics.
{Additionally, we further include an CoT-analogous causal structure identification task inspired by \cite{zevcevic2023causal, wei2022chain}}.

\noindent \textbf{CoT-analogous Causal Structure Identification.}
Unlike human cognition, LLMs significantly depend on elaborate prompt engineering to grasp task descriptions accurately. {Advanced prompting techniques have begun to play an important role in LLM reasoning, and related works are reviewed in Appendix I-C.
Among existing studies, the CoT~\cite{wei2022chain} prompting is one of the most influential approaches, which refers to generating a series of intermediate reasoning steps to solve a reasoning problem.} In this paper, we design a CoT-analogous causal structure identification task, which involves breaking down the causality identification task into a coherent series of intermediate reasoning steps that lead to the final causal relationship. 
For example, given four variables $\mathbf{A}$, $\mathbf{B}$, $\mathbf{C}$, and $\mathbf{D}$, a causality identification task is to determine whether $\mathbf{A}$ causes $\mathbf{D}$ with three reasoning steps: $\mathbf{A}$ causes $\mathbf{B}$, $\mathbf{B}$ causes $\mathbf{C}$, and $\mathbf{C}$ causes $\mathbf{D}$. Therefore, the CoT-analogous task tests LLMs' causality identification by providing a coherent series of intermediate reasoning steps, reflecting LLMs' causal learning capabilities.

{Notably, in the correlation and causal skeleton identification tasks, LLMs do not operate directly on raw data. Instead, they answer binary natural-language questions derived from the ground-truth causal graph under standard DAG semantics.}

\subsection{Prompt Format Construction}	
The prompts employed in current studies often only include variable names, contain limited semantics, and fail to harness the full potential of LLMs for long-text comprehension and prior knowledge integration.
In CausalBN-Bench, four prompt formats are incorporated: variable name, variable name + training data, variable name + background knowledge, and the combination of the three.
They delineate the upper bounds of LLMs' performance across different information scopes and illustrate variation in LLMs' efficiency when parsing textual versus data-centric information.

\noindent \textbf{Variable Name.}
As shown in Col.~3 of Table \ref{tab:ref_summary}, the variable name is the most prevalent prompt format in existing works, which is derived from real-world causal networks. 
In this work, we design six prompt templates for different evaluation tasks based on empirical experiences and integrate the advantages of prompts from previous evaluations~\cite{jin2023can}, which are provided in Appendix~III. 
Additionally, for each prompt template, we design five distinct prompts, presented in Appendix~III. The impact of these prompts will be carefully tested in Section \ref{sec:prompt}.

Additionally, we introduce an alternative evaluation mode named modified variable names to enhance the understanding of LLMs for the real-world implications of variable names. 
For example, in Asia dataset, there is a variable name \textit{Asia} which literally refers to the Asia continent. LLMs might interpret this name from geographical and climatic aspects of Asia, leading to an inference that there is no correlation or causality between \textit{Asia} and other variables like \textit{Lung cancer}. 
However, \textit{Asia} actually signifies the event \textit{Visiting to Asia}. Given the prevalence of respiratory diseases in many Asian countries, the event \textit{Visiting to Asia} could likely lead to respiratory diseases, subsequently causing lung cancer. This suggests a potential correlation and causality between \textit{Asia} and \textit{Lung cancer}. 
Therefore, we consider both the original variable name, like \textit{Asia}, and the modified variable name, like \textit{Visiting to Asia}, to explore the causal understanding capabilities of LLMs.

\noindent \textbf{Variable Name + Background Knowledge.}
As aforementioned, previous research uses a prompt format solely based on variable names. 
LLMs search across various domains associated with these variable names and make corresponding interpretations within the context.
Therefore, background knowledge related to the variable name should be considered when designing prompts for LLMs.
This has also been primarily studied previously.
For example,~\cite{chen2023mitigating} posits that incorporating background knowledge, particularly the specific knowledge of the related domain, into prompts can enhance the LLMs' understanding of relevant names. Therefore, we combine the variable names and related background knowledge as a prompt format to test whether LLMs can enhance their causal understanding capabilities by incorporating additional knowledge, especially in zero-shot scenarios.
In this paper, we gather background knowledge for each variable name within each dataset from Wikipedia
and some other encyclopedic websites. This background knowledge is then integrated with the variable name and appended to each prompt. The detailed processes are provided in Section~\ref{sec:bk}.

\noindent \textbf{Variable Name + Training Data.}
Traditional causal learning algorithms utilize training data as input, typically represented as a matrix where columns denote nodes and rows denote observed samples (i.e., cases) \cite{JMLR:v21:19-232}. For instance, for Alarm dataset comprising 37 variables, if there are 500 observed samples, then the input training data for causal learning can be represented as a 37\(\times\)500 matrix. 
However, current LLM-based causal learning evaluation methods rely on textual contexts, unlike classic causal learning algorithms.
This discrepancy prevents direct comparison between LLM-based and traditional causal learning algorithms.
We therefore design a prompt format using high-dimensional matrices of training data to evaluate LLM performance. 
{In addition to enabling direct comparison between LLM-based and classical methods,} this prompt format also helps evaluate LLMs' long-text comprehension capabilities.
It is noteworthy that this is the first work in causal learning that directly uses training data in prompts to evaluate causal understanding capabilities for LLMs. 
In practice, we provide the variable name and the high-dimensional matrix as a single prompt to the evaluated LLMs.
The detailed processes are provided in Section \ref{sec:sd}.

\noindent \textbf{Variable Name + Training Data + Background Knowledge.}
To further explore the causal learning capabilities of LLMs in prior knowledge utilization and long-text comprehension, we design a prompt format that combines the variable name, background knowledge, and training data. Based on this format, {we can further test whether LLMs can achieve better performance} by incorporating additional textual and numerical information. In this paper, we use the variable name, background knowledge, and training data as prompts, and provide detailed experimental procedures and results in Section \ref{sec:bk+sd}.

\subsection{{CausalBN-Bench vs. Existing Evaluation Benchmarks}}
{In causal learning, several benchmarks have been proposed to evaluate LLM-based approaches \cite{kiciman2023causal, tu2024carl, wang2024causalbench,sheth2025causalgraph2llm,babakov2025causalgraphbench}, including LLM-based causal discovery, causal reasoning, and other causal tasks with LLMs (see Appendix A for a detailed review). Because causal discovery provides an important basis for downstream reasoning when the true causal graph is unknown, we focus on evaluating LLMs in causal structure learning \cite{pearl2009causality}. Existing LLM-based causal discovery benchmarks\cite{kiciman2023causal,sheth2025causalgraph2llm,babakov2025causalgraphbench} are limited by small or private datasets, narrow task coverage, limited prompt settings, and insufficient use of background knowledge and training data. They also evaluate only a small number of models, limiting the generalizability of the results and failing to reflect LLM performance across causal learning tasks adequately. However, real-world causal relationships are often multivariate and structurally complex.} 
To address this gap, our benchmark CausalBN-Bench is constructed from real-world datasets in the causal community with ground-truth structures, thereby enabling a systematic assessment of LLMs' capabilities under diverse and complex causal topologies. The datasets of CausalBN-Bench enable integration and comparison to traditional data-driven methods. 

\begin{figure*}[htb] 
\center{\includegraphics[width=0.8\linewidth]  {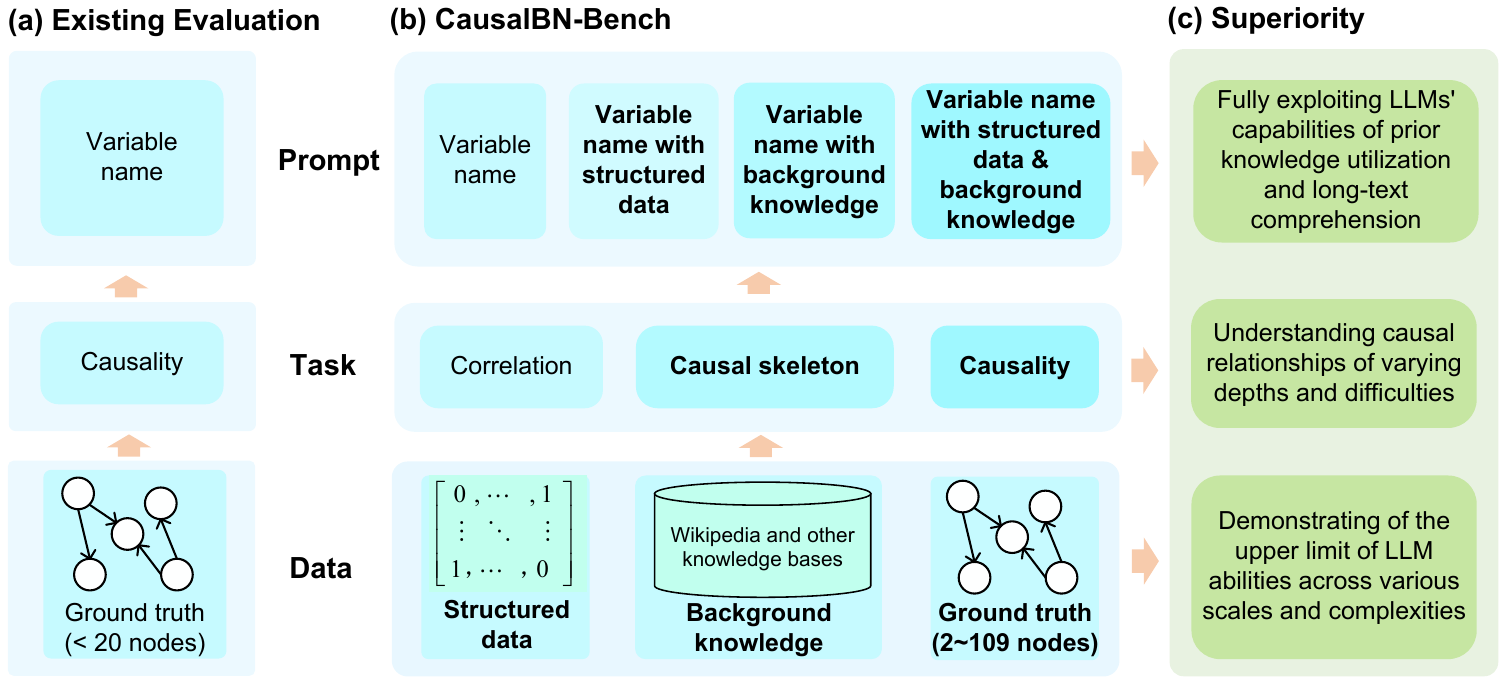}}\caption{{Construction process of CausalBN-Bench and comparison to existing evaluations.} (a) Existing evaluations are confined to causal learning datasets with no more than 51 nodes, concentrating on causality assessment, and utilizing variable names directly as the input format to LLMs. (b) CausalBN-Bench integrates training data, background knowledge, and a diverse set of ground truths, ranging from 2 to 109 nodes from the causal  community in its data component; it defines three principal evaluation tasks: identifying correlation, causal skeleton, and causality respectively, in the task component; it introduces three new prompt formats: variable names with training data, variable names with background knowledge, and a combination of both in the prompt component. (c) The advantages of CausalBN-Bench include better exploitation of prior knowledge utilization and long-text comprehension with LLMs, understanding causal relationships of varying depths and difficulties, and demonstrating the upper limits of LLM abilities across various scales and complexities.} 
\label{fig:construction of CausalBN-Bench} 
\end{figure*}

{Building on this richer dataset design, 
we juxtapose CausalBN-Bench}\footnote{The implementation of CausalBN-Bench is available at \url{https://github.com/Rainy-ZhouYu/CausalBN-Bench}} with previous evaluations in terms of data, evaluation tasks, and prompt formats in Fig.~\ref{fig:construction of CausalBN-Bench} to make its advantages clear. {Detailed discussions of CausalBN-Bench and existing causal benchmarks are provided in Appendix~I-D.
Compared with existing evaluations \cite{tu2024carl, wang2024causalbench,chen2024causal,babakov2025causalgraphbench,sheth2025causalgraph2llm,li2025can,chen2023mitigating} in Fig.~\ref{fig:construction of CausalBN-Bench}(a), CausalBN-Bench in Fig.~\ref{fig:construction of CausalBN-Bench}(b) integrates (i) training data, (ii) background knowledge, and (iii) a broad collection of ground truths from the causal learning community with 2–109 nodes.} This design enables a comprehensive assessment of the attainable performance of LLMs across dataset scales and causal-network complexities.
From the task perspective, CausalBN-Bench establishes three core evaluation tasks: correlation, causal skeleton, and causality identification, to thoroughly explore the capabilities of LLMs in understanding causal relations at different depths and difficulties. 
Regarding prompts, beyond conventional formats based solely on variable names like \cite{yang2024critical}, CausalBN-Bench introduces three additional formats: variable names with training data (as commonly assumed by traditional causal-learning algorithms), variable names with background knowledge, and variable names with both. In this way, CausalBN-Bench rigorously tests LLMs’ ability to incorporate prior information and to comprehend long textual inputs.
{In this paper, the term benchmark in CausalBN-Bench refers to a standardized evaluation suite that includes datasets with ground-truth causal networks, task definitions, prompt formats, and evaluation metrics to assess the causal learning capabilities of LLMs. To avoid ambiguity, we clarify that CausalBN-Bench supports four evaluation perspectives: cross-LLM comparison under the same prompts and datasets, cross-prompt comparison under the same LLM and datasets, comparison between LLM-based methods and traditional causal learning algorithms, and absolute evaluation against ground-truth networks using standard metrics. Based on CausalBN-Bench, researchers can develop more advanced LLM-based methods to narrow the gap between LLM-based and traditional causal methods.}

\section{Evaluation on Different Tasks}\label{sec:exp_task}
In this section, we first introduce evaluation metrics and evaluated LLMs. Then, we present the evaluation results and analysis on different tasks, including correlation, causal skeleton, causality, and CoT-analogous causal identification.

\subsection{Evaluation Metrics}
{
We employ the following metrics to assess the performance: 
\begin{itemize}
\item \textbf{F1 score}: represents the harmonic mean of precision and recall, with values ranging from 0 to 1. A higher F1 score indicates superior structural accuracy.
\item \textbf{Structural hamming distance (SHD)}:  measures the average structural Hamming distance, penalizing discrepancies between the ground-truth structure and the learned graph. A lower SHD value indicates a more accurate graph \cite{9712367}.
\item {\bf Structural intervention distance (SID)}:  measures the accuracy of causality model predictions. It calculates the prediction error of a causal graph relative to a reference graph, and a lower SID value indicates greater consistency between the model's predicted and true structures. 
\item \textbf{Network sparsity:} measures the sparsity of a predicted causal network relative to a fully connected directed network on the same variable set. A larger value indicates a sparser causal structure \cite{pearl2009causality}.
\item \textbf{In-degree and out-degree} denote the numbers of incoming and outgoing edges of a node, respectively, reflecting the numbers of its direct causes and direct effects in the predicted causal network \cite{pearl2009causality, peters2017elements}.
\item \textbf{Maximal inference length} denotes the longest shortest directed path among all reachable ordered pairs of variables in the predicted causal network. A larger value indicates longer directed causal chains.
\end{itemize}
The detailed definitions for network sparsity, in/out-degree, and maximal inference length are provided in Appendix~II-A.}
\subsection{Evaluated LLMs}
In order to evaluate causal learning ability gaps among different LLMs, we utilize five series of open-source LLMs: 
\begin{itemize}
\item BERT series: BERT-large \cite{lewis2020bart}, RoBERTa-large  \cite{liu2019roberta}, DeBERTa-large \cite{he2021deberta}, and DistilBERT-mnli \cite{shleifer2020pre};
\item LLAMA series \cite{touvron2023llama}: LLAMA-7B, LLAMA-13B and LLAMA-33B;
\item OPT series \cite{zhang2022opt}: OPT-1D3B, OPT-2D7B, OPT-6D7B and OPT-66B;
\item InternLM series \cite{team2023internlm}: InternLM-7B and InternLM-20B;
\item Falcon series \cite{falcon40b}: Falcon-7B and Falcon-40B;
\end{itemize}
along with one series of closed-source LLMs:
\begin{itemize}
\item GPT series \cite{achiam2023gpt}: GPT3.5-Turbo, GPT4 and GPT4-Turbo;
\end{itemize}
in zero-shot scenarios. 
{More details about these LLMs and the evaluation setting are provided in Appendices~II-B and~II-C.}
We evaluate these LLMs with different tasks and report the experiment results in the next subsection. 
It is worth noting that due to the poor performance on Pathfinder dataset, we exclude the results on it.

\subsection{Evaluation on Correlation Identification} \label{sec:correlation}
We define two types of correlations, i.e., direct and indirect correlation, and test the performance of LLMs on identifying them, respectively, whose prompts are provided in Appendix~III-A. As stated in Section~\ref{subsec:eva_task}, direct correlation refers to a direct correlational factor between a pair of variables, while indirect correlation refers to the correlation recognized from the causal graph. {The detailed generation of ground-truth labels is provided in Appendix~II-D.}

\noindent \textbf{Results on Direct Correlation Identification.}
We summarize the results of LLMs on the direct correlation identification task in Table~\ref{tab:correlation}. Overall, all models achieve moderate success on small-scale datasets, but their performance degrades substantially as dataset complexity increases. On small datasets, LLMs achieve reasonable F1 scores, whereas on medium datasets, performance drops significantly, with average F1 values falling into the 0.15-0.25 range. On large-scale datasets, LLMs nearly fail, and a similar trend is observed for accuracy.
Closed-source LLMs consistently outperform open-source models due to their larger parameters and broader corpora. Among open-source models, the LLAMA series generally achieves the best results, followed by the InternLM series. Notably, some open-source models exhibit dataset-specific strengths. For instance, InternLM-20B performs better on Cancer and Earthquake, LLAMA-33B on Water, and Falcon-40B on Barley, approaching the performance of closed-source LLMs. This suggests that pre-training on domain-relevant corpora can provide advantages on specialized datasets. This is also illustrated in Fig. 16 in Appendix~V-A.
\begin{table}[htbp]
\centering
\caption{Performance of evaluated LLMs on correlation identification}
\label{tab:correlation}
\scalebox{1.0}{
\begin{tabular}{lcccc}
	\toprule
	\multirow{1.5}[4]{*}{Causal networks} & \multicolumn{2}{c}{Direct Correlation$^{\dagger}$} & \multicolumn{2}{c}{Indirect Correlation$^{\dagger}$} \\
	\cmidrule{2-5}          & F1 score* & Accuracy* & F1 score* & Accuracy* \\
	\midrule
	Asia  & 0.4126 & 0.4059 & 0.3923 & 0.4008 \\
	Cancer & 0.5279 & 0.5333 & 0.5150 & 0.5161 \\
	Earthquake & 0.5673 & 0.5775 & 0.5574 & 0.5730 \\
	Survey & 0.3479 & 0.3675 & 0.3309 & 0.3568 \\
	Sachs & 0.5181 & 0.5242 & 0.5075 & 0.5150 \\
	Child & 0.1687 & 0.1879 & 0.1545 & 0.1919 \\
	Insurance & 0.2601 & 0.3705 & 0.2647 & 0.3702 \\
	Water & 0.2492 & 0.2551 & 0.2379 & 0.2459 \\
	Mildew & 0.2379 & 0.2466 & 0.2236 & 0.2443 \\
	Alarm & 0.1791 & 0.2425 & 0.1777 & 0.2257 \\
	Barley & 0.2262 & 0.2644 & 0.2221 & 0.2531 \\
	Hailfinder & 0.1257 & 0.1243 & 0.1158 & 0.1183 \\
	Hepar II & 0.1592 & 0.1669 & 0.1485 & 0.1610 \\
	Win95PTS & 0.1541 & 0.1416 & 0.1369 & 0.1368 \\
	\bottomrule
	\end{tabular}}
	\newline
	\newline
	
	\begin{minipage}{\linewidth}
\raggedright
{$^{\dagger}$ Labels of direct correlation $y_{\mathrm{direct}}$ and indirect correlation $y_{\mathrm{indirect}}$ are derived from the ground-truth DAGs using adjacency for $y_{\mathrm{direct}}$ and d-connection under the empty conditioning set for $y_{\mathrm{indirect}}$, with the additional constraint that $y_{\mathrm{indirect}}$ excludes adjacent pairs.}
\newline
{{*} Averaged F1 score and accuracy are computed over the LLM set listed in Table~VIII in Appendix~II-B. The detailed performance of each LLM is provided in Appendix~V-A.}
\end{minipage}
\end{table}

\noindent \textbf{Results on Indirect Correlation Identification.}
We report the results of LLMs on indirect correlation identification in Table~\ref{tab:correlation}. Overall, LLMs demonstrate limited capability, with reasonable performance only on small-scale and a few medium-scale datasets, and near failure on large-scale datasets. On small datasets, LLMs achieve moderate F1 scores, but as the number of nodes increases, their performance declines sharply. For most medium datasets, average F1 scores fall within 0.15 to 0.25, while on large datasets, both F1 scores and accuracies remain close to chance levels from 0.11 to 0.16. 
As illustrated in Fig.~17 in Appendix~V-B, closed-source LLMs consistently outperform open-source models across all dataset scales, particularly on large datasets where the performance gap becomes most pronounced. Among open-source models, the LLAMA series generally performs best on small and medium datasets, while the OPT series occasionally achieves better results on large datasets due to its larger parameter size. In contrast, the performance differences between InternLM and Falcon models remain marginal.
Overall, all LLMs exhibit stronger performance on direct correlation identification than on indirect correlation identification. This suggests that recognizing indirect correlations, which requires inferring multi-step dependencies, is considerably more challenging. Model performance correlates with parameter size. GPT-4-Turbo achieves the strongest results among closed-source models, while LLAMA-13B shows stable performance among open-source models, albeit with higher training cost requirements.

\subsection{Evaluation on Causal Skeleton Identification}	\label{sec:causal_skeleton}
We examine the performance of LLMs at identifying causal skeletons, with the prompts depicted in Appendix~III-B.
The results are presented in Fig. \ref{exp:ske} and the detailed results are provided in Appendix~V-C. 
From them, it can be seen that most LLMs struggle with
recognizing causal skeletons.
Some LLMs, such as InternLM series, exhibit commendable performance on this evaluation task. 
As shown in Fig. \ref{exp:ske_reda_Acc}, most LLMs get an accuracy of around 60\%, with InternLM-7B and Falcon-7B continuing to exhibit subpar performance, while InternLM-20B and LLAMA-7B maintain strong performance. 
This might be attributed to the fact that InternLM-7B and Falcon-7B do not acquire robust causal reasoning capabilities during pre-training. Conversely, InternLM-20B and LLAMA-7B are better at identifying causal skeletons.

In the causal skeleton identification task, CausalBN-Bench tests the causal learning abilities of LLMs from the causal inference aspect. In contrast, the other two tasks are more focused on LLMs utilizing prior knowledge. Therefore, LLMs that perform well in correlation identification may not necessarily do as well in causal skeleton identification. For instance, InternLM-20B shows average performance on correlation identification tasks but excels at causal skeleton identification tasks. On the other hand, GPT4-Turbo leads in identifying correlation tasks across almost all LLMs, but it shows only moderate performance on causal skeleton tasks.

\begin{figure}[ht]
\centering
\subfigure[F1 score]{\includegraphics[width=0.48\linewidth]{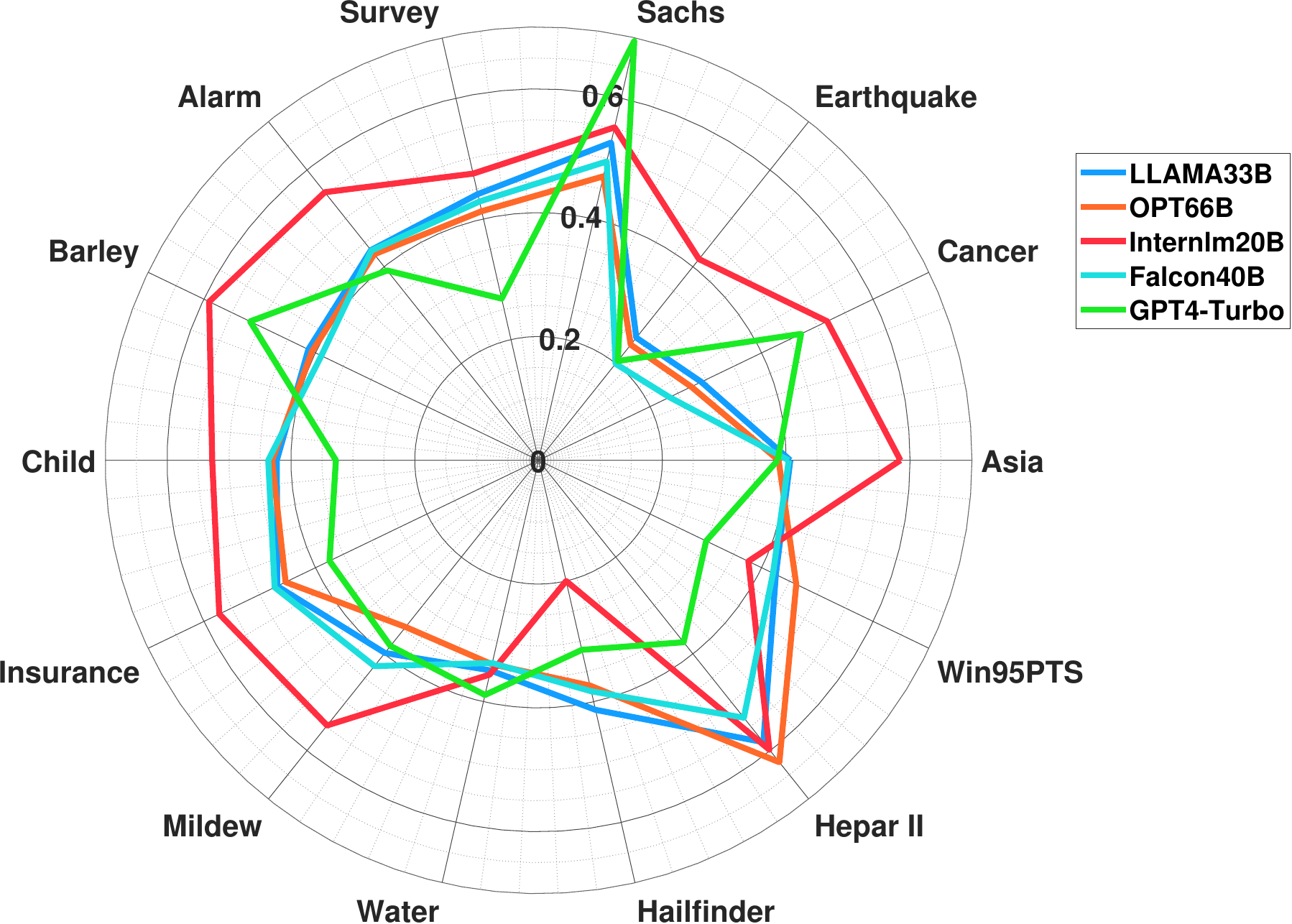}\label{exp:ske_reda_F1}}
\subfigure[Accuracy]{\includegraphics[width=0.48\linewidth]{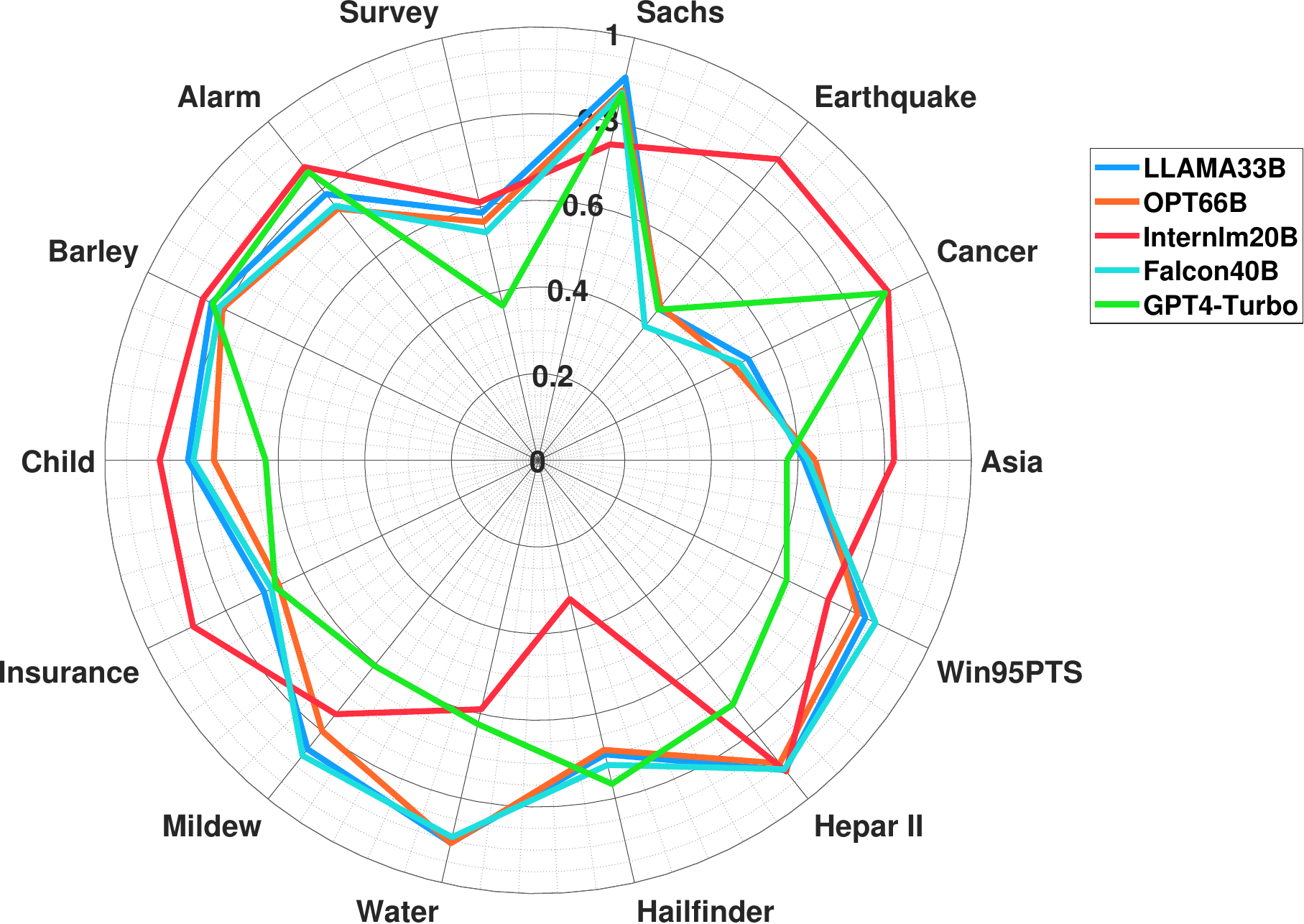}\label{exp:ske_reda_Acc}}\\
\caption{Performance of causal skeleton identification for different datasets.}
\label{exp:ske}
\end{figure}

\subsection{Evaluation on Causality Identification}\label{sec:causality} 
We design two evaluation methods, with prompts provided in Appendix III-C.
The first directly queries LLMs about the existence of causal relations between variable pairs using five prompt variants, analogous to the correlation identification task.
The second infers causal directions after the skeleton has been identified.
These methods provide a comprehensive assessment of LLMs’ capacity to reason about causality.
We present the results for the first method in Fig.~\ref{fig:cas1}, and the detailed results can be found in Appendix V-D.
From Fig. \ref{fig:cas1_F1}, 
open-source LLMs achieve limited performance, with most F1 scores ranging from 0.2 to 0.4, although LLAMA-7B occasionally exceeds 0.5 on small datasets. In contrast, closed-source LLMs achieve better results on small datasets, around 0.5 to 0.6 on medium datasets, and 0.5 on large datasets. Accuracy results in Fig.~\ref{fig:cas1_Acc} follow a similar pattern. Specifically, open-source models rarely exceed 60\%, whereas closed-source LLMs surpass 80\% on most datasets. Among all models, GPT-4-Turbo exhibits the strongest overall performance, while Falcon-40B leads among open-source LLMs.
\begin{figure}[t] 
\centering
\subfigure[F1 score]{\includegraphics[width=0.48\linewidth]{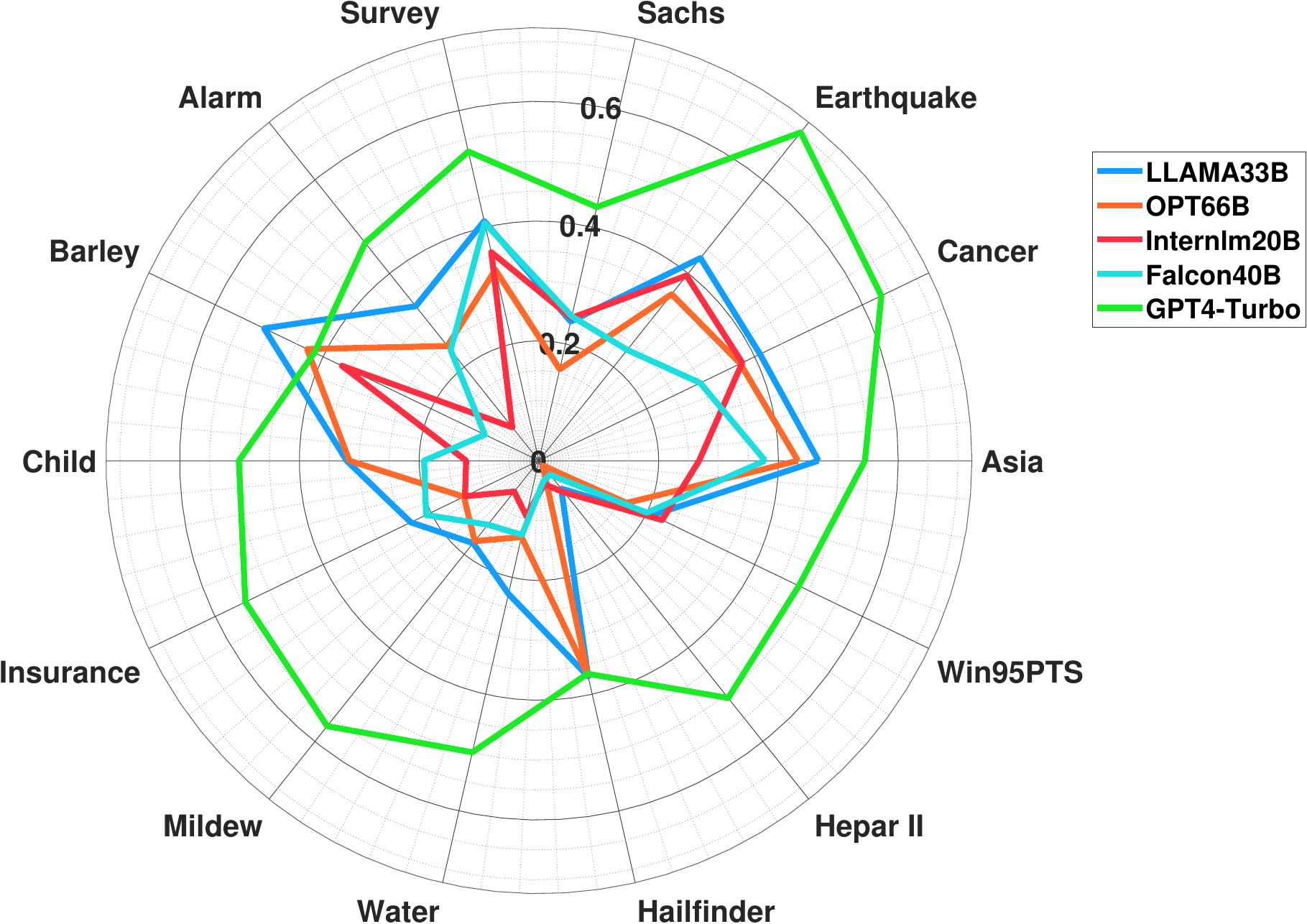}\label{fig:cas1_F1}}
\subfigure[Accuracy]{\includegraphics[width=0.48\linewidth]{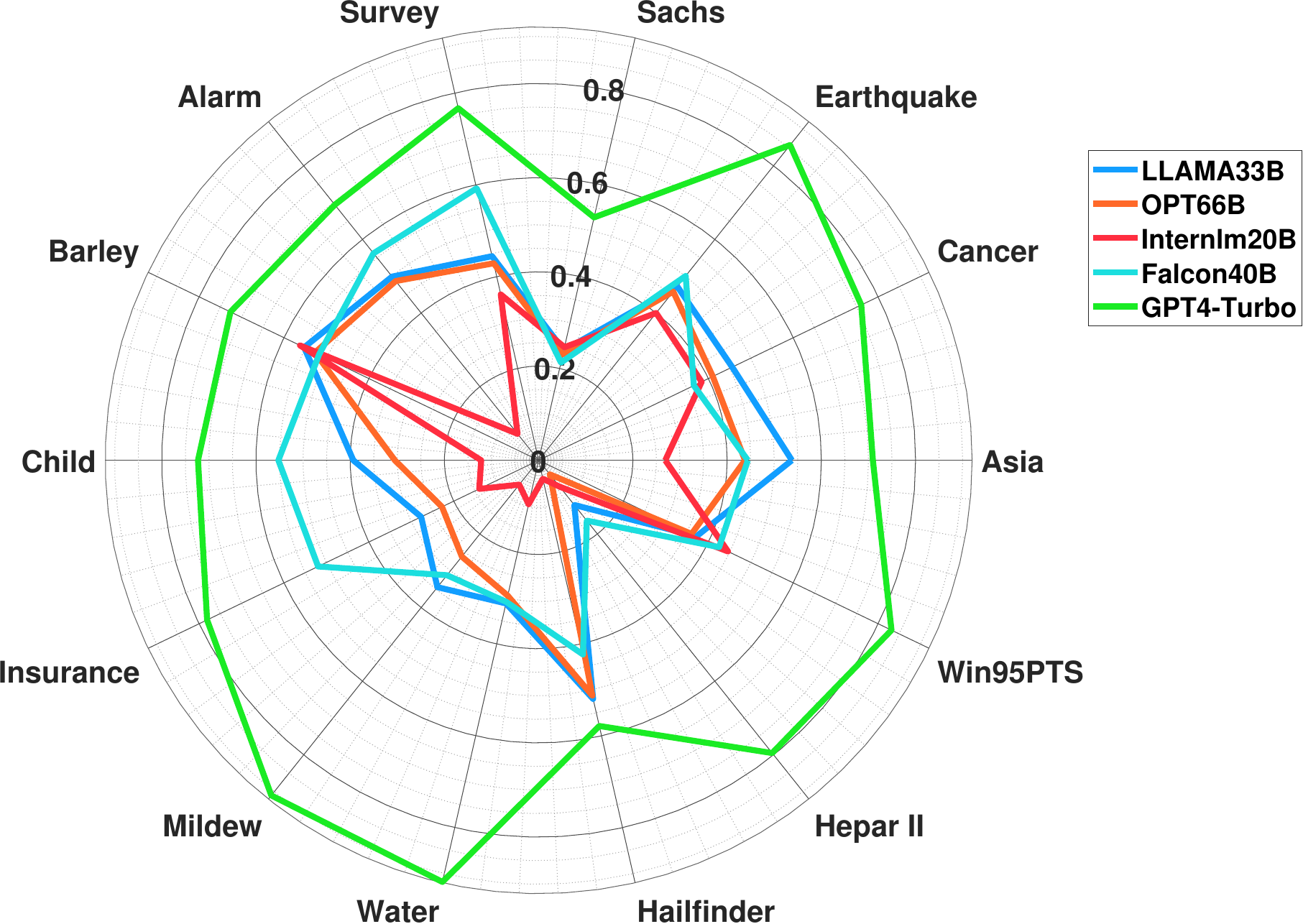}\label{fig:cas1_Acc}}\\
\caption{Performance of causality identification for the first method.}
\label{fig:cas1}
\end{figure}

For the second method, we provide the results in Table XIII of Appendix~V-E. Overall, LLMs fail to effectively identify causal relationships under this setting. Across all datasets, F1 scores rarely exceed 0.4 and accuracies remain below 50\%, with performance deteriorating sharply on medium and large datasets where F1 scores drop to 0.1–0.2 and accuracies to 20–30\%. For SHD and SID, LLMs perform worse than classical methods. Furthermore, the generated graphs tend to be dense, with sparsity values between 0.6 and 0.7 across most datasets, indicating that LLMs produce overly connected rather than sparse causal graphs.
Next, we examine the causality identification capabilities of LLMs based on Fig.~\ref{fig:cas2} and Fig.~20 of Appendix~V-E. Specifically, from Fig. \ref{fig:cas2_F1}, it can be seen that open-source LLMs demonstrate commendable causal abilities on small scale datasets, with LLAMA series LLMs equaling or surpassing closed-source LLMs. 
This could be attributed to the simplicity of tasks on small scale datasets, which open-source LLMs are adept at recognizing and understanding. However, on medium and large scale datasets, closed-source LLMs maintain impressive performance, achieving the best results. 
Especially, LLAMA series also sustains robust performance on medium to large scale datasets, with their F1 scores on Win95PTS and Hepar II even exceeding closed-source LLMs. Other open-source LLMs demonstrate a specific capability for identifying causality on small scale datasets. However, their performance on medium to large scale datasets is less satisfactory, particularly for LLMs with smaller parameter scales. For closed-source LLMs, GPT4-Turbo exhibits the best performance on large scale datasets, GPT3.5-Turbo obtains the highest F1 score on medium scale datasets, and GPT4 has a certain edge on small scale datasets. The same conclusions of accuracy are shown in Fig. \ref{fig:cas2_Acc}. For a comprehensive overview of these results, please refer to Appendix~V-E.
\begin{figure}[t]
\centering
\subfigure[F1 score]{\includegraphics[width=0.48\linewidth]{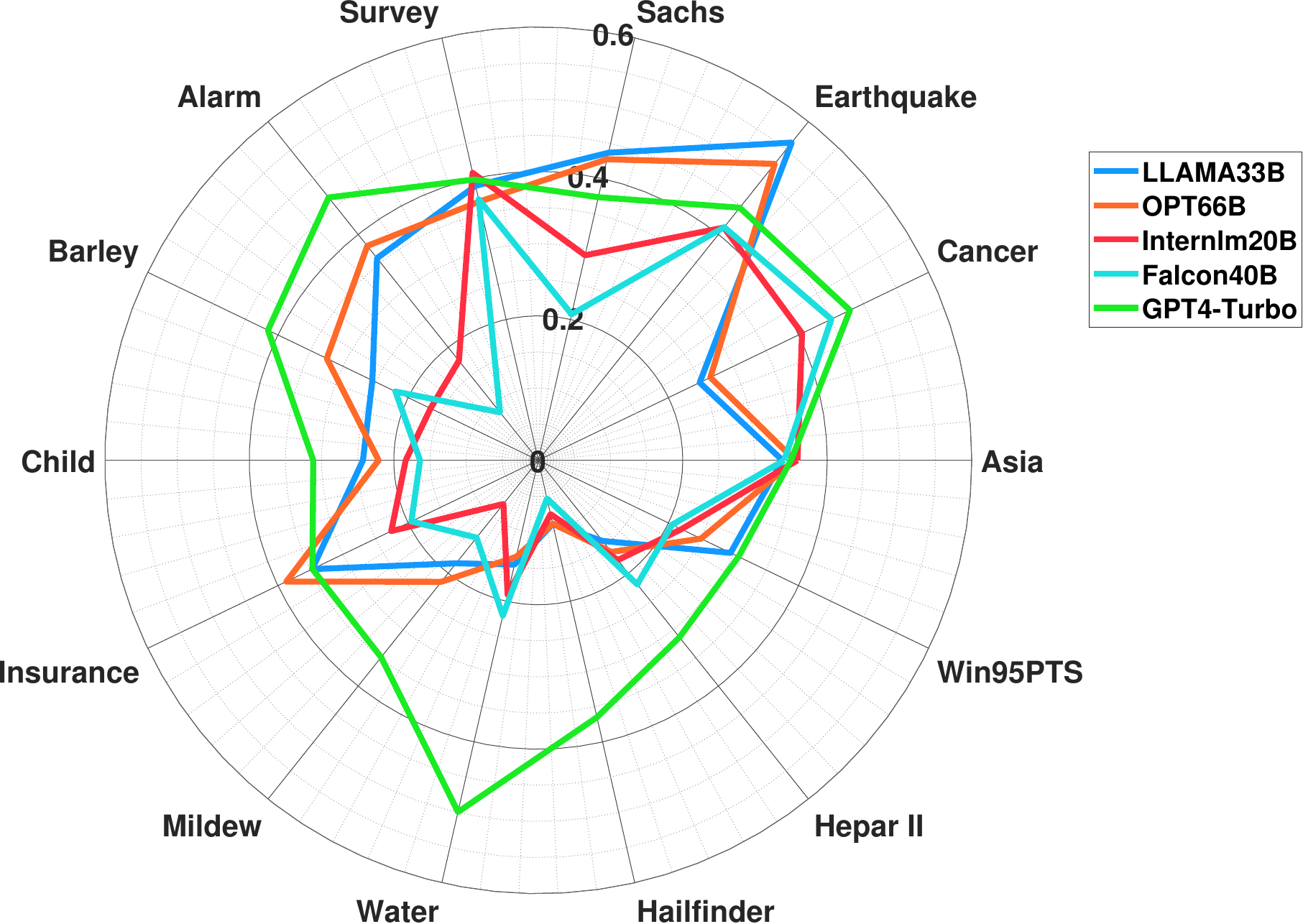}\label{fig:cas2_F1}}
\subfigure[Accuracy]{\includegraphics[width=0.48\linewidth]{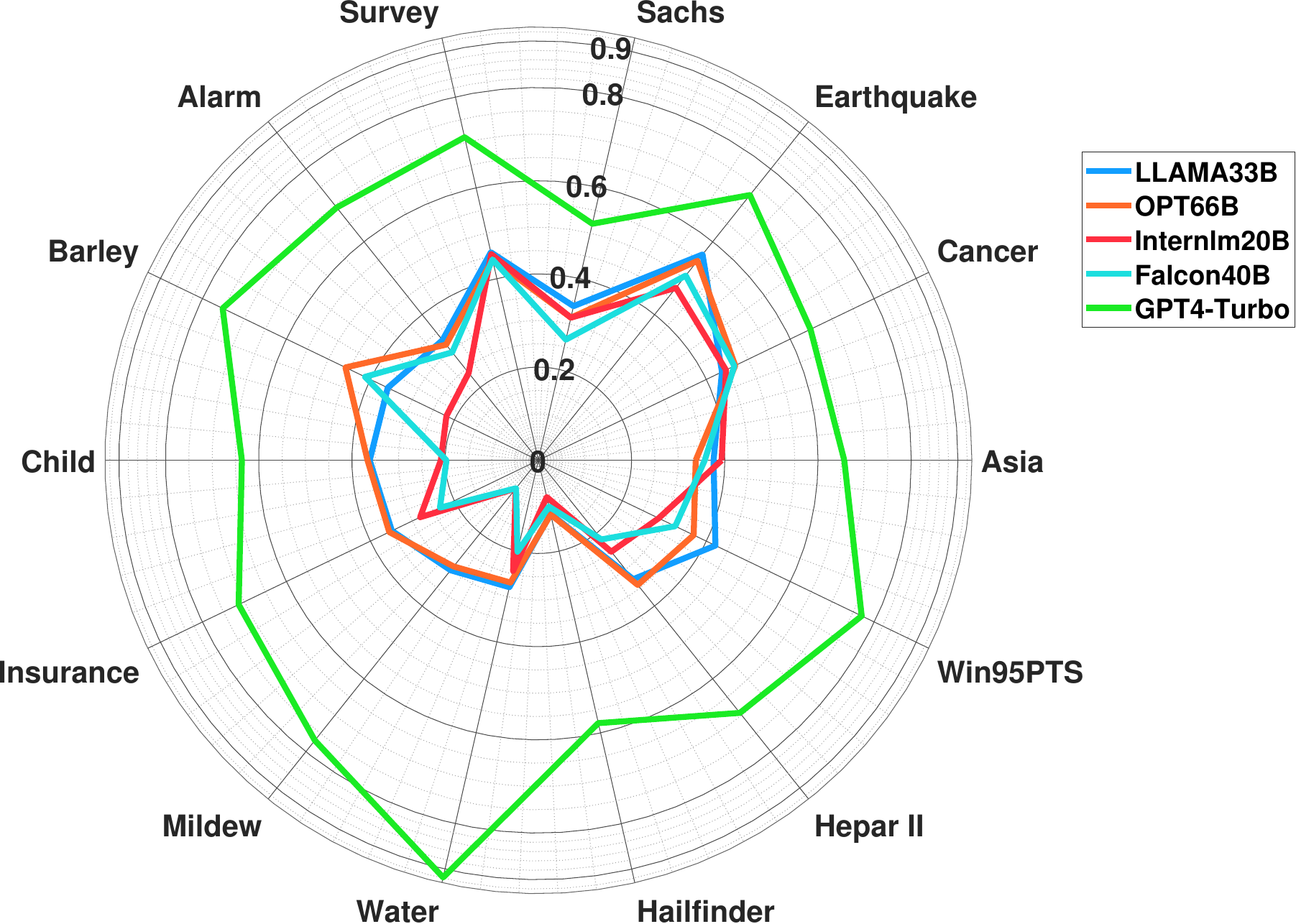}\label{fig:cas2_Acc}}\\
\caption{Performance on causality identification for the second method.}
\label{fig:cas2}
\end{figure}

We visualize the performance of SHD and SID for the second method in Figs.~21 and~22 of Appendix~V-E. From them, closed-source LLMs consistently outperform open-source models and yield lower SHD and SID across most datasets. Among open-source LLMs, LLAMA series achieves the best performance on small and medium datasets, while Falcon series performs relatively well on large datasets. Network sparsity in Fig.~23 in Appendix~V-E shows that LLMs with lower SHD and SID typically generate sparser graphs. Closed-source LLMs maintain sparsity levels below 0.5 and reflect more parsimonious structures, whereas Falcon series remain denser at around 0.4. Other open-source LLMs produce graphs with even higher density.

Additionally, the second method utilizes the d-separation technique to derive six distinct descriptive features of causal relationships, which include two types of direct causality (i.e., parent and child) and four types of indirect causality (i.e., ancestral, descendant of a chain structure, a collider structure, and a confounder structure). 
The experimental results are provided in Table \ref{tab:structure}. It can be seen that the performance differences in identifying chain, collider, and confounder structures across LLMs are not significant; existing LLMs are better at recognizing chain structures and less adept at identifying collider structures.

\begin{table}[tbp]
\centering
\caption{Performance of different structures in causality identification}	
\label{tab:structure}
\scalebox{0.8}{
\begin{tabular}{lcccccc}
	\toprule
	\multirow{1.5}[4]{*}{Causal Structure} & \multicolumn{2}{c}{Open-source LLM} & \multicolumn{2}{c}{Closed-source LLM} & \multicolumn{2}{c}{All LLMs} \\
	\cmidrule{2-7}          & F1 score & Accuracy & F1 score & Accuracy & F1 score & Accuracy \\
	\midrule
	\multirow{1.5}[2]{*}{Direct} & 0.2769  & 0.3542  & 0.4608  & 0.4760  & 0.3174  & 0.3810  \\
	& 0.2242  & 0.2700  & 0.4439  & 0.4654  & 0.2725  & 0.3130  \\
	\midrule
	\multirow{1.5}[2]{*}{Chain} & 0.2640  & 0.3441  & 0.5185  & 0.5729  & 0.3200  & 0.3944  \\
	& 0.1879  & 0.2071  & 0.4607  & 0.4758  & 0.2479  & 0.2662  \\
	\cmidrule{1-1}    Average of Chain & 0.2260  & 0.2756  & 0.4896  & 0.5244  & 0.2840  & 0.3303  \\
	\midrule
	{Confounder} & 0.1628  & 0.1636  & 0.4410  & 0.4442  & 0.2240  & 0.2253  \\
	Collider  & 0.1632  & 0.1699  & 0.3662  & 0.3528  & 0.2079  & 0.2101  \\
	\bottomrule
	\end{tabular}}
\end{table}%

\subsection{Evaluation on CoT-analogous Causal Structure Identification} \label{sec:CoT}
In addition to the above tasks, we also evaluate LLMs with CoT-analogous causal structure identification task. 
First, we design CoT prompting, presenting the LLM with a sequence of question-and-answer pairs that mimic the questioning style used to prompt the model. These question prompts, along with comprehensive CoT findings, are documented in Appendix~III-D. The evaluation results are provided in Table~XVI of Appendix~V-F and demonstrate notable performance enhancements across all models on both datasets when using CoT prompts. Meanwhile, we find that nearly all LLMs with parameter sizes greater than 6 billion can effectively complete CoT-analogous causal structure identification experiments.
This indicates that LLMs exhibit strong reasoning capabilities in causal learning, enabling them to identify causal relationships. The results also indicate that CausalBN-Bench supports popular prompt techniques and provides guidance for further improving LLMs' causal learning abilities.

\section{Evaluation on Different Prompt Formats} \label{sec:exp_prompt}
In this section, we perform evaluations regarding different prompt formats in CausalBN-Bench to test the capabilities of LLMs %
to utilize prior knowledge and comprehend long texts for causal learning tasks.
Due to the excessively large parameter scale of LLMs and the overly complex tasks, which result in unacceptably long inference time for LLMs, we exclude the results of LLMs with parameter scales exceeding 30B, and focus on the causality identification task.

\subsection{Evaluation on Using Variable Name + Background Knowledge as Prompt}
\label{sec:bk} 
We first identify the background domain of each dataset (e.g., Cancer in the medical domain, Win95PTS in computer science) and collect background knowledge for each variable from Wikipedia and other encyclopedic sources. This information is appended to variable names to form enriched prompts, with details provided in Appendix~IV-B.
The evaluation results are summarized in Fig. \ref{fig:kno_cau}, with complete results in Fig. 24 of Appendix VI-A. 
Overall, LLMs still struggle to identify causality under this prompt format. Closed-source models consistently achieve higher F1 and accuracy, while yielding lower SHD and SID in Figs. 25 and 26 in Appendix~VI-A, compared to open-source models. 
Among open-source models, the LLAMA series performs best on small and medium datasets, while Falcon models show relative strengths on large datasets. 
Network sparsity analysis in Fig.~27 of Appendix~VI-A confirms that models with lower SHD and SID produce sparser graphs.
We also present the results in Table~\ref{tab:vb_bg}. When background knowledge is appended to variable names, all LLMs show a decrease in F1 and accuracy, together with higher SHD and SID, indicating that performance deteriorates under this prompt format. {Figs.~28 and~29 in Appendix~VI-B further compare the two formats.} While some open-source models achieve marginal gains on specific datasets, such as Earthquake and Water, their overall F1 scores are lower with enriched prompts. In contrast, closed-source models exhibit consistent improvements across all datasets, particularly on medium and large scales.

\begin{figure}[t]
\centering
\subfigure[F1 score]{\includegraphics[width=0.48\linewidth]{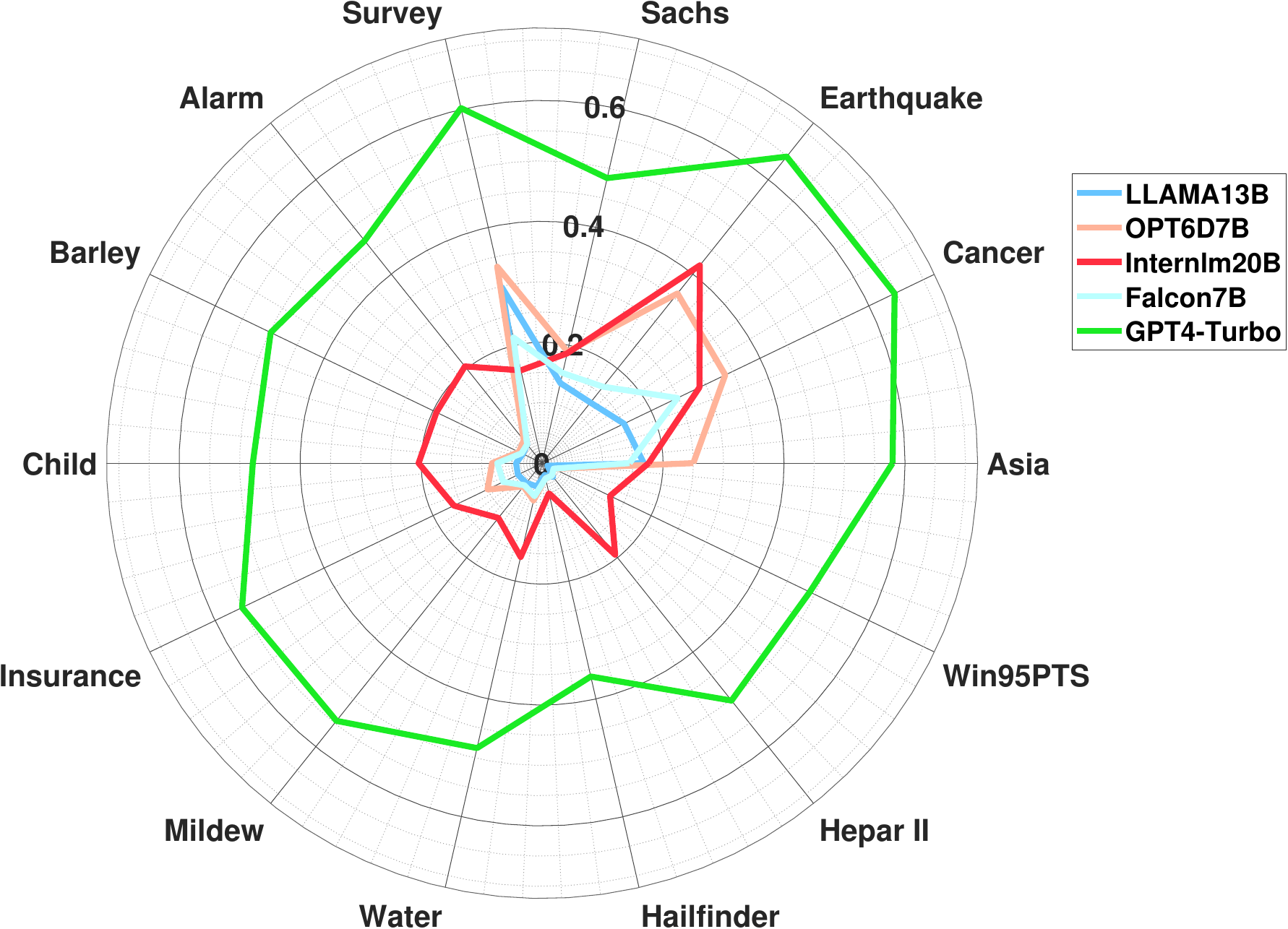}\label{fig:kon_cau_F1}}
\subfigure[Accuracy]{\includegraphics[width=0.48\linewidth]{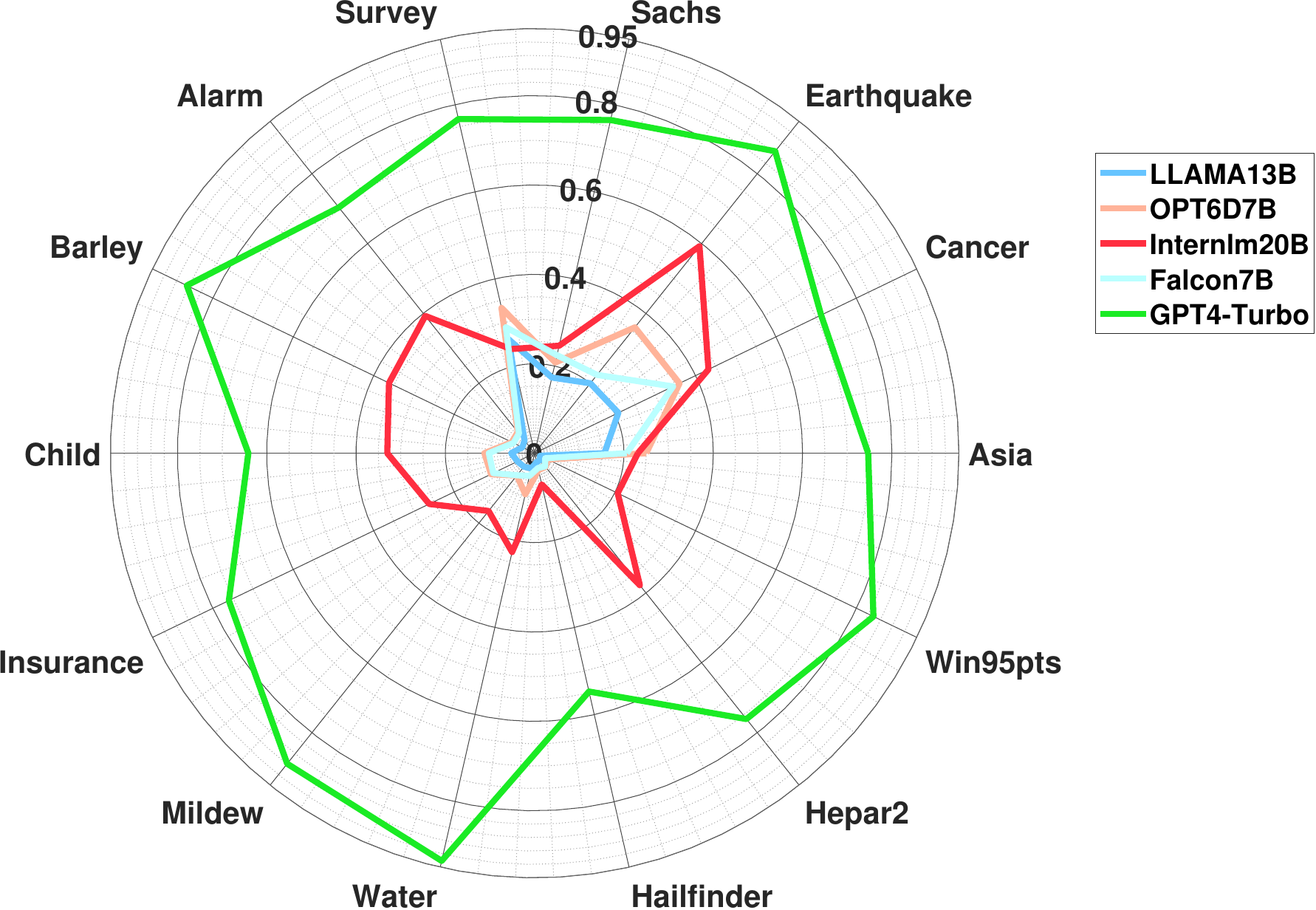}\label{fig:kno_cau_Acc}}\\
\caption{Performance of causality identification for prompt format ``variable name + background knowledge''.}
\label{fig:kno_cau}
\end{figure}

Interestingly, augmenting prompts with background knowledge does not improve performance for most open-source LLMs. In contrast, closed-source LLMs benefit substantially from enriched prompts, especially on medium and large datasets. This counterintuitive finding may be explained by two factors:
\begin{itemize}
\item While combining variable names with background knowledge introduces more information, it makes prompts complex and lengthy, which might hinder the comprehension and reasoning abilities of open-source LLMs. 
\item Due to the prohibitively long inference time with these complex prompts, this evaluation task does not include LLMs with large parameter scales like Falcon-40B, OPT-66B, or LLAMA-33B. 
We suspect that the parameter scale of the tested open-source LLMs limits their ability to understand prompts of variable names with background knowledge. Notably, InternLM-20B, the evaluated open-source LLM with the largest parameters, shows an improvement in F1 score across seven causal datasets when using enriched prompts.
Closed-source LLMs are unaffected by inference time limitations due to API access and boast parameter scales up to 180B. Thus, they demonstrate superior text comprehension and reasoning, allowing them to better understand and reason through datasets with variable names and background knowledge. This is supported by SHD in Fig. 29 in Appendix~VI-B.
\end{itemize}

\subsection{Evaluation on Using Variable Name + Training Data as Prompt} \label{sec:sd}

We evaluate LLMs using enriched prompts that combine variable names with training data, with detailed prompt construction provided in Appendix~IV-C. Following common practice, we utilize 500, 1000, or 1500 observed samples. Due to token limitations, only GPT-series LLMs can process the training data after splitting samples into smaller segments, while open-source models fail to handle such inputs. Thus, we restrict our analysis to GPT models, and summarize the results in Table~XIX of Appendix~VI-C. LLMs exhibit notable gains on small and medium datasets compared to variable names alone. This improvement is attributable to the richer observed samples available at these scales. In contrast, performance on large datasets does not improve and in some cases deteriorates, suggesting that LLMs struggle to process larger numerical matrices.
This phenomenon is interesting because it is well established that incorporating task-related additional information can enhance LLMs' causal learning ability \cite{kiciman2023causal}.
Upon analysis, we attribute this to two factors: 
\begin{itemize}
\item {Training data is directly input to LLMs in the form of numerical matrices. {However, existing LLMs are considered to struggle with handling large scale numerical problems and reasoning} \cite{huang2024exploring}. LLMs are better at handling non-numerical problems and text-based inputs, such as natural language processing tasks \cite{chang2023survey}. }
\item  As the number of nodes increases, combining five submatrices into a complete dataset through interaction with LLMs might fail.
For example, with Alarm dataset, we only need to input data with dimensions of 37 $\times$ 100 in five iterations. However, for Win95PTS dataset, we need to input training data with dimensions of 76 $\times$ 100 five times, and LLMs may struggle to remember each complex data sample in this larger search space.
\end{itemize}

We also compare this format with ``variable names + background knowledge" in Table~\ref{tab:vb_bg}. On small and medium datasets, training data generally yield superior results, whereas on large datasets, neither background knowledge nor training data provides substantial benefits. 
This gap may arise because:
\begin{itemize}
\item Variable names in many datasets may not convey specific meanings, limiting the effectiveness of using background knowledge. For example, some variable names in Water dataset, such as \textit{s2251} and \textit{s2615}, can hardly be associated with precise background knowledge in Wikipedia or other encyclopedias, and LLMs cannot accurately utilize their rich prior knowledge in these cases. 
\item Training data directly provides numerical values, allowing LLMs to utilize their limited numerical processing capabilities more effectively than background knowledge. 
\end{itemize}

\begin{table}[t]
\centering
\caption{Performance of prompt formats ``variable name" and ``variable name + background knowledge"}
\label{tab:vb_bg}
\scalebox{0.5}{
\begin{tabular}{lcccccccccc}
	\toprule
	Metric & \multicolumn{2}{c}{F1 score} & \multicolumn{2}{c}{Accuracy} & \multicolumn{2}{c}{SHD} & \multicolumn{2}{c}{SID} & \multicolumn{2}{c}{{Network Sparsity}} \\\cmidrule(r){1-2}
	\cmidrule(r){2-3} \cmidrule(lr){4-5} \cmidrule(lr){6-7} \cmidrule(lr){8-9} \cmidrule(l){10-11}
	Causal Network & Var.  & Var. + B. K. & Var. & Var. + B. K. & Var. & Var. + B. K. & Var. & Var. + B. K.  & Var. & Var. + B. K. \\
	\midrule
	Asia & 0.3252 & 0.3034 & 0.4192 & 0.3908 & 33.95 & 39.40 & 7.436 & 7.778 & 0.6651 & 0.7652 \\
	Cancer & 0.3719 & 0.3939 & 0.4725 & 0.4527 & 12.88 & 13.80 & 4.819 & 4.921 & 0.7054 & 0.8075 \\
	Earthquake & 0.4246 & 0.4219 & 0.5070 & 0.5389 & 12.10 & 11.72 & 4.526 & 4.424 & 0.7142 & 0.6395 \\
	Sachs & 0.3101 & 0.2658 & 0.3316 & 0.3324 & 76.12 & 80.54 & 10.57 & 11.22 & 0.7892 & 0.8094 \\
	Survey & 0.3994 & 0.3526 & 0.4811 & 0.4439 & 17.88 & 20.06 & 5.338 & 5.748 & 0.6667 & 0.7463 \\
	Barley & 0.2496 & 0.1814 & 0.3953 & 0.3233 & 75.92 & 1527.83 &17.22 & 46.87	 & 0.3454	 & 0.6960 \\
	Child	& 0.2176	& 0.2317	&0.3518	&0.3545	&221.57	&259.27	&18.20	&20.05	&0.6099	&0.7168 \\
	Insurance	&0.2400	&0.2179	&0.3387	&0.3039	&431.20	&513.94	&24.83	&27.55	&0.6399	&0.7793\\
	Mildew	&0.1506	&0.2067	&0.2267	&0.3397	&904.55	&811.98	&33.38 &34.60	&0.7806	&0.7011 \\
	Water	&0.2057	&0.1974	&0.2982	&0.3782	&591.56	&569.17	&33.72	&27.60	&0.6214	&0.5794\\
	Hailfinder	&0.1332	&0.1435	&0.2232	&0.2216	&2144.57	&2442.81	&51.81	&57.30	&0.6599	&0.8097 \\
	Hepar II	&0.1764	&0.1852	&0.3337	&0.3037	&2845.52	&3421.87	&64.13	&71.03	&0.6152	&0.7320\\
	Win95PTS	&0.2131	&0.1770	&0.3977	&0.3363	&1037.62	&3801.99	&52.01	&74.95	&0.3532	&0.6777 \\
	\bottomrule
	\end{tabular}}
	
\end{table}

\subsection{Evaluation on Using Variable Name + Background Knowledge + Training Data as Prompt}
\label{sec:bk+sd}
We examine the performance of LLMs when prompted with variable names and background knowledge, and with training data as the prompt. Specifically, we first combine variable names with the training data as input prompts and interact with the LLM to recognize it. We split the 500 observed samples for each node of the causal dataset into five 100-dimensional segments in dimensional order and fed them to the LLM, whose prompts are provided in Appendix IV-D. Then, the LLM  prompts us to input the corresponding part number and sample data. After receiving each part's number and training data, the LLM reassembles the split sample data according to the part numbers and data, thereby recognizing the training data. Then, we maintain the interaction and introduce variable names, along with background knowledge, as inputs to pose questions to the LLM. This combined format substantially improves causality identification compared with the other three prompt formats. The results suggest that incorporating both semantic context and numerical evidence enables LLMs to better capture causal relations.
We also observe that training data contribute more to performance gains than background knowledge, yet integrating both yields the strongest results overall. Importantly, this improvement depends on clear and semantically meaningful variable names, indicating that LLMs rely on semantic associations rather than on contextual descriptions or numerical distributions alone. Among all formats, variable names combined with background knowledge and training data deliver the best performance across LLMs, so this prompt format will be used to compare with traditional causal learning methods.

\section{Evaluation on Other Influencing Factors}	
\label{sec:exp_ana}
In this section, we further analyze the factors that affect LLMs' causal learning capabilities.
Below, we first consider the effect of the used data, then investigate the causal networks generated by LLMs and analyze their in/out-degrees.
Then, we examine prompt robustness to explore the impact of sentence templates and variable names.  
Finally, we compare the performance of the best LLM-based method on  CausalBN-Bench with that of traditional causal learning algorithms.

\subsection{Data Identification Analysis}\label{sec:fine-gain}
We examine the impact of data granularity on the performance of LLMs, with results in Table XX of Appendix VII-A. In this context, ``understanding" refers not to producing correct answers, but to correctly parsing the question, recognizing the input, and subsequently engaging in causal learning.
Overall, most LLMs can process data from coarse to fine granularity, except for BERT series and OPT-1D3B, which fail to recognize prompts and generate invalid outputs. When dealing with training data, substantial limitations of LLMs are observed. Open-source LLMs with fewer than 100 billion parameters fail to interpret such data, while GPT-3.5-Turbo can only handle inputs up to $7 \times 500$. GPT-4-Turbo can process table-like or matrix inputs up to $78 \times 500$, but its performance still declines as data granularity increases. Notably, none of the evaluated models can handle training data extended to $1,000$ dimensions.
These findings indicate that existing LLMs are adequate for recognizing coarse and moderately fine data, yet their ability to process highly fine-grained inputs remains limited.

\subsection{Causal Network Analysis}\label{sec:in_out_degree}
In this subsection, we analyze two structural metrics in causal learning: out-degree and in-degree \cite{pearl2009causality}.
Out-degree denotes the number of directed edges emanating from a node, indicating how many other nodes it influences, while in-degree denotes the number of directed edges pointing to a node, reflecting how many other nodes directly influence it.
These metrics are critical for identifying key nodes of influence and dependency, thereby revealing the structure and dynamics of causal relationships~\cite{peters2017elements}.
As shown in Table~XXI of Appendix~VII-A, the DAGs generated by LLMs remain inferior to those by traditional causal learning methods.
Across multiple datasets, the DAGs generated by LLMs exhibit larger average out-degrees and in-degrees, suggesting a tendency to produce overly dense structures.
Among LLMs, closed-source models yield degree values closer to those of classical methods (see the last column ``Average degree" in Table~XXI) than open-source models, yet both categories still fall short of matching the performance of traditional approaches.

\begin{figure}[t]
\centering
\subfigure{
	\includegraphics[width=0.97\linewidth]{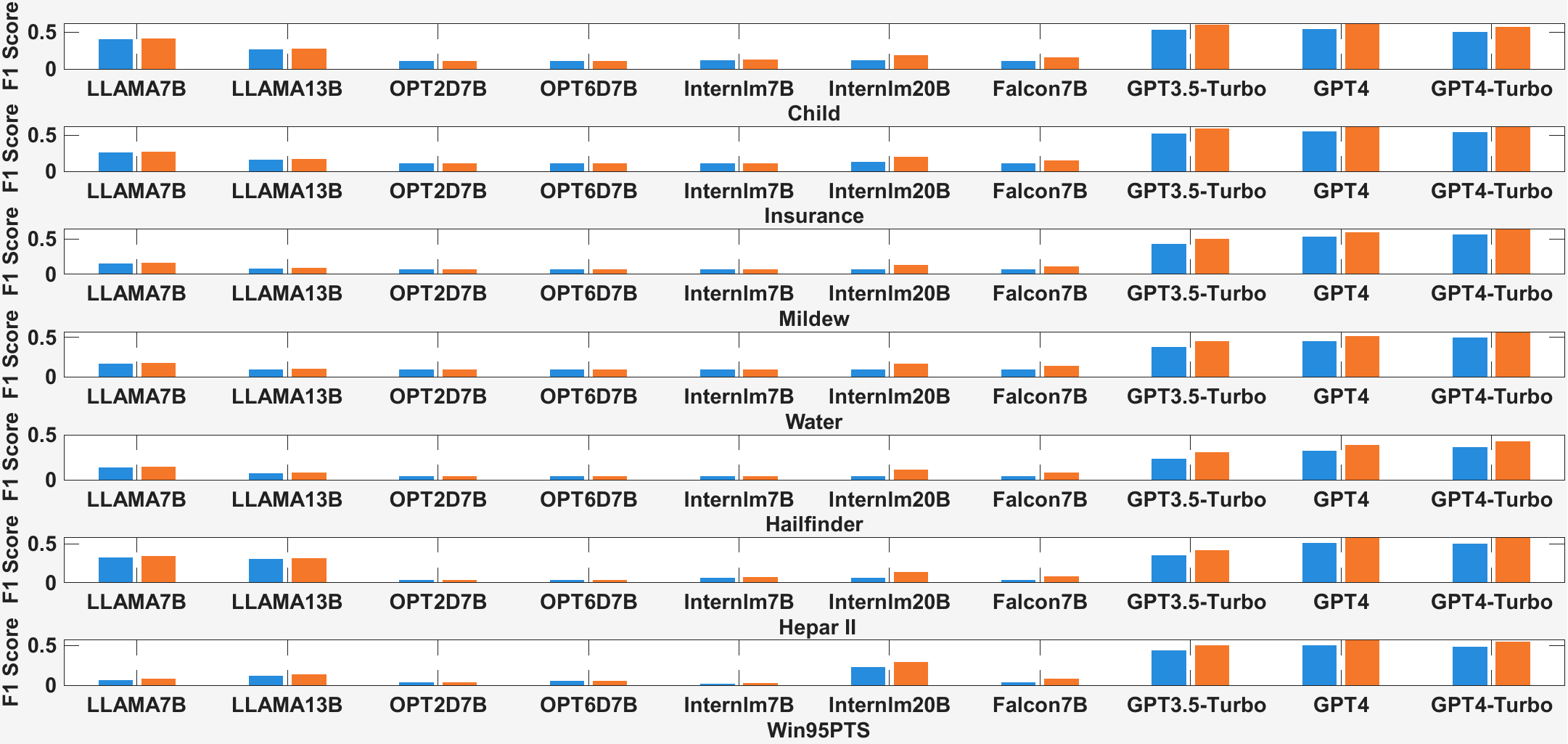}
	\label{fig:cas_var_f1_v1}
}
\vspace{-20pt}
\subfigure{
	\includegraphics[width=0.97\linewidth]{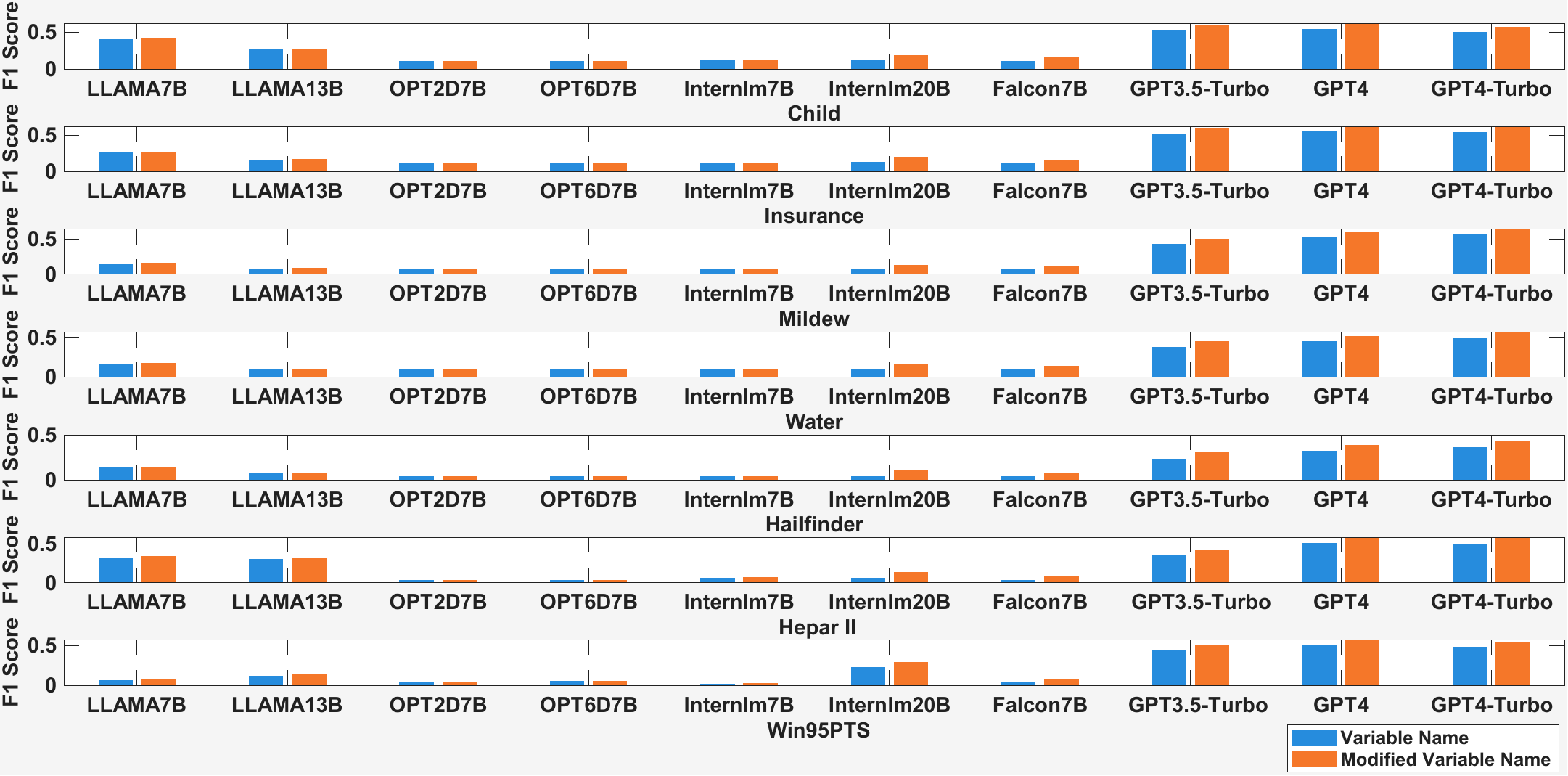}
	\label{fig:cas_var_f1_v2}
}
\caption{{F1 score for prompts using variable names and modified ones.}}
\label{fig:cas_var_f1}
\end{figure}

\subsection{Prompt Robustness Analysis}\label{sec:prompt}

According to \cite{chang2023survey}, the performance of LLMs is significantly related to variable names and sentence templates of prompts. Thus, we test the prompt robustness in the two aspects: \noindent \textbf{Variable Refactorization.}
The robustness of LLMs is evaluated by modifying original variable names to more descriptive forms, such as \textit{Asia} and \textit{Visiting to Asia} discussed in Section~\ref{sec:problem}-C. Since variable names encode meta-semantic information, their representation can directly affect LLMs' ability to identify correlations and causalities. 
To assess this effect, we searched Wikipedia and websites about the causal learning community, such as Bnlearn, to determine the actual meaning of each variable.
Unlike the ``variable name + background knowledge" format, this prompt format remains ``variable name" but with clearer names.
The experimental results, as shown in Fig. \ref{fig:cas_var_f1} and Fig.~30 in Appendix~VII-C, indicate that refactored variable names generally improve both F1 score and accuracy across most LLMs. This suggests that more descriptive names enhance LLMs' recognition of causal relationships. However, negligible improvements are observed on datasets such as Water and Mildew, where variable names (e.g., \textit{s1254}, \textit{s2548}) lack semantic meaning in external knowledge sources. This indicates that performance gains depend on the availability of interpretable and meaningful variable names.

\noindent \textbf{Sentence Paraphrase.}
{We examine the robustness of LLMs to prompt phrasing in the causality identification task using five templates, including three existence-oriented (i.e., Types 1, 2, and 5) and two direction-oriented (i.e., Types 3 and 4). Table~\ref{tab:rw_causality} shows that prompt phrasing has a substantial effect on performance. For existence identification, performance improves from Type 1 to Type 5, with Type 5 achieving the best results. For direction identification, Type 3 consistently outperforms Type 4. Closed-source LLMs outperform open-source LLMs across all prompt types, especially on direction identification. These results indicate that the optimal prompt wording depends on whether the task focuses on causal existence or causal direction.}
Comparison with correlation identification results in Table XXII in Appendix VII-D further reveals that prompt phrasing exerts a stronger influence on causality tasks than on correlation tasks.

\begin{table}[t]  
\centering  
\caption{Sentence paraphrase for causality identification task}
\label{tab:rw_causality}	\scalebox{0.45}{{
		\begin{tabular}{cllccccc}    \toprule    \multicolumn{2}{c}{Causality identification task} & \multicolumn{1}{c}{LLMs} & F1    & Accuracy & SHD   & SID   & Network sparsity \\    \midrule    \multirow{9}[6]{*}{Existence} & \multirow{3}[2]{*}{Type 1: Are Var. A and Var. B causally related?} & Open-Source LLMs & 0.2614 & 0.3018 & 612.47 & -     & 0.9052 \\          &       & Closed-Source LLMs & 0.3729 & 0.6126 & 438.91 & -     & 0.5124 \\          &       & All LLMs & 0.2897 & 0.3665 & 564.83 & -     & 0.8461 \\\cmidrule{2-8}          & \multirow{3}[2]{*}{Type 2: Is there a causal connection between Var. A and Var. B?} & Open-Source LLMs & 0.2756 & 0.3151 & 598.30 & -     & 0.8927 \\          &       & Closed-Source LLMs & 0.3898 & 0.6354 & 401.25 & -     & 0.4868 \\          &       & All LLMs & 0.3041 & 0.3812 & 536.62 & -     & 0.8295 \\\cmidrule{2-8}          & \multirow{3}[2]{*}{Type 5: Is there causality between Var. A and Var. B?} & Open-Source LLMs & 0.2861 & 0.3274 & 585.66 & -     & 0.8871 \\          &       & Closed-Source LLMs & 0.4086 & 0.6627 & 372.14 & -     & 0.4659 \\          &       & All LLMs & 0.3155 & 0.3968 & 517.39 & -     & 0.8167 \\    \midrule    \multirow{6}[4]{*}{Direction} & \multirow{3}[2]{*}{Type 3: Does {Var. A} cause Var. B?} & Open-Source LLMs & 0.2409 & 0.2906 & 852.82 & 25.74 & 0.8487 \\          &       & Closed-Source LLMs & 0.3845 & 0.8342 & 178.52 & 26.28 & 0.1335 \\          &       & All LLMs & 0.2738 & 0.4152 & 698.17 & 25.86 & 0.6847 \\\cmidrule{2-8}          & \multirow{3}[2]{*}{Type 4: Does Var. A influence Var. B?} & Open-Source LLMs & 0.2052 & 0.2217 & 943.64 & 25.99 & 0.9413 \\          &       & Closed-Source LLMs & 0.3462 & 0.5983 & 610.96 & 29.04 & 0.4668 \\          &       & All LLMs & 0.2375 & 0.3081 & 867.34 & 26.69 & 0.8325 \\    \bottomrule    \end{tabular}} }
		\end{table}
		
		\subsection{Few-shot In-Context Learning Evaluation}
		
		{To examine whether a small number of task examples can improve LLMs' causal learning, we conduct few-shot in-context learning (ICL) experiments across three evaluation tasks. Following \cite{chen2024causal}, we consider 1-shot and 3-shot settings and sample examples, which are different from the target test graph to avoid answer leakage. F1 score and accuracy are reported in Tables XXIII and XXIV in Appendix VII-E. The results show that few-shot ICL consistently improves performance on all three tasks, with the largest gains on causal skeleton identification, moderate gains on causality identification, and relatively limited gains on correlation identification. In general, the improvements are more evident in larger, more complex causal networks. Detailed settings and further analysis are provided in Appendix VII-E.}
		
		\subsection{Parameter-efficient Fine-tuning Evaluation}
		{We further examine whether parameter-level adaptation can improve LLM performance on causal learning tasks. Specifically, we introduce fine-tuned baselines for the causality task where each sample is formulated as a binary question over variables in a causal network. Supervision data are constructed using a subgraph-driven strategy based on the ground-truth graph to reduce information leakage. We consider both in-distribution and out-of-distribution settings, and use LoRA \cite{hu2022lora} for parameter-efficient adaptation. F1 score and accuracy in Table XXV of Appendix VII-F show that fine-tuning improves performance on the causality task. The gains are largest and most stable in the in-distribution setting, while the out-of-distribution setting also improves many LLMs, but with smaller gains. This suggests that parameter-level adaptation can enhance causal discrimination with supervised data, although cross-network generalization remains challenging. Detailed results and analysis are provided in Appendix VII-F.}

		\subsection{{Performance Comparison of LLMs vs. Traditional Causal Learning Methods}} 
		\label{sec:com_classical}
		{Table \ref{tab:comparison_classic} provides SHD of {the best-performing LLM method} 
(GPT4-Turbo with prompt format ``variable names, background knowledge and training data'') with traditional causal learning algorithms, including classic algorithms (i.e., HC \cite{buntine1994operations}, MMHC \cite{tsamardinos2006max}, PC \cite{spirtes1991algorithm}, and GES \cite{chickering2002optimal}) and SOTA algorithms (i.e., MIGA \cite{YanMutual2023}, DAG-GNN \cite{yu2019dag}, ENSAOBS \cite{wang2024improved}, and LiNGAM \cite{shimizu2011directlingam}). The detailed rationale for choosing these traditional methods is provided in Appendix VIII-B.}
It is evident that on small-scale datasets, the LLM-based method achieves performance comparable to classical approaches but still lags behind SOTA algorithms, such as MIGA and ENSAOBS. On medium scale datasets, the LLM-based method falls short of classical methods and is significantly outperformed by SOTA methods. In large scale datasets, the LLM-based method is substantially inferior to both classical and SOTA methods.  
Then, we conclude that the current capabilities of LLM-based methods have not yet reached the level of existing methods. Additionally, LLM-based methods are inferior to traditional causal learning algorithms at scales above 50 nodes, usually failing to reach 50\% of their performance metrics.

\begin{table}[tbp]
\centering
\caption{SHD of different causal learning algorithms}
\scalebox{0.56}{{
		\begin{tabular}{lccccccccc}    
			\toprule    
			\multicolumn{1}{c}{\multirow{2}[4]{*}{Datasets}} & Best LLM method & \multicolumn{4}{c}{Classical methods} & \multicolumn{4}{c}{{SOTA methods}} \\\cmidrule{2-10}          & GPT4-Turbo with B. K. + T. D. & HC    & MMHC  & PC    & GES   & DAG-GNN & ENSAOBS & MIGA  & LiNGAM \\    \midrule    
			Asia  & 8     & 10    & 4     & 8     & 6     & 8     & 1     & 2     & 12 \\    
			Cancer & 4     & 4     & 3     & 3     & 4     & 5     & 4     & 3     & 7 \\    
			Earthquake & 1     & 7     & 3     & 2     & 5     & 4     & 1     & 2     & 6 \\    
			Sachs & 54    & 12    & 14    & 10    & 15    & 17    & 13    & 4     & 34 \\    
			Survey & 10    & 4     & 3     & 1     & 3     & 4     & 4     & 1     & 9 \\    
			Alarm & 88    & 49    & 27    & 36    & 41    & 70    & 29    & 12    & 130 \\    
			Barley & 174   & 95    & 81    & 77    & 86    & 98    & 84    & 65    & 225 \\    
			Child & 42    & 20    & 15    & 29    & 23    & 23    & 12    & 11    & 72 \\    
			Insurance & 152   & 48    & 29    & 45    & 44    & 66    & 31    & 21    & 175 \\    
			Mildew & 103   & 59    & 35    & 34    & 41    & 75    & 51    & 29    & 162 \\    
			Water & 70    & 56    & 55    & 68    & 63    & 62    & 66    & 47    & 142 \\    Hailfinder & 1225  & 59    & 49    & 56    & 58    & 67    & 40    & 38    & 340 \\    
			Hepar II & 1286  & 101   & 92    & 117   & 104   & 123   & 92    & 79    & 445 \\    
			Win95PTS & 909   & 153   & 89    & 119   & 126   & 189   & 161   & 56    & 520 \\   
			\bottomrule    
\end{tabular}}}%
\label{tab:comparison_classic}%
\end{table}%

\section{Conclusion and Future Trends} 
\label{sec:discussion}
In this section, we present conclusions based on the evaluation outcomes and analyses, followed by insights into future directions on augmenting the causal learning ability of LLMs.

\subsection{Conclusion}
To explore the causality identification capabilities of LLMs, we develop a comprehensive benchmark called CausalBN-Bench. 
Through systematic evaluation across diverse datasets, prompt formats, and tasks, several key findings emerge:
\begin{itemize}
\item While LLMs exhibit clear promise, their efficacy in causal learning still lags behind human capabilities. Their performance remains inferior to classical and SOTA causal learning algorithms, particularly on medium- and large-scale datasets. Closed-source models consistently outperform open-source counterparts, yet even the best LLMs fall short of established methods. They usually fail to reach 50\% of the performance metrics of classical and SOTA algorithms above 50 nodes. Furthermore, in small scale networks, LLMs exhibit superior performance in identifying correlations compared to causality. However, in large scale networks, there are no statistically significant differences across the three evaluation tasks, indicating that the performance is not influenced by the difficulty of the evaluation tasks. When LLMs identify correlation and causality, they primarily rely on contextual semantic knowledge rather than recognizing differences in underlying distributions.

\item In causal networks, LLM performance is not uniform, fluctuating with task depth and difficulty. Open-source models struggle more with causality identification than closed-source ones, and overall accuracy and F1 scores decline as dataset size increases. LLMs are generally proficient in recognizing chain structures but perform poorly on colliders, while showing relative strength in CoT-analogous long-chain causal structures. Structural analysis further reveals that LLM-generated DAGs are considerably denser than those derived from traditional algorithms, with many correct edges but also a substantial number of erroneous ones.

\item Background knowledge can play an important role in the causality identification of LLMs, but its benefits are not consistent. When variable names are clear and semantically meaningful, background knowledge substantially improves performance. For ambiguous variable names, however, it provides little advantage, suggesting that LLMs rely more on semantic associations of explicit entities than on contextual descriptions or numerical distributions. Training data also exerts a strong influence: recent closed-source LLMs can effectively exploit such data, whereas earlier models and most open-source variants remain largely dependent on textual information. In addition, performance is sensitive to sentence structures and the semantics of variable names, underscoring the challenges LLMs face in causal reasoning.
\end{itemize}

{Based on the above findings, current LLMs still struggle with causal learning in many scenarios such as large-scale causal networks. Further discussions about error analysis, performance degradation on large-scale causal networks, and practical implications are provided in Appendices VIII-C, VIII-D, and VIII-E.}

\subsection{{Future Trends}}
Finally, we propose future trends for LLMs in causal learning and offer perspectives on methods to enhance LLMs' ability to identify causal relationships.
\begin{itemize}
\item{\bf Continuous evolution for CausalBN-Bench to keep pace with LLM advancements:}
With the rapid development of LLMs, their causal learning capabilities will further improve. CausalBN-Bench will also evolve in tandem with the development of LLMs, including the addition of datasets with more nodes and more complex networks, as well as the evaluation of causal structures similar to some more advanced LLM prompting techniques, such as Tree-of-Thought \cite{yao2024tree}. 
CausalBN-Bench will be continuously updated to maintain a comprehensive evaluation of the causal learning capabilities of LLMs.
\item {\bf Enhancing the abilities of causality identification for LLMs:} 
We aim to improve the causal learning abilities of LLMs, which can better facilitate mechanism understanding and explainability on other complex tasks, such as counterfactuals and interventions. From the results on CausalBN-Bench, existing LLMs are not good at identifying collider structures and perform poorly on large-scale sparse networks. Therefore, we will fine-tune LLMs using causal structures with specific properties purposefully and strategically in future work.
\item {\bf Strengthening LLMs' utilization of background knowledge and training data for causal learning-related tasks:} We recommend fine-tuning LLMs to integrate better background knowledge and process training data, where we aim to improve the representation of input data to distinguish and leverage the semantic information of variable names and enhance understanding of different sentence structures.
Additionally, designing expert LLMs to recognize medium to large scale causal networks with unclear variable names and to recognize numerical inputs are essential research directions.
\end{itemize}

{In the future, we intend to extend CausalBN-Bench beyond causal learning toward effect estimation. A detailed discussion is provided in Appendix VIII-F. Inspired by causal effect estimation, we are also concerned about a causal strength evaluation task, a quantitative analysis of each ranked variable in their cause or effect within the range of [0,1], whose prompts are given in Appendix~IX-A}. This experiment aims to pave the way for future research to more precisely examine how LLMs assess causal strength and quantify it, thereby assisting scholars in more thoroughly utilizing and recognizing causal relationships between variables.
{To further explore this direction, we introduce a token-generation probability evaluation for causal strength \cite{takayamaintegrating}. For each variable pair, we sample the same prompt and average the generation probability of the token ``yes'', whose detailed results are provided in Appendix~IX-B}. This experiment provides an alternative to discrete evaluation as an initial step toward continuous causal strength assessment.

\vspace{0.1in}
\vspace{-10pt}
\bibliographystyle{IEEEtran}
\bibliography{Causal}

\appendices

\setcounter{section}{0}
\renewcommand{\thesection}{\Roman{section}}

\onecolumn

\begin{center}
	\Huge {Appendix to ``CausalBN-Bench:~A Comprehensive Benchmark for Causal Learning Capability of \\Large Language Models''}
\end{center}
\vspace{5pt}
\noindent{\large \itshape This appendix provides supplementary materials for the main manuscript and is organized as follows:}
\vspace{5pt}
\begin{itemize}
	\item \ref{app:related_work}~\nameref{app:related_work}
	\begin{itemize}
		\item \ref{app:causal_learning}.~\nameref{app:causal_learning}
		\item \ref{app:causal_LLM}.~\nameref{app:causal_LLM}
		\item \ref{app:rela_work_reasoning}.~\nameref{app:rela_work_reasoning}
		\item \ref{app:deteld_comparision_des}.~\nameref{app:deteld_comparision_des}
	\end{itemize}
	\item \ref{app:exp_setup}.~\nameref{app:exp_setup}
	\begin{itemize}
		\item \ref{app:detailed_metrics}.~\nameref{app:detailed_metrics}
		\item \ref{app:LLMs_v1}.~\nameref{app:LLMs_v1}
		\item \ref{app:add_exp_set}.~\nameref{app:add_exp_set}
		\item \ref{app:lable_gt_direct}.~\nameref{app:lable_gt_direct}
		
	\end{itemize}
	\item \ref{app:prompt_task}.~\nameref{app:prompt_task}
	\begin{itemize}
		
		\item \ref{app:correlation}.~\nameref{app:correlation}
		\item \ref{app:causal_skeleton}.~\nameref{app:causal_skeleton}
		\item \ref{app:causal}.~\nameref{app:causal}
		\item \ref{app:cot}.~\nameref{app:cot}
		
	\end{itemize}
	\item \ref{app:prompt_format}.~\nameref{app:prompt_format}
	\begin{itemize}
		\item \ref{app:var}.~\nameref{app:var}
		\item \ref{app:var_bk}.~\nameref{app:var_bk}
		\item \ref{app:var_sd}.~\nameref{app:var_sd}
		\item \ref{app:var_bk_sd}.~\nameref{app:var_bk_sd}
	\end{itemize}
	\item \ref{app:add_exp_evaluation}.~\nameref{app:add_exp_evaluation}
	\begin{itemize}
		\item \ref{app:dir_relat}.~\nameref{app:dir_relat}
		\item \ref{app:indir_relat}.~\nameref{app:indir_relat}
		\item \ref{app:add_ske}.~\nameref{app:add_ske}
		\item \ref{app:add_cau1}.~\nameref{app:add_cau1}
		\item \ref{app:add_cau2}.~\nameref{app:add_cau2}
		\item \ref{app:cot_table}.~\nameref{app:cot_table}
	\end{itemize}
	\item \ref{app:add_exp_prompt}.~\nameref{app:add_exp_prompt}
	\begin{itemize}
		\item \ref{app:add_kno_cau}.~\nameref{app:add_kno_cau}
		\item \ref{app:kno_cas_comparion}.~\nameref{app:kno_cas_comparion}
		\item \ref{app:diff_prompt}.~\nameref{app:diff_prompt}
	\end{itemize}
	\item \ref{app:add_exp_other}.~\nameref{app:add_exp_other}
	\begin{itemize}
		\item \ref{app:fine-grained}.~\nameref{app:fine-grained}
		\item \ref{app:causal_network}.~\nameref{app:causal_network}		
		\item \ref{app:acc_modified_var}.~\nameref{app:acc_modified_var}
		\item \ref{app:sen_par_cor}.~\nameref{app:sen_par_cor}
		\item \ref{app:fewshot}.~\nameref{app:fewshot}
		\item \ref{app:fine-tuned}.~\nameref{app:fine-tuned}		
	\end{itemize}
	\item \ref{app:discussion}.~\nameref{app:discussion}
	\begin{itemize}
		\item \ref{app:task_design}.~\nameref{app:task_design}
		\item \ref{app:tranditional_cl}.~\nameref{app:tranditional_cl}
		\item \ref{app:error_an}.~\nameref{app:error_an}
		\item \ref{app:large_network}.~\nameref{app:large_network}
		\item \ref{app:practical}.~\nameref{app:practical}	
		\item \ref{app:future_reasoning}.~\nameref{app:future_reasoning}
		
	\end{itemize}
	\item \ref{app:future}.~\nameref{app:future}
	\begin{itemize}
		\item \ref{app:cau_str}.~\nameref{app:cau_str}
		\item \ref{app:future_probality}.~\nameref{app:future_probality}	
	\end{itemize}
\end{itemize}

\noindent{\large \textbf{Remark:}} The implementation of a case of CausalBN-Bench is available at \url{https://anonymous.4open.science/r/CausalBN-Bench-FEE8}. The complete source code and datasets will be made publicly available after the review process is complete.

\clearpage
\section{Comprehensive Related Work on CausalBN-Bench}\label{app:related_work}
\subsection{Preliminaries of Causal Learning}\label{app:causal_learning}

\subsubsection{Graphical Models}
Causal learning aims to recover a causal model from data, which are typically sampled from conditional probability. To this end, we first introduce the causal modeling, which is mainly divided into two categories, structural causal model (SCM) and causal network (also called Bayesian network (BN)).

The SCM, formally represented as $M = (V, U, F, P)$, offers a rigorous framework for delineating causal systems. In SCM, $V$ is the observable variable, $U$ represents the unobservable variable, $F$ is a set of functions articulating causal mechanisms, and $P$ represents the probability distributions of these functions. This model enables a systematic exploration of the dynamics among system variables, mapping out explicit causal connections and pathways.

A causal network, denoted as $B$, is defined as a triple $(G, X, P_X)$, where $G = (V, E)$ represents a directed acyclic graph (DAG) \cite{pearl2009causality}. In this DAG, $V$ is the set of nodes, and $E \subset V \times V$ is the set of edges, which delineate the causal relationships between nodes in a directed manner. The set $X = \{X_1, X_2, \ldots, X_n\}$ comprises random variables, each associated with a node in $V$. Additionally, $P_X$ denotes the joint distribution of these variables. Within the context of a directed graph, an edge $X \rightarrow Y$ delineates a directional relationship, which is uniquely different from $Y \rightarrow X$. A path within a graph is defined as a sequence of vertices that are consecutively connected, with a cycle being identified when a path initiates and concludes at the identical vertex. A DAG represents a directed graph devoid of cycles, frequently employed to depict causal frameworks. A fundamental principle in BNs, known as the faithfulness assumption, asserts that $P_X$ exclusively reflects the independence conditions evident in $G$. This paper adopts the faithfulness assumption, ensuring that the BN faithfully represents the underlying causal structure.

\subsubsection{Traditional Causal Learning Methods}
Causal learning and BN structure learning are closely related fields that focus on uncovering and refining causal connections within datasets. At their core, both domains aim to ascertain the causal architecture, encapsulated as a DAG, derived from observational data. This process involves analyzing a collection of variables, denoted as $\{X_1, X_2, \ldots, X_n\}$, to determine the optimal causal graph $G^*$ that accurately reflects the inherent causal mechanisms. Accordingly, traditional causal learning techniques are dichotomized into two principal categories \cite{vowels2022d}: 
\begin{itemize}
	\item \textbf{Constraint-based methods}: Constraint-based methodologies initiate the process of uncovering BN structures by pinpointing conditional independence relationships among variables using statistical techniques, including Pearson’s chi-square test. Subsequently, the BN architecture that optimally aligns with these identified relationships is formulated. Notable algorithms in this domain encompass the PC algorithm \cite{spirtes1991algorithm} and the grow–shrink algorithm. The exponential growth in computational power, notably with the deployment of GPUs, has paved the way for analyzing voluminous, high-dimensional datasets. Current approaches have evolved to reformulate the BN structure learning challenge as a constrained continuous optimization problem. This paradigm shift leverages DL technologies \cite{yu2019dag, wang2024improved} to tackle the intricacies of this problem effectively.		
	
	\item \textbf{Score-based methods}: Score-based methods assess the adequacy of potential Bayesian network structures through a specific scoring metric. Such methods treat the BN structure learning problem as a combinatorial optimization problem. Traditional search techniques aimed at finding a single solution, including the hill climbing algorithm (HC) \cite{buntine1994operations} and simulated annealing, often encounter difficulties due to entrapment in local optima. To overcome these limitations, evolutionary computation based methods, as population-based search strategies including the genetic algorithm \cite{YanMutual2023}, aim to unearth optimal structures by exploring a broader solution space.

\end{itemize}

\subsubsection{D-Separation}
DAGs are typically assumed to adhere to the Markov property, indicating that the implied joint distribution factorizes according to the recursive decomposition characteristic of BNs \cite{pearl2009causality}:

\begin{equation}
	P(X) = \prod_{i=1}^{d} P(X_i | {pa}_i)
\end{equation}

This decomposition is related to the concept of d-separation. For two vertices $X_i$ and $X_k$, they are d-separated by a disjoint set of variables $S$ if $X_j \in S$ in any of the following structural scenarios \cite{pearl2009causality}:
\begin{align*}
	X_i &\rightarrow X_j \rightarrow X_k \\
	X_i &\leftarrow X_j \leftarrow X_k \\
	X_i &\leftarrow X_j \rightarrow X_k \\
\end{align*}

They are also d-separated if $X_j$ and none of its descendants are in $S$ in the following structural scenario:

\begin{equation}
	X_i \rightarrow X_j \leftarrow X_k
\end{equation}

If the DAG's d-separation properties uphold an assumption of faithfulness, they imply Markov conditional independencies in the joint distribution, denoted as $X_i \perp\!\!\!\perp X_k | X_j$. As for the DAG, disjoint (i.e., non-overlapping) sets of variables, $A$ and $B$ are d-separated by $S$ in $G$ if $A \perp\!\!\!\perp_{{d-sep}} B | S$, and conversely, are d-connected if this conditional independence does not hold in the graph. The assumption of d-faithfulness is that any conditional independencies implied by the graph, according to its d-separation properties, are reflected in $P_X$. For $P_X$ and $G$, the assumption of d-faithfulness holds if:

\begin{equation}
	A \perp\!\!\!\perp_{P_X} B | C \Rightarrow A \perp\!\!\!\perp_{{d-sep}} B | C
\end{equation}

An example of a violation of d-faithfulness occurs when the influences of two paths cancel each other out, leading to a DAG with different implied conditional independencies than those present in the joint distribution.

\subsubsection{Markov Property}
In causal learning, the Markov property is of fundamental importance. It asserts that within a BN each variable is conditionally independent of its non-descendants, given its parents. This assumption of local independence permits the decomposition of the variables' joint probability distribution. Concretely, the joint probability distribution $P_X$ of variables can be decomposed into a product of conditional probabilities for each variable given its parents, denoted as $P_X(X_1, X_2, \ldots, X_n) = \prod_{i=1}^{n} P(X_i | {{Parents}}(X_i))$. This principle not only facilitates causal inference within intricate networks but also underpins numerous causal learning algorithms. These methodologies exploit the Markov property to derive DAGs from data, by discerning and modeling the conditional independencies and dependencies. In this paper, we harness the concepts of d-separation in DAGs and the Markov property to define d-separation-related correlation and to determine the causal skeleton.

\subsection{Causal Learning for LLMs}\label{app:causal_LLM}
{In recent years, the rapid progress of large language models (LLMs) has led to remarkable achievements across a wide range of fields \cite{causal_survey_liu_2025}. By processing massive amounts of text data, these models can effectively distill human knowledge and have demonstrated strong capabilities in reasoning and decision-making. As a result, researchers increasingly believe that LLMs may also absorb human experience, common sense, and heuristic knowledge related to causal learning and optimization design \cite{wu2024causality}. This potential is especially evident in tasks involving reasoning and decision-making, such as planning and mathematical problem solving. Therefore, thanks to their rich prior knowledge,  text comprehension abilities, and emerging potential for causal reasoning, LLMs are considered promising tools for supporting causal learning tasks. Existing studies on LLM-based causal learning can generally be divided into three categories: LLM-based causal discovery methods, LLM-based causal reasoning methods, and other causal tasks with LLMs \cite{ma2025causal,wu2025llm}.}
\subsubsection{LLM-Based Causal Discovery Methods}
{LLM-based causal discovery methods can be divided into three types: direct causal discovery with LLMs, prompt-based causal identification, and hybrid frameworks that combine LLMs with traditional causal discovery methods. Specifically,}
{
	\begin{itemize}
		\item \textbf{Direct causal discovery with LLMs}: this type directly uses LLMs to identify causal relationships, mainly including pairwise causal discovery and full causal graph discovery. Pairwise causal discovery focuses on determining the causal direction between two variables or deciding whether a causal relation exists. These methods typically provide the LLM with variable names and limited contextual information, and ask the model to judge the causal direction. Some studies have shown that models such as GPT-4 can outperform certain traditional methods on specific tasks. At the same time, other works have emphasized that LLM judgments may rely more on statistical patterns embedded in training corpora than on genuine causal reasoning \cite{kiciman2023causal,jin2023can,zevcevic2023causal,chi2024unveiling}. In contrast, full causal graph discovery aims to recover the complete causal graph over all variables in a system. Compared with pairwise discovery, this setting is more complex and usually requires combining LLMs' language understanding, domain knowledge, and data-driven structural learning procedures. For example, Jiralerspong \etal \cite{jiralerspong2024efficient} proposed an efficient LLM-based causal graph discovery framework that uses breadth-first search to reduce query complexity from quadratic to linear and further improves performance by incorporating observational data; Antonucci \etal \cite{antonucci2023zero} studied the ability of LLMs to infer causal relations from natural language in a zero-shot setting and further extended the framework to causal graph recovery through iterative pairwise queries; Long \etal \cite{long2023can} investigated the potential of GPT-3 for assisting causal graph construction in the medical domain and analyzed how prompt design, linking verbs, and language properties affect causal edge identification; and Lyu \etal \cite{zhiheng2022can} examined whether LLMs can distinguish causes from effects and proposed a zero-shot post hoc analysis method based on natural language prompts \cite{ban2023causal}.
		\item \textbf{Prompt-based causal identification}: This line of work studies how prompt construction influences causal identification performance. Prompt formulation is important for both pairwise and whole-graph causal discovery because the way the question is presented affects whether the LLM can correctly interpret variable meanings and infer possible causal relations. Existing studies mainly enrich prompts from three aspects. The first is {variable-level information} \cite{constantinou2025using}, in which variable names, definitions, or attributes are explicitly provided to clarify the entities involved. The second is {background knowledge} \cite{zhang2023understanding,zhang2024causal}, such as task descriptions, domain-specific knowledge, or scientific documents, which provide context that may not be sufficiently captured during pretraining. The third is {training data} \cite{cai2023knowledge}, where the model is provided with raw observations, statistical summaries, or other structured representations so that its causal judgments can be grounded in data evidence. In essence, this direction seeks to strengthen the causal structure identification ability of LLMs by designing prompts that provide clearer and richer information.
		\item \textbf{Hybrid frameworks that combine LLMs with traditional causal discovery methods:} the goal of these hybrid approaches is to combine the rich semantic and commonsense knowledge encoded in LLMs with the structural learning strengths of conventional statistical causal discovery algorithms, to improve the robustness and accuracy of graph reconstruction. Based on the coupling between the two components, existing studies can be broadly organized into four categories. The first category uses LLMs for initialization or prior elicitation, providing candidate causal relations, prior knowledge, or structural constraints for downstream causal discovery procedures \cite{ban2023query,naik2024applying,darvariu2024large}. The second category treats LLMs as post-hoc refiners, in which a traditional causal discovery method first produces an initial graph, and the LLM is then used to revise, interpret, or augment the result \cite{castelnovo2023marrying,khatibi2024alcm,long2023causal}. The third category involves iterative interaction, where LLMs and classical causal discovery algorithms cooperate over multiple rounds to progressively update and improve the inferred structure \cite{abdulaal2023causal,ban2023causal,takayamaintegrating}. The fourth category includes more modular forms of integration, in which LLMs are embedded into particular components of the pipeline, for example by replacing or enhancing modules such as scoring mechanisms or conditional independence testing. Representative studies \cite{cohrs2024large,zhang2024causal} cover several directions, including using LLMs to provide structural priors, combining retrieval-augmented generation with causal graph construction, and developing constrained or autonomous causal discovery systems with LLM support.
	\end{itemize}
}
\subsubsection{LLM-Based Causal Reasoning Methods}
{Unlike structural discovery, LLM-based causal reasoning mainly focuses on the capabilities, limitations, and enhancement strategies of LLMs in causal effect estimation and counterfactual reasoning \cite{liu2024discovery}.
	\begin{itemize}
		\item \textbf{Causal effect estimation}: existing studies \cite{lin2023text,kiciman2023causal} evaluate whether LLMs truly possess causal reasoning ability and also explore how prompt engineering, external knowledge, or tool use can improve their ability to estimate intervention effects \cite{jin2023cladder}.
		For example, Chi \etal \cite{chi2024unveiling} argues that current LLMs rely heavily on causal knowledge already present in training corpora rather than on robust causal reasoning, and further demonstrates the importance of external knowledge enhancement. Kim \etal \cite{kim2025technical} studies how LLMs can be used to generate synthetic data that are more suitable for causal inference, with the goal of better preserving key causal quantities such as the average treatment effect. In addition, several studies attempt to improve causal effect estimation through prompt design and external tools. For instance, Abdali \etal \cite{abdali2023extracting} use in-context learning and self-consistency mechanisms to extract key causal variables from unstructured text and construct causal graphs; Pawlowski \etal \cite{pawlowski2023answering} combine contextual enhancement and tool enhancement to refine causal outputs; CausalCoT \cite{jin2023cladder} employs chain-of-thought prompting to decompose complex queries and improve causal inference accuracy; and Dhawan \etal \cite{dhawan2024end} integrates traditional causal inference techniques to automatically estimate causal effects from unstructured text.
		\item \textbf{Counterfactual reasoning}:  an increasing number of studies have investigated how LLMs perform in ``what-if" scenarios and how their performance can be improved. Chen \etal \cite{chen2025counterbench} introduces a dedicated benchmark for counterfactual reasoning and reveals clear weaknesses of current LLMs on this task. Gendron \etal \cite{gendron2024counterfactual} proposes a complete framework that extracts causal variables from natural language text, constructs causal graphs, and performs counterfactual reasoning. At the same time, datasets such as IfQA \cite{yu2023ifqa}, Clomo \cite{huang2024clomo}, and Crass \cite{frohberg2022crass} have been designed to systematically evaluate counterfactual reasoning ability in LLMs. Tan \cite{tan2023causal} studies how modifying intermediate nodes in chain-of-thought reasoning affects final conclusions. Counterfactual explanation \cite{bhattacharjee2023towards} and commonsense counterfactual generation \cite{miao2023generating} focus on generating counterfactual samples that are both interpretable and consistent with common sense. Chen \etal \cite{chen2023disco} can generate high-quality counterfactual data at scale to improve model robustness. In addition, Kiciman \etal \cite{kiciman2023causal} further demonstrate emerging counterfactual reasoning abilities of LLMs in tasks such as causal DAG generation. Beyond these settings, Acharya \etal \cite{acharya2025causcibench} extends causal reasoning evaluation to scientific analysis scenarios and assesses LLM performance across a complete causal analysis pipeline, including variable selection, method selection, effect estimation, and result interpretation. Overall, this line of research has moved beyond simply asking whether LLMs can restate superficial causal knowledge and instead increasingly focuses on whether they can perform reliable causal inference in higher-level tasks involving intervention, estimation, and counterfactual analysis.
	\end{itemize}
}
\subsubsection{Other causal tasks with LLMs} 
{Beyond causal discovery and causal reasoning, another line of work does not focus on a single task, but instead investigates the broader bidirectional integration between LLMs and causal learning, or studies other related causal tasks.
	One example is causal attribution \cite{cai2023knowledge,kiciman2023causal,liu2023trustworthy}. These tasks often take the form of questions such as ``why" or ``what is the cause," and aim to identify the key reason behind an outcome or determine whether a factor is a necessary or sufficient cause. Existing studies suggest that, thanks to the human knowledge and common sense encoded in LLMs, they show promise in attribution problems in domains such as law and medicine.
	Another example is causal explanation \cite{bhattacharjee2023llms,gat2023faithful,gao2023chatgpt}. In recent years, many studies have explored the role of LLMs in generating causal explanations. Although debate continues over whether LLMs truly possess causal reasoning ability, a large body of empirical evidence suggests that they can still serve as effective causal explainers, especially because of their ability to organize logic in natural language, generate explanations, and answer ``why" questions \cite{gao2023chatgpt}.
	In addition, several survey papers have examined the relationship between LLMs and causal learning from a broader perspective. Liu \etal \cite{liu2025large} systematically reviews both ``causal inference for LLMs" and ``LLMs for causal inference," summarizing how causal perspectives can improve reasoning, robustness, and interpretability in LLMs, and how LLMs can in turn support causal discovery and causal effect estimation. Wu \etal \cite{wu2024causality} discusses how causality can be integrated into the full LLM lifecycle, including pretraining, fine-tuning, alignment, reasoning, and evaluation. Ma \cite{ma2025causal} further provides a task-level overview of LLM applications in causal discovery, causal effect estimation, and other causal tasks. The main value of these works lies in constructing an overall research landscape, clarifying task boundaries, and organizing methodological developments, thereby offering a unified framework for future method design.}

{Although causal discovery and causal reasoning are related but relatively independent directions, causal discovery often provides an important basis for downstream reasoning in real applications, especially when the true causal graph is unknown. In such cases, tasks such as causal effect estimation, intervention analysis, and some forms of counterfactual reasoning often rely on an identified causal structure  \cite{pearl2009causality}. Based on this consideration, since the task design and evaluation metrics of CausalBN-Bench are mainly centered on structural causal learning (i.e., causal discovery), we focus the comparison on existing LLM-based causal discovery studies. Then, we further clarify the distinctions between CausalBN-Bench and existing LLM-based causal discovery methods.
	Many existing LLM-based causal discovery methods suffer from the following limitations.}
{
	\begin{itemize}
		\item The causal networks used in these works either originate from private datasets or only comprise a limited number of nodes.
		\item  The evaluation tasks lack diversity.
		Most works focus on pairwise causality and full causal graph identification for evaluation while neglecting simpler correlation identification and more complex causal network identification tasks, which could better demonstrate LLMs' ability to grasp causal relationships at different scales and levels of difficulty. 
		\item  The prompt settings are often limited in semantic richness. In most cases, only variable names are used, while background knowledge and training data are either only partially explored or not systematically exploited.
		\item The diversity of the examined LLMs is limited. Only a small number of LLMs are evaluated, potentially undermining the generalizability of the assessments.	
\end{itemize} }

{Compared with these existing studies, CausalBN-Bench offers several advantages. 
	\begin{itemize}
		\item {\bf Diverse scales of datasets from the causal learning community:}
		CausalBN-Bench is an extension of the causal learning research community's efforts, designed to offer a robust evaluation framework. It incorporates 15 commonly used real-world causal learning datasets of diverse scales, enabling rigorous and quantitative measurement of LLMs' causal learning capacities with extensive evaluation results in the causal community as a reference.	
		\item {\bf Evaluation tasks of varying depths and difficulties: }
		CausalBN-Bench offers three tasks of different difficulties, i.e., correlation, causal skeleton, and causality identification, respectively, to holistically assess the causal learning capabilities of existing LLMs. 
		Additionally, we have provided an example on causal structures similar to a particular prompt technique, which identifies the long cause-effect chain to evaluate causal learning abilities on multi-step reasoning, like the popular Chain-of-Thought (CoT) prompt technique.
		\item {\bf Diverse prompts with rich information: }
		CausalBN-Bench offers four distinct prompt formats, encompassing variable name and its combinations with background knowledge and training data, respectively, and the combination of the three. With these diverse prompts, CausalBN-Bench can well assess LLMs' causal learning capacities through looking into their abilities to utilize prior information and comprehend long texts in understanding causal relations.
		\item {\bf Demonstration of the upper limit of LLMs' causal learning capability across various scales and complexities:} CausalBN-Bench evaluates causal relations of varying scales and complexities. CausalBN-Bench covers causal learning datasets with scales ranging from 5 to 109 nodes, far exceeding what current evaluation works have explored. Meanwhile, it evaluates various types of causal structures and discusses different densities in causal learning networks.
		\item 	{\bf Unified Comparison with Traditional Causal Baselines:} CausalBN-Bench includes multiple traditional causal discovery methods as baselines, which supports unified comparisons among LLMs, prompt variants, LLM-based methods, and classical statistical baselines under the same evaluation setting.
	\end{itemize}
}

{Although CausalBN-Bench is primarily designed as a causal discovery benchmark, we also include complementary analyses such as causal strength analysis. Inspired by LLM-based causal reasoning methods, causal strength task provides an initial quantitative view of how LLMs assess causal strength. Additionally, we also evaluate other factors that may influence LLMs’ causal capabilities, including {data identification analysis}, {causal network analysis}, and {prompt robustness analysis}.}

\subsection{Related Work on LLM Reasoning}\label{app:rela_work_reasoning}
{LLM reasoning generally refers to the capability of large language models to perform deliberate, multi-step inference over complex information, so as to support problem solving, decision making, and structured judgment \cite{sun2025survey}. Recent studies \cite{sun2025survey,chen2025towards} show that LLM reasoning has evolved from early single-path linear inference to more advanced paradigms that integrate multi-path search, aggregation, reflection, and planning. Specifically, Chain-of-Thought (CoT) \cite{wei2022chain} decomposes complex tasks into explicit intermediate reasoning steps, enabling LLMs to solve challenging problems in a stepwise manner. Building on this idea, self-consistency  \cite{wang2023self} improves robustness by sampling multiple reasoning paths and aggregating them through majority voting or consistency-based selection, rather than relying on a single reasoning chain. }

{Subsequent methods further extend reasoning from chains to more structured search spaces. Tree-of-Thought \cite{yao2023tree} models reasoning as tree-structured exploration over multiple candidate intermediate states, while Graph-of-Thought \cite{besta2024graph} generalizes this process to graph-structured dependencies among thoughts. In parallel, Reasoning via Planning \cite{hao2023reasoning} treats the LLM as both a reasoning agent and a world model, and combines it with principled planning algorithms such as Monte Carlo Tree Search to explore large reasoning spaces more strategically. Together, these methods represent a clear transition from single-path reasoning to multi-path search and aggregation.}

{Another important direction is reflection- and verification-enhanced reasoning, including methods such as ReAct \cite{yao2023react}, Reflexion \cite{shinn2023reflexion}, SelfCheck \cite{miao2024selfcheck}, and Self-Refine \cite{madaan2023self}. Instead of treating one-pass reasoning as final, these approaches introduce action, feedback, self-checking, and refinement to improve reliability and reduce hallucination. From the perspective of recent survey \cite{chen2025towards}, the development of reasoning methods is no longer only about generating longer outputs, but about strengthening three key capabilities: deep reasoning, extensive exploration, and feasible reflection.}

{Motivated by this development trend, our CoT-analogous causal structure identification task in CausalBN-Bench treats causal structure identification not merely as a one-shot prediction problem, but as a stepwise, decomposable, and verifiable reasoning process. This design is consistent with the broader evolution of LLM reasoning methods \cite{sun2025survey,chen2025towards}, where complex inference is increasingly handled through intermediate reasoning steps, multi-path exploration, and self-reflective verification, rather than direct one-step prediction.}

\subsection{Detailed Comparison Between CausalBN-Bench and Existing Causal Benchmarks and Methods}\label{app:deteld_comparision_des}
{To better position CausalBN-Bench in the growing literature on LLM-based causal learning, we further compare it with several recent representative studies, including CaLM \cite{chen2024causal}, CausalGraphBench \cite{babakov2025causalgraphbench}, CausalGraph2LLM \cite{sheth2025causalgraph2llm}, and two recent LLM-assisted causal learning methods. As summarized in Table~\ref{app:ref_summary}, these studies and CausalBN-Bench address related yet distinct dimensions of LLM-based causal learning. Specifically,}
{
	\begin{itemize}
		\item  CaLM provides a broad benchmark for evaluating causal reasoning in language models. Its main strength lies in wide coverage across causal targets, adaptation settings, evaluation metrics, and error types, and it evaluates a larger number of LLMs than CausalBN-Bench. However, for the causal learning component, CaLM mainly focuses on pairwise causal discovery. In contrast, CausalBN-Bench is designed specifically for learning causal graph structures. It covers a sequence of tasks with increasing difficulty, including correlation identification, causal skeleton identification, and causality identification, and extends further to pairwise causal discovery and whole causal graph discovery. Therefore, the strength of CaLM is broad coverage, while the strength of CausalBN-Bench is a more focused and fine-grained evaluation of structural causal learning.
		\item CausalGraphBench and CausalGraph2LLM focus on different graph-based settings. CausalGraphBench mainly studies end-to-end causal graph discovery, while CausalGraph2LLM focuses on reasoning over encoded causal graphs. Compared with these benchmarks, CausalBN-Bench is more directly aligned with standard causal discovery settings used in the causal discovery community. It is built on commonly used causal network datasets with verified ground-truth graphs, which support structured, reproducible, and diagnostic evaluation using unified structural metrics. In addition, CausalBN-Bench evaluates multiple input settings, including variable names, background knowledge, training data, and their combinations, which provides a broader testbed for analyzing how different forms of information affect LLM-based causal learning.
		\item Li \etal~\cite{li2025can} and Ban \etal~\cite{chen2023mitigating} are not benchmarks in the strict sense, but LLM-assisted causal discovery methods. Their focus is on improving causal discovery performance through intervention design or knowledge integration. By contrast, CausalBN-Bench is intended as a standardized evaluation benchmark rather than a new causal discovery algorithm. For this reason, these studies are better viewed as complementary method-level advances. In contrast, CausalBN-Bench provides a common evaluation framework for comparing such methods and diagnosing their strengths and weaknesses.
\end{itemize}}

{
	Another point concerns graph scale. CausalGraphBench includes several larger causal networks than CausalBN-Bench, mainly because it contains very-large causal networks such as Munin1 and Neuro. We also conducted exploratory tests on very large causal networks, including Pathfinder, Munin1, and Neuro. However, our results show that once the graph size exceeds that of Win95PTS, the causal learning ability of current LLMs drops sharply, especially for open-source models. In addition, even strong closed-source models face clear limits when prompts combine LLM reasoning with training data, due to current constraints on context length and structured-input processing. For this reason, CausalBN-Bench adopts a graph-scale range that better matches the current capability boundary of LLMs while keeping the evaluation valid, reproducible, and comparable. For completeness, we will release the corresponding prompts for Pathfinder, Munin1, and Neuro in our open-source code. However, we do not include these results in CausalBN-Bench because the performance on such very large networks is consistently poor and does not provide a stable or informative comparison.}

{Overall, these comparisons show that CausalBN-Bench is not intended to replace existing studies, but to complement them from a more focused benchmark perspective. Its main advantage lies in its alignment with causal graph structure learning, its hierarchical task design, its support for multiple input settings, and its diagnostic evaluation under unified structural metrics. }
\newpage
\begin{table}[htbp]
	\centering
	\label{app:ref_summary}
	\caption{
		Previous causality identification evaluations for LLMs.}
	\scalebox{0.9}{
		\begin{tabular}{cccc}
			\toprule
			\textbf{Previous Studies} & {\textbf{No. of Nodes}} & {*}{\textbf{Prompt format}} & \textbf{No. of Evaluated LLMs} \\
			\midrule
			\cite{zhiheng2022can} & 2$\sim$10 & Variable name & 1 \\
			\midrule
			\cite{long2023causal} & 8$\sim$37  & Variable name & 2 \\
			\midrule
			\cite{long2023can} & 3$\sim$4   & Variable name & 1 \\
			\midrule
			\cite{pawlowski2023answering} & 5$\sim$40  & Variable name & 2 \\
			\midrule
			\cite{tang2023towards} & NA & Variable name & 2 \\
			\midrule
			\cite{zhang2023understanding} & 2$\sim$20 & Variable name & 2 \\
			\midrule
			\cite{zevcevic2023causal} & 2$\sim$10  & Variable name & 4 \\
			\midrule
			\multirow{1}[3]{*}{\cite{jin2023can}} & \multirow{1}[3]{*}{2$\sim$6}   & \multirow{1}[3]{*}{Variable name} &3 series  \\
			&   &  & (9 LLMs) \\
			\midrule
			\cite{tu2023causal}  & 2$\sim$10  & Variable name & 3 \\
			\midrule
			\cite{zhang2024causal} & 11    & Variable name &  1  \\
			\midrule
			\cite{jiralerspong2024efficient} & 8$\sim$20  & Variable name &  1 \\
			\midrule
			\cite{antonucci2023zero}  & 2$\sim$20 & Variable name & 1 \\
			\midrule
			\cite{gao2023chatgpt}  & NA & Variable name & 4 \\
			\midrule
			\cite{jin2023cladder}  & 3$\sim$4 & Variable name & 8 \\
			\midrule
			\multirow{2}{*}{\cite{11141551}} & \multirow{2}{*}{2$\sim$25} & Variable name        & \multirow{2}{*}{9} \\
			&                             & Background Knowledge &                     \\
			\midrule
			\multirow{2}{*}{\cite{liu2024llms}} & \multirow{2}{*}{NA} & Variable name        & \multirow{2}{*}{1} \\
			&                             & Background Knowledge &                     \\
			\midrule
			\cite{tu2024carl}  & 10$\sim$51 & Variable name & 5 \\
			\midrule
			\cite{wang2024causalbench}  & NA & Variable name & 10 \\
			\midrule
			\cite{chen2024causal}  & 9 & Variable name & 28 \\
			\midrule
			\multirow{2}{*}{\cite{babakov2025causalgraphbench}} & \multirow{2}{*}{5$\sim$222} & Variable name        & \multirow{2}{*}{3} \\
			&                             & Background Knowledge &                     \\
			\midrule
			\multirow{2}{*}{\cite{sheth2025causalgraph2llm}} & \multirow{2}{*}{20$\sim$37} & Variable name        & \multirow{2}{*}{6} \\
			&                             & Background Knowledge &                     \\
			\midrule

			\multirow{2}{*}{\cite{li2025can}} & \multirow{2}{*}{37} & Variable name        & \multirow{2}{*}{1} \\
			&                             & Background Knowledge &                     \\
			\midrule
			
			\multirow{2}{*}{\cite{chen2023mitigating}} & \multirow{2}{*}{48} & Variable name        & \multirow{2}{*}{1} \\
			&                             & Background Knowledge &                     \\
			\midrule
			\multirow{1.5}[3]{*}{CausalBN-Bench} 		 &\multirow{1.5}[3]{*}{2$\sim$109} &\multirow{1.5}[3]{*}{Four types \textsuperscript{*}}   &6 series  \\ 
			& &    & (19 LLMs) \\
			\bottomrule			
		\end{tabular}%
	}
	
\end{table}	{*} `Four types' refers to variable name, variable name +  training data, variable name + background knowledge, and variable name +  training data + background knowledge.

\clearpage
\section{Detailed Experimental Settings} \label{app:exp_setup}
\subsection{Detailed Definitions of Evaluation Metrics, Including Network Sparsity, In-/Out-Degree, and Maximal Inference Length}\label{app:detailed_metrics}
{In this subsection, we clarify the meanings of network sparsity, node in-degree and out-degree in the causal network. Let $G=(V,E)$ denote a directed causal network with $n=|V|$ nodes. Let $A \in \{0,1\}^{n \times n}$ be its adjacency matrix, where $A_{i,j}=1$ indicates a directed edge $X_i \rightarrow X_j$ and $A_{i,j}=0$ otherwise, with $A_{i,i}=0$. Then, the evaluation metrics are defined as follows:}

{
	\begin{itemize}
		\item	\textbf{Network sparsity} describes how many directed edges are absent relative to a fully connected directed network on the same node set \cite{pearl2009causality}, which can be defined as
		\begin{equation}
			\mathrm{sparsity}(G) \;=\; 1 - \frac{|E|}{n(n-1)}, 
			\quad \text{where} \quad 
			|E|=\sum_{i\neq j} A_{i, j}.
		\end{equation}
		A higher sparsity value indicates a sparser learned causal structure.
		\item	\textbf{In-degree and out-degree}  measure node-level connectivity. The in-degree of $X_j$ counts how many edges point into $X_j$ and thus how many direct causes it has \cite{pearl2009causality, peters2017elements}, which can be defined as 
		\begin{equation}
			\deg_{\mathrm{in}}(X_j) \;=\; \sum_{i\neq j} A_{i,j}.
		\end{equation}
		The out-degree of $X_i$ counts how many edges point out of $X_i$ and thus how many direct effects it has, which can be defined as 
		\begin{equation}
			\deg_{\mathrm{out}}(X_i) \;=\; \sum_{j\neq i} A_{ij}.
		\end{equation}
		\item \textbf{Maximal inference length} measures the longest minimal directed path length between any reachable ordered pair of variables in the predicted causal network. A larger maximal inference length indicates that the causal network contains longer directed causal chains.
	\end{itemize}
}

\subsection{Evaluated LLMs} \label{app:LLMs_v1}
{In order to evaluate causal learning ability gaps among different LLMs, we utilize five series of open-source LLMs:} 
\begin{itemize}
	\item BERT series: BERT-large \cite{lewis2020bart}, RoBERTa-large  \cite{liu2019roberta}, DeBERTa-large \cite{he2021deberta}, and DistilBERT-mnli \cite{shleifer2020pre};
	\item LLAMA series \cite{touvron2023llama}: LLAMA-7B, LLAMA-13B and LLAMA-33B;
	\item OPT series \cite{zhang2022opt}: OPT-1D3B, OPT-2D7B, OPT-6D7B and OPT-66B;
	\item InternLM series \cite{team2023internlm}: InternLM-7B and InternLM-20B;
	\item Falcon series \cite{falcon40b}: Falcon-7B and Falcon-40B;
\end{itemize}
along with one series of closed-source LLMs \cite{achiam2023gpt}:
\begin{itemize}
	\item GPT3.5-Turbo;
	\item GPT4;
	\item GPT4-Turbo;
\end{itemize}
in zero-shot scenarios. These LLMs are very recently and widely used by researchers, and more details of these LLMs are provided in Table \ref{tab:LLM}.

\begin{table}[h]
	\centering
	\caption{Main used LLMs in CausalBN-Bench}
	\label{tab:LLM}
	\begin{tabular}{@{}cccc@{}}
		\toprule
		LLMs & Name & Parameters &Layers  \\ \midrule
		
		\multirow{6}{*}{BERT} & BERT-large & 336M &24   \\ \cmidrule(l){2-4}
		& RoBERTa-large & 350M &24  \\ \cmidrule(l){2-4}
		& DeBERTa-v3-large & 506M &24   \\ \cmidrule(l){2-4}
		& DistilBERT-mnli & 110M &24   \\ 
		\midrule
		\multirow{4}{*}{LLAMA} & LLAMA-7B & 6.7 Billion &32  \\ \cmidrule(l){2-4} 
		& LLAMA-13B & 13 Billion &40 \\ \cmidrule(l){2-4} 
		& LLAMA-33B & 32.5 Billion &60 \\ 
		
		\midrule
		
		\multirow{6}{*}{OPT} & OPT-1D3B & 1.3 Billion &24  \\ \cmidrule(l){2-4} 
		& OPT-2D7B & 2.7 Billion &32  \\ \cmidrule(l){2-4}
		& OPT-6D7B & 6.7 Billion &32  \\ \cmidrule(l){2-4}
		& OPT-66B & 66 Billion &72  \\ 			
		\midrule
		\multirow{3}{*}{InternLM} & InternLM-7B & 7 Billion &32  \\ \cmidrule(l){2-4}
		& InternLM-20B & 20 Billion &60   \\ 
		\midrule
		\multirow{3}{*}{Falcon} & Falcon-7B & 7 Billion &32  \\ \cmidrule(l){2-4}			
		& Falcon-40B & 40 Billion &60  \\ 
		\midrule
		\multirow{6}{*}{ChatGPT} & GPT2 & 124M & 12\\ \cmidrule(l){2-4}
		& GPT3.5-Turbo & 175 Billion & Unknown \\ \cmidrule(l){2-4}
		& GPT4 & 1800 Billion & Unknown  \\ \cmidrule(l){2-4}
		& GPT4-Turbo & 1800 Billion & Unknown  \\

		\bottomrule
	\end{tabular}
\end{table}

\subsection{Additional Experimental Configurations}\label{app:add_exp_set}
{The main results reported in Tables II–V are obtained under a single-shot prompting setting with deterministic decoding. Specifically, we use a fixed prompt template, set $temperature = 0$ and $top-p = 1$, limit the maximum output length to 512 tokens, and apply a unified parsing rule to map each model output to the binary decision ``yes" or ``no". For causality identification evaluation tasks, both $X \rightarrow Y$ and $Y \rightarrow X$ are queried and evaluated independently. To verify reproducibility, we further repeat the full evaluation pipeline 10 times. Since the prompt template and decoding configuration are deterministic, repeated runs yield identical results to a single run. Therefore, the reported F1 score and accuracy remain numerically unchanged across repeated evaluations.}

\subsection{Detailed Generation of Ground-Truth Labels for Direct and Indirect Correlation} \label{app:lable_gt_direct}
{
	This subsection specifies how the ground-truth labels are generated for the direct and indirect correlation identification tasks. All labels are derived deterministically from the ground-truth causal network, represented as a DAG, and do not rely on empirical data samples or statistical testing. Specifically, let $G=(V, E)$ denote the ground-truth DAG, where $V=\{X_1,\dots, X_n\}$ is the set of variables and $E$ is the set of directed edges.}

{\textbf{Direct correlation label}~
	For each unordered variable pair $\{X_i,X_j\}$, we define the direct correlation label $y_{\mathrm{direct}}(X_i,X_j)$ as
	\begin{align}
		y_{\mathrm{direct}}(X_i,X_j)=
		\begin{cases}
			1, & \text{if } (X_i \rightarrow X_j)\in E \text{ or } (X_j \rightarrow X_i)\in E,\\
			0, & \text{otherwise.}
		\end{cases}
	\end{align}
	Thus, $y_{\mathrm{direct}}(X_i, X_j)=1$ if and only if $X_i$ and $X_j$ are adjacent in the ground-truth DAG, which is equivalent to adjacency in the DAG skeleton.}

{\textbf{Indirect correlation label}~
	For each unordered variable pair $\{X_i,X_j\}$, we define the indirect-correlation label $y_{\mathrm{indirect}}(X_i,X_j)$ as
	\begin{align}
		y_{\mathrm{indirect}}(X_i,X_j)=
		\begin{cases}
			1, & \text{if } X_i \text{ and } X_j \text{ are d-connected in } G \text{ under the empty conditioning set,}\\
			& \text{and } y_{\mathrm{direct}}(i,j)=0,\\
			0, & \text{otherwise.}
		\end{cases}
	\end{align}
	That is, $y_{\mathrm{indirect}}(X_i, X_j)=1$ if $X_i$ and $X_j$ are marginally dependent according to the ground-truth DAG semantics, while excluding directly adjacent pairs. This label captures dependence induced through paths via intermediate nodes.}

{To reproduce the labels, one can enumerate all unordered pairs $\{X_i, X_j\}$, compute adjacency in $E$ for $y_{\mathrm{direct}}$, and compute d-connection under the empty conditioning set for $y_{\mathrm{indirect}}$ while excluding adjacent pairs. No numerical correlation computation and no conditional independence testing on raw samples are required in this label generation procedure.}

\clearpage

\section{Prompts for Different Evaluation Tasks}\label{app:prompt_task}
\subsection{Correlation}\label{app:correlation}

\begin{figure}[htb] 
	\center{\includegraphics[width=0.5\linewidth]  {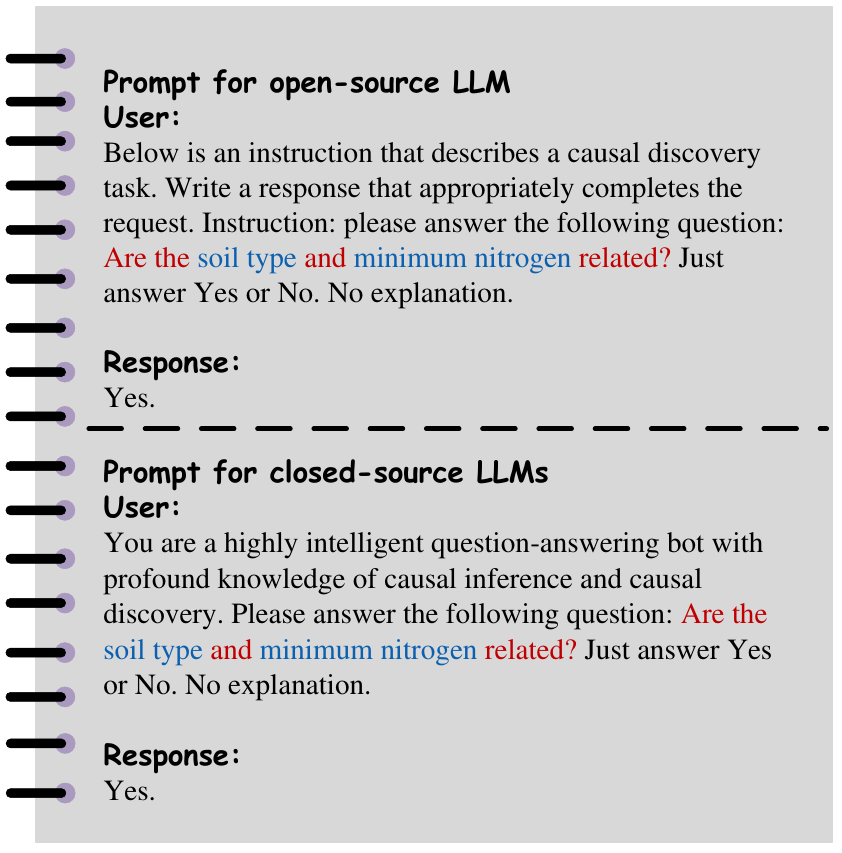}}\caption{Prompt for open-source and closed-source LLMs in the correlation task.} 
	\label{fig:correlation} 
\end{figure}

The prompt related to correlation evaluation is shown in Fig. \ref{fig:correlation}, where the sentence structure in red and the variable names in blue can be varied. We take the Variable A (\textit{Var. A}) and Variable B (\textit{Var. B}) as an example, and the red sentence structures include the following five categories:

\begin{itemize}
	\item Are \textit{Var. A} and \textit{Var. B} related?
	\item Are \textit{Var. A} and \textit{Var. B} correlated?
	\item Is there a correlation between \textit{Var. A} and \textit{Var. B}? 
	\item Is there a relation between \textit{Var. A} and \textit{Var. B}?
	\item Do \textit{Var. A} and \textit{Var. B} have a connection? 
\end{itemize}
Meanwhile, the blue variable names include two categories (taking the variable "Asia" as an example).
\begin{itemize}
	\item Asia
	\item Visiting to Asia
\end{itemize}

\clearpage
\subsection{Causal Skeleton}\label{app:causal_skeleton}
\begin{figure}[htb] 
	\center{\includegraphics[width=0.5\linewidth]  {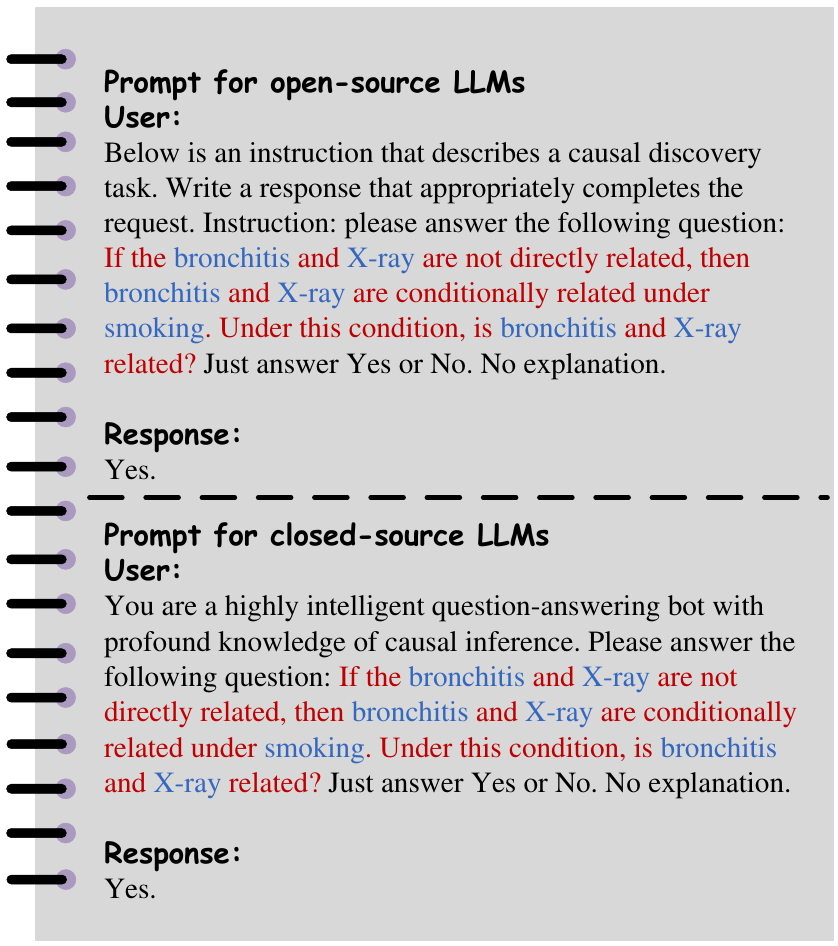}}\caption{Prompt for LLMs in the causal skeleton evaluation task.} 
	\label{fig:causal_skeleton} 
\end{figure}

The prompt for the causal skeleton is shown in Fig. \ref{fig:causal_skeleton}, where the sentence structure in red and the variable names in blue can be varied. Based on conditional correlations calculated using classical methods of causal learning, we can divide the causal skeleton into the following three types of sentence structures:
\begin{itemize}
	\item Are \textit{Var. A} and \textit{Var. B} directly related?
	\item If \textit{Var. A} and \textit{Var. B} are not directly related, then \textit{Var. A} and \textit{Var. B} are conditionally related under \textit{Var. C}. Under this condition, are \textit{Var. A} and \textit{Var. B} related? 
	\item If \textit{Var. A} and \textit{Var. B} are not directly related, then \textit{Var. A} and \textit{Var. B} are not conditionally related under \textit{Var. C}. Under this condition, are \textit{Var. A} and \textit{Var. B} related? 
\end{itemize}
Meanwhile, the blue variable names include two categories (Taking the variable ``Asia" as an example).
\begin{itemize}
	\item Asia
	\item Visiting to Asia
\end{itemize}

\clearpage
\subsection{Causality}\label{app:causal}

\begin{figure}[ht]
	\begin{center}
		\subfigure[Type I]{\includegraphics[width=8cm]{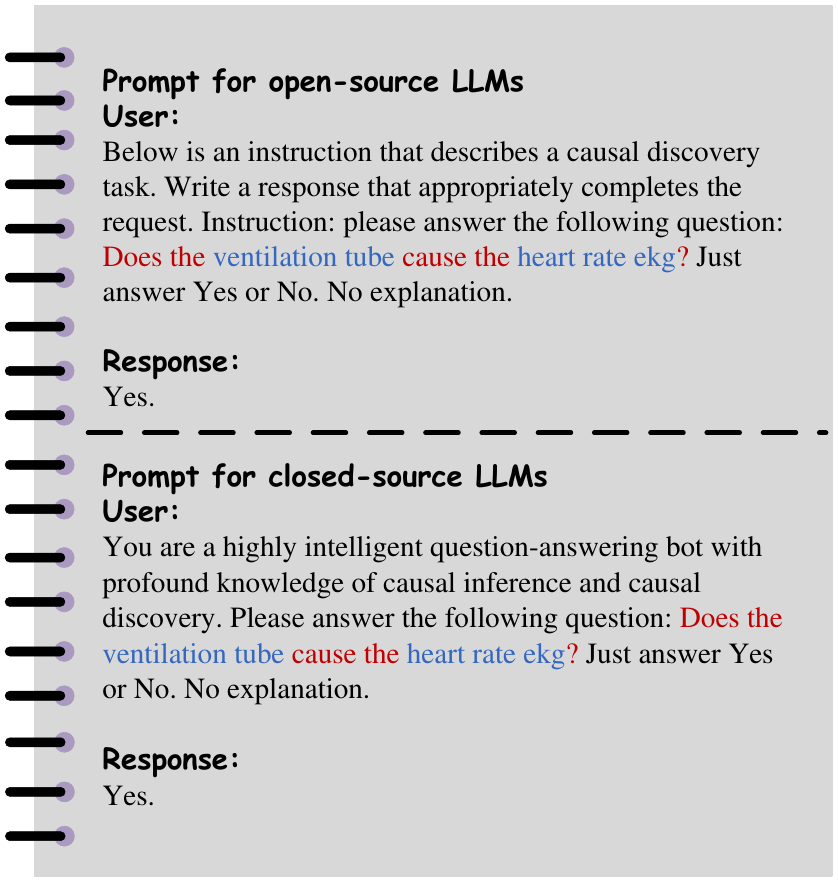}}
		\subfigure[Type II]{\includegraphics[width=8.55 cm]{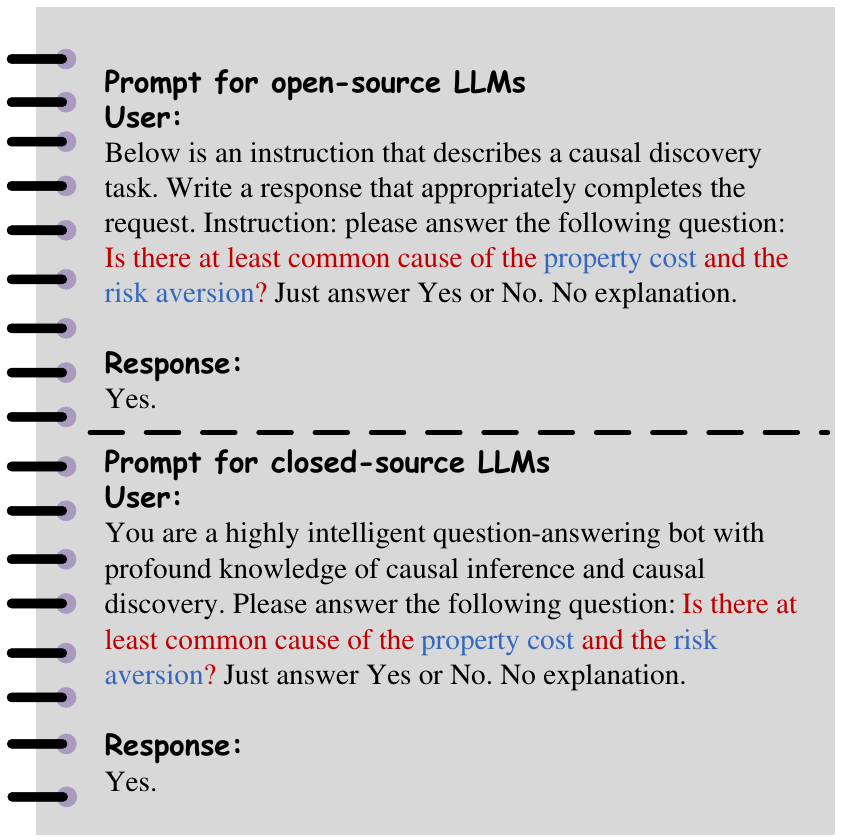}}
		\caption{Prompt for LLMs in the causality evaluation task.}
		\label{fig:Causality}
	\end{center}
\end{figure}

The causality prompt is provided in Fig. \ref{fig:Causality}, which includes two types of prompts. The first type of prompt (Type I) is set up similarly to correlation experiments, aiming to explore whether there is a causal relationship between pairs of variables. It mainly includes the following five sentence structures:
\begin{itemize}
	\item Are \textit{Var. A} and \textit{Var. B} causally related?
	\item Is there a causal connection between \textit{Var. A} and \textit{Var. B}?
	\item Does \textit{Var. A} cause \textit{Var. B}? 
	\item Does \textit{Var. A} influence \textit{Var. B}?
	\item Is there causality between \textit{Var. A} and \textit{Var. B}?
\end{itemize}
The second type of prompt (Type II) focuses more on the specific structure of the causal relationships between pairs of variables, such as collider, Chain, etc. It mainly includes the following six sentence structures:
\begin{itemize}
	\item Does \textit{Var. A} directly cause the \textit{Var. B}?
	\item Does \textit{Var. A} cause something else which causes \textit{Var. B}?
	\item Is there at least one common effect of \textit{Var. A} and the \textit{Var. B}?
	\item Is the \textit{Var. B} a cause for the \textit{Var. A} but not a direct one?
	\item Is there at least one common effect of the \textit{Var. A} and the \textit{Var. B}?
	\item Is there at least one common cause of the \textit{Var. A} and the \textit{Var. B}?
\end{itemize}

\newpage
\subsection{CoT-Analogous Causal Structure Identification}\label{app:cot}
\begin{figure}[ht] 
	\center{\includegraphics[width=0.5\linewidth]  {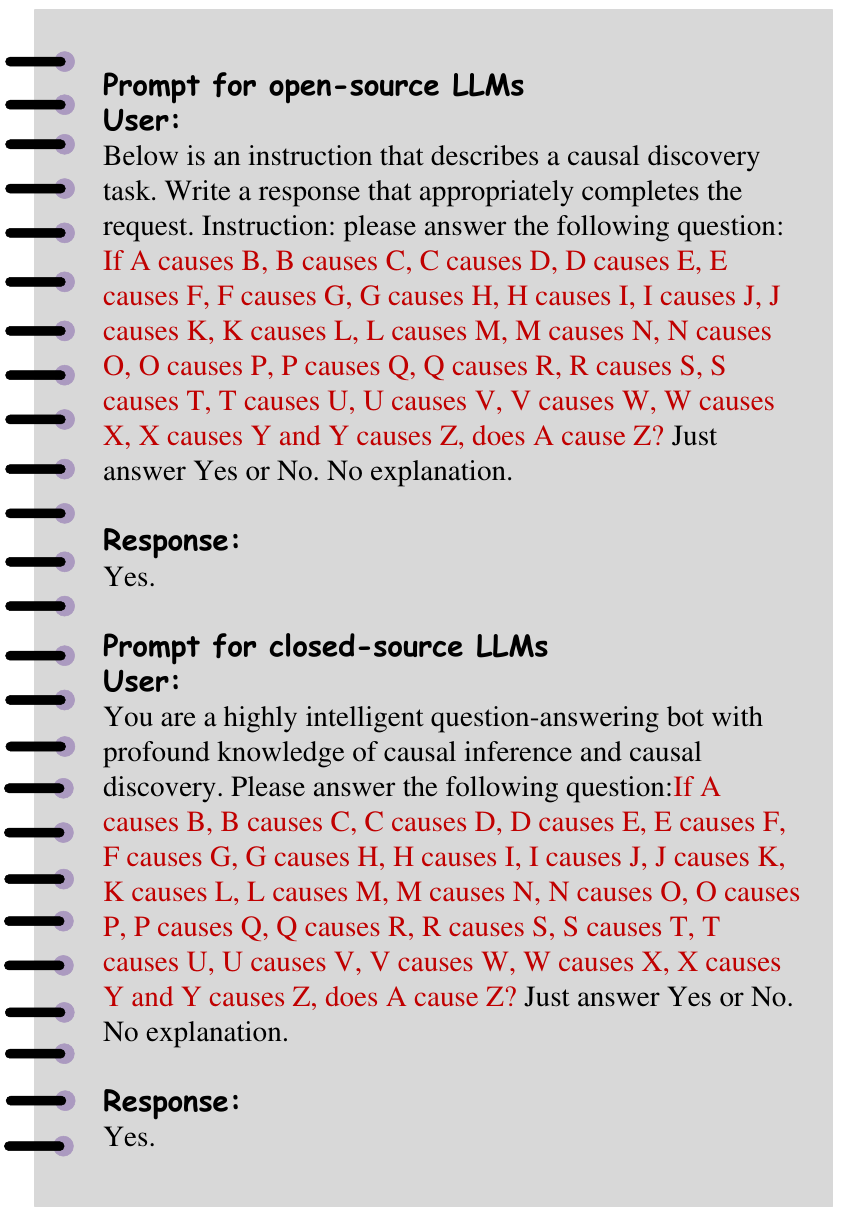}}\caption{Prompt for LLMs with CoT-analogous causal structure identification.} 
	\label{fig:CoT} 
\end{figure}

The prompt for CoT-analogous causal structure identification is shown in Fig. \ref{fig:CoT}, where the prompt in the red part is passed from A to B in sequence. Therefore, the order of this prompt is as follows:
\begin{itemize}
	\item If A causes B and B causes C, does A cause C? 
	\item If A causes B, B causes C and C causes D, does A cause D? 
	\item If A causes B, B causes C, C causes D and D causes E, does A cause E? 
	\item If A causes B, B causes C, C causes D, D causes E and E causes F, does A cause F? 
	\item If A causes B, B causes C, C causes D, D causes E, E causes F and F causes G, does A cause G? 
	\item If A causes B, B causes C, C causes D, D causes E, E causes F, F causes G and G causes H, does A cause H? 
	\item If A causes B, B causes C, C causes D, D causes E, E causes F, F causes G, G causes H and H causes I, does A cause I?
	\item If A causes B, B causes C, C causes D, D causes E, E causes F, F causes G, G causes H, H causes I and I causes J, does A cause J?
	\item If A causes B, B causes C, C causes D, D causes E, E causes F, F causes G, G causes H, H causes I, I causes J and J causes K, does A cause K?
	\item If A causes B, B causes C, C causes D, D causes E, E causes F, F causes G, G causes H, H causes I, I causes J, J causes K and K causes L, does A cause L?
	\item If A causes B, B causes C, C causes D, D causes E, E causes F, F causes G, G causes H, H causes I, I causes J, J causes K, K causes L and L causes M, does A cause M?
	\item If A causes B, B causes C, C causes D, D causes E, E causes F, F causes G, G causes H, H causes I, I causes J, J causes K, K causes L, L causes M and M causes N, does A cause N?
	\item If A causes B, B causes C, C causes D, D causes E, E causes F, F causes G, G causes H, H causes I, I causes J, J causes K, K causes L, L causes M, M causes N and N causes O, does A cause O?
	\item If A causes B, B causes C, C causes D, D causes E, E causes F, F causes G, G causes H, H causes I, I causes J, J causes K, K causes L, L causes M, M causes N, N causes O and O causes P, does A cause P?
	\item If A causes B, B causes C, C causes D, D causes E, E causes F, F causes G, G causes H, H causes I, I causes J, J causes K, K causes L, L causes M, M causes N, N causes O, O causes P and P causes Q, does A cause Q?
	\item If A causes B, B causes C, C causes D, D causes E, E causes F, F causes G, G causes H, H causes I, I causes J, J causes K, K causes L, L causes M, M causes N, N causes O, O causes P, P causes Q and Q causes R, does A cause R?
	\item If A causes B, B causes C, C causes D, D causes E, E causes F, F causes G, G causes H, H causes I, I causes J, J causes K, K causes L, L causes M, M causes N, N causes O, O causes P, P causes Q, Q causes R and R causes S, does A cause S?
	\item If A causes B, B causes C, C causes D, D causes E, E causes F, F causes G, G causes H, H causes I, I causes J, J causes K, K causes L, L causes M, M causes N, N causes O, O causes P, P causes Q, Q causes R, R causes S and S causes T, does A cause T?
	\item If A causes B, B causes C, C causes D, D causes E, E causes F, F causes G, G causes H, H causes I, I causes J, J causes K, K causes L, L causes M, M causes N, N causes O, O causes P, P causes Q, Q causes R, R causes S, S causes T and T causes U, does A cause U?
	\item If A causes B, B causes C, C causes D, D causes E, E causes F, F causes G, G causes H, H causes I, I causes J, J causes K, K causes L, L causes M, M causes N, N causes O, O causes P, P causes Q, Q causes R, R causes S, S causes T, T causes U and U causes V, does A cause V?	\item If A causes B, B causes C, C causes D, D causes E, E causes F, F causes G, G causes H, H causes I, I causes J, J causes K, K causes L, L causes M, M causes N, N causes O, O causes P, P causes Q, Q causes R, R causes S, S causes T, T causes U, U causes V and V causes W, does A cause W?
	\item If A causes B, B causes C, C causes D, D causes E, E causes F, F causes G, G causes H, H causes I, I causes J, J causes K, K causes L, L causes M, M causes N, N causes O, O causes P, P causes Q, Q causes R, R causes S, S causes T, T causes U, U causes V, V causes W and W causes X, does A cause X?
	\item If A causes B, B causes C, C causes D, D causes E, E causes F, F causes G, G causes H, H causes I, I causes J, J causes K, K causes L, L causes M, M causes N, N causes O, O causes P, P causes Q, Q causes R, R causes S, S causes T, T causes U, U causes V, V causes W, W causes X and X causes Y, does A cause Y? 
	\item If A causes B, B causes C, C causes D, D causes E, E causes F, F causes G, G causes H, H causes I, I causes J, J causes K, K causes L, L causes M, M causes N, N causes O, O causes P, P causes Q, Q causes R, R causes S, S causes T, T causes U, U causes V, V causes W, W causes X, X causes Y and Y causes Z, does A cause Z?
	
\end{itemize}

\clearpage
\section{Prompts for Different Input Formats}\label{app:prompt_format}
\subsection{Variable Name}\label{app:var}
The prompt based on the variable name is the same as the prompt in the causality evaluation task {which is shown in} Fig. \ref{fig:Causality}.

\subsection{Variable Name + Background Knowledge}\label{app:var_bk}
\begin{figure}[ht]
	\begin{center}
		\subfigure[Open-Source LLMs]{\includegraphics[width=8cm]{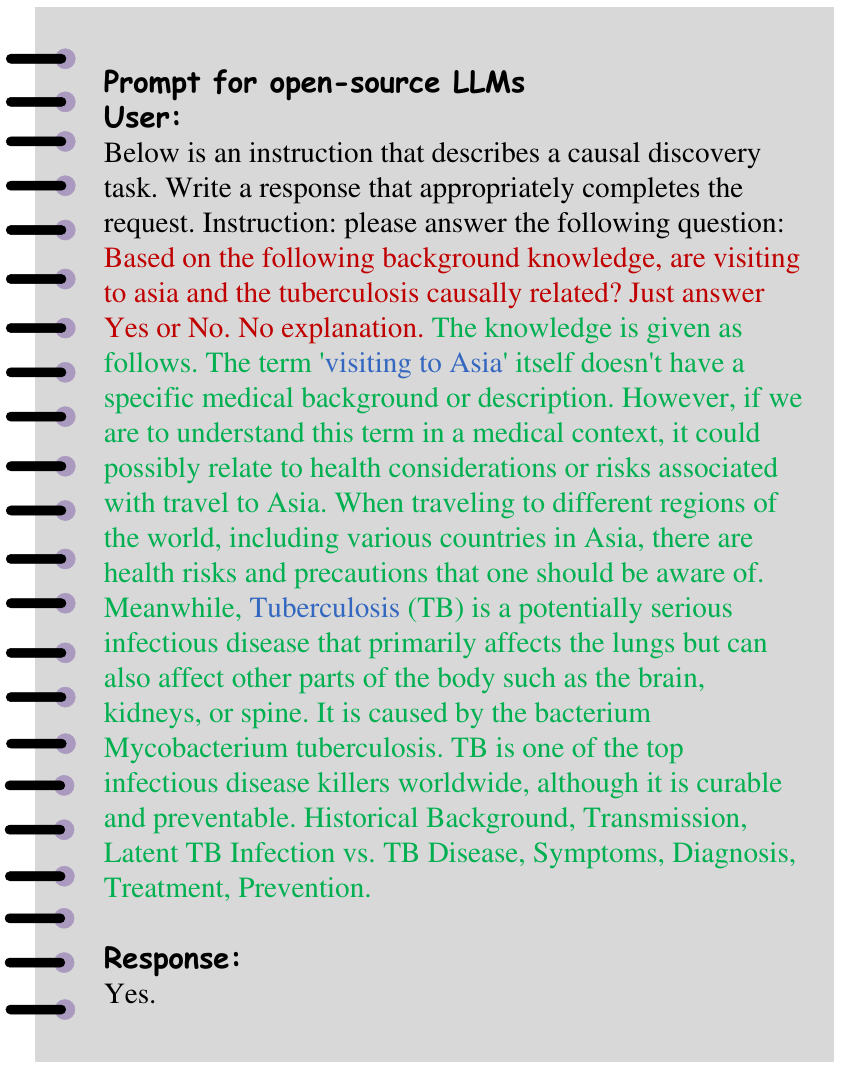}}
		\subfigure[Closed-Source LLMs]{\includegraphics[width=8 cm]{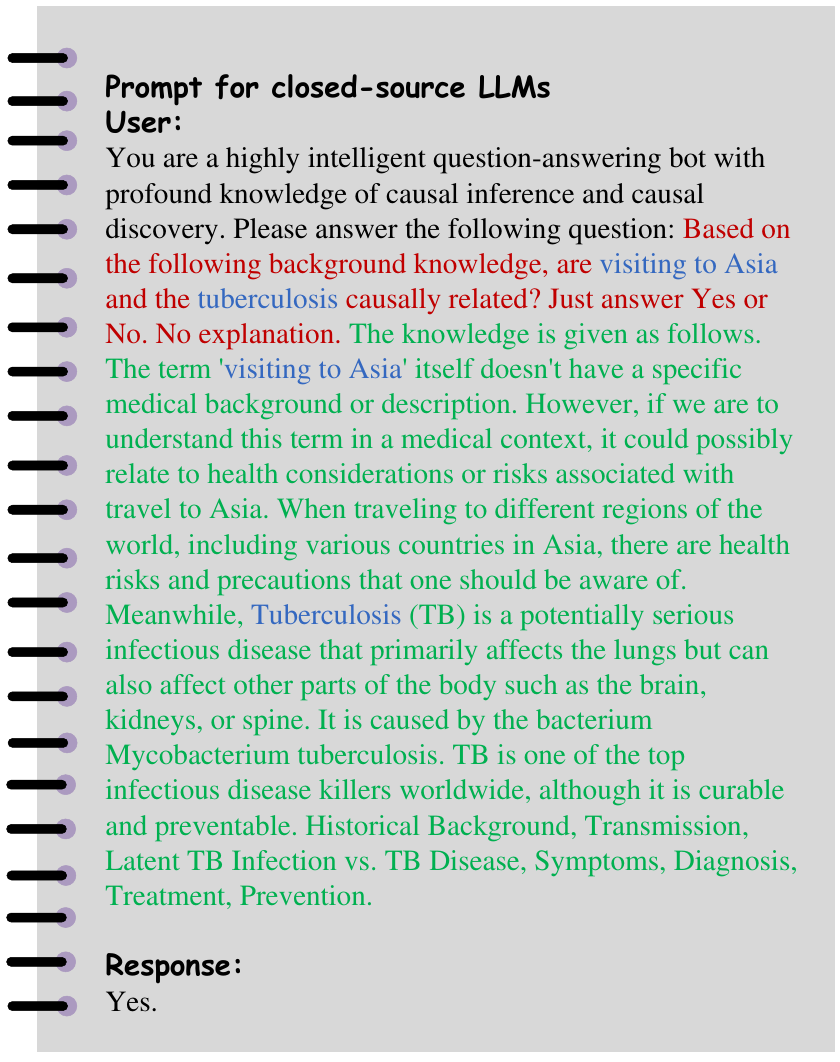}}
		\caption{Prompt format ``variable name + background knowledge.''}
		\label{fig:Causality+Know}
	\end{center}
\end{figure}

The prompt that combines variable names and background knowledge input is shown in Fig. \ref{fig:Causality+Know}, where the red part is the basic interchangeable task sentence structure, the blue part is the variable names, and the green part is the background knowledge. In the green part, the background knowledge of two variables is described in sequence.

\clearpage
\subsection{Variable Name + Training Data}\label{app:var_sd}
\begin{figure}[!h] 
	\center{\includegraphics[width=1\linewidth]  {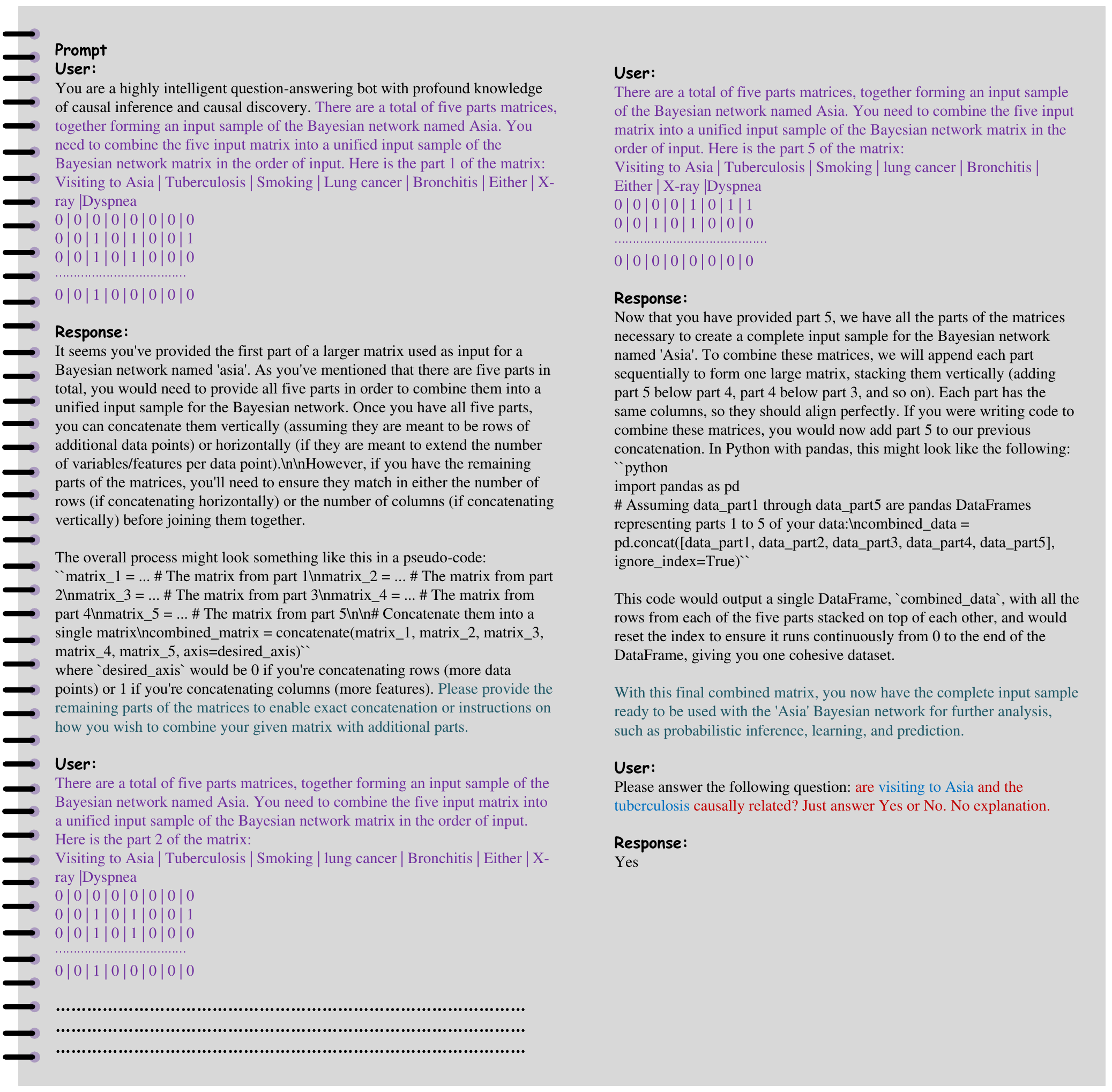}}\caption{Prompt format ``variable name + training data.''} 
	\label{fig:Causality+data} 
\end{figure}

The prompt that combines variable names and training data input is provided in Fig. \ref{fig:Causality+data}, where the purple part emphasizes how users can split 1 $\times$ 500 observed sample data into 5 $\times$ 100 sample data, and then through multiple rounds of interaction with GPT, complete the process. The dark green part is the feedback from GPT after receiving our data information. After completing the reception and combination of matrices, GPT will ask the user for the next command, such as inference, learning, and prediction. Finally, as shown in the red part, we carry out related evaluation tasks.

\clearpage
\subsection{Variable Name + Background Knowledge + Training Data}\label{app:var_bk_sd}
\begin{figure}[!h] 
	\center{\includegraphics[width=1\linewidth]  {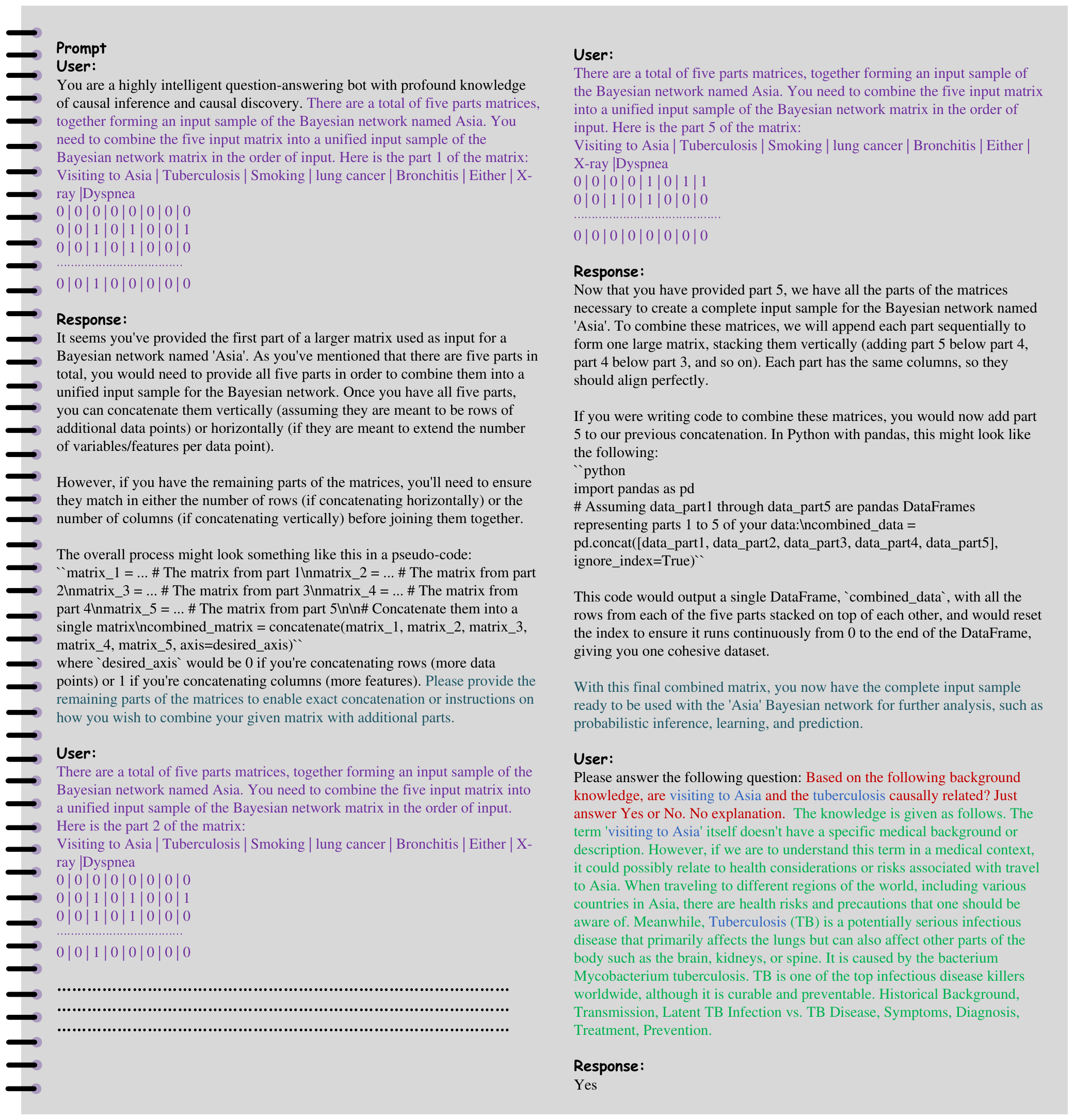}}\caption{Prompt format ``variable name + background knowledge + training data.''} 
	\label{fig:Causality+data+knowledge} 
\end{figure}

The prompt that combines variable names, background knowledge, and training data input is shown in Fig. \ref{fig:Causality+data+knowledge}, where the purple part is the same as in Fig. \ref{fig:Causality+data}, emphasizes how users can split 1 $\times$ 500 sample data into 5 $\times$ 100 sample data, and then through multiple rounds of interaction with GPT, implement the process. The light green part is the feedback from GPT after receiving our data information. After completing the reception and combination of matrices, GPT will prompt the user for the next instruction, such as inference, learning, and prediction. Finally, as shown in the red part, we carry out related Causality evaluation tasks. The green part describes the background knowledge of two variables in sequence.

\clearpage
\section{Detailed Evaluation Results for Different Evaluation Tasks}\label{app:add_exp_evaluation}

\subsection{Detailed Evaluation Results for the Direct Correlation Identification Task}\label{app:dir_relat}
\begin{figure}[htb] 
	\center{\includegraphics[width=0.9\linewidth]  {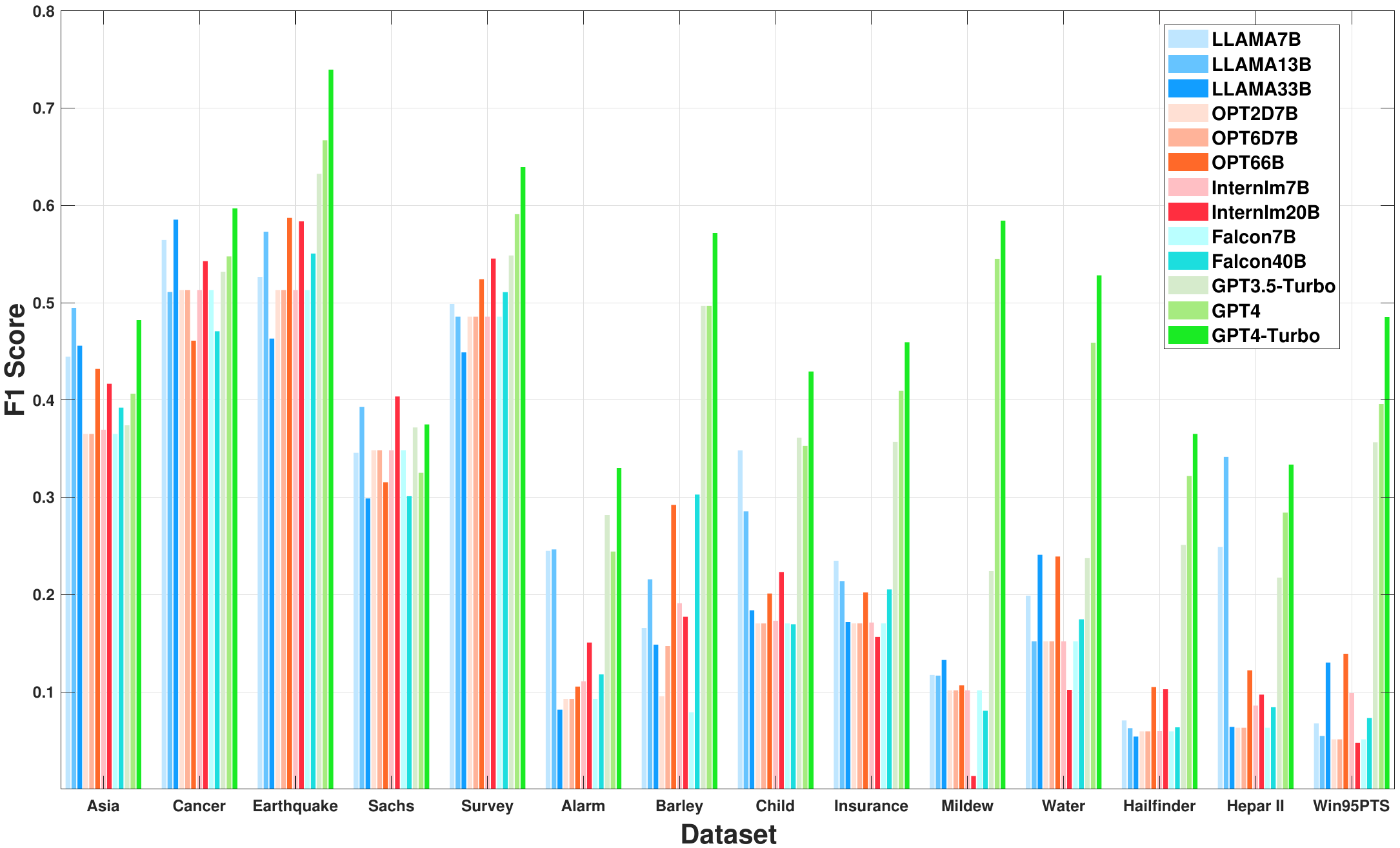}}\caption{F1 scores of LLMs on direct correlation identification.} 
	\label{exp:dir_relat} 
\end{figure}

We also illustrate the performance of LLMs on identifying direct correlation in Fig. \ref{exp:dir_relat}. 
We can see closed-source LLMs maintain higher F1 scores and accuracies compared to open-source LLMs. Among the open-source LLMs, LLAMA series LLMs generally outperform their competitors, i.e. OPT series, InternLM series and Falcon series, followed by InternLM series. 
It is noteworthy that some open-source LLMs demonstrate superior performance on certain causality discovery datasets. For instance, InternLM-20B on Earthquake and Cancer datasets, LLAMA-33B on Water dataset, and Falcon40B on Barley dataset, all achieve performance that substantially exceeds that of other open-source LLMs and approaches that of closed-source LLMs. This variation may be attributed to these models having absorbed more knowledge relevant or analogous to specific datasets during the pre-training phase. For example, since Cancer dataset is from the medical domain, it is plausible that during the pre-training of InternLM-20B and OPT-66B, medical samples related to cancer have been included, endowing them with a superior ability to recognize direct correlations on Cancer dataset.

\subsection{Detailed Evaluation Results for the Indirect Correlation Identification Task}\label{app:indir_relat}
\begin{figure}[htb] 
	
	\center{\includegraphics[width=0.9\linewidth]  {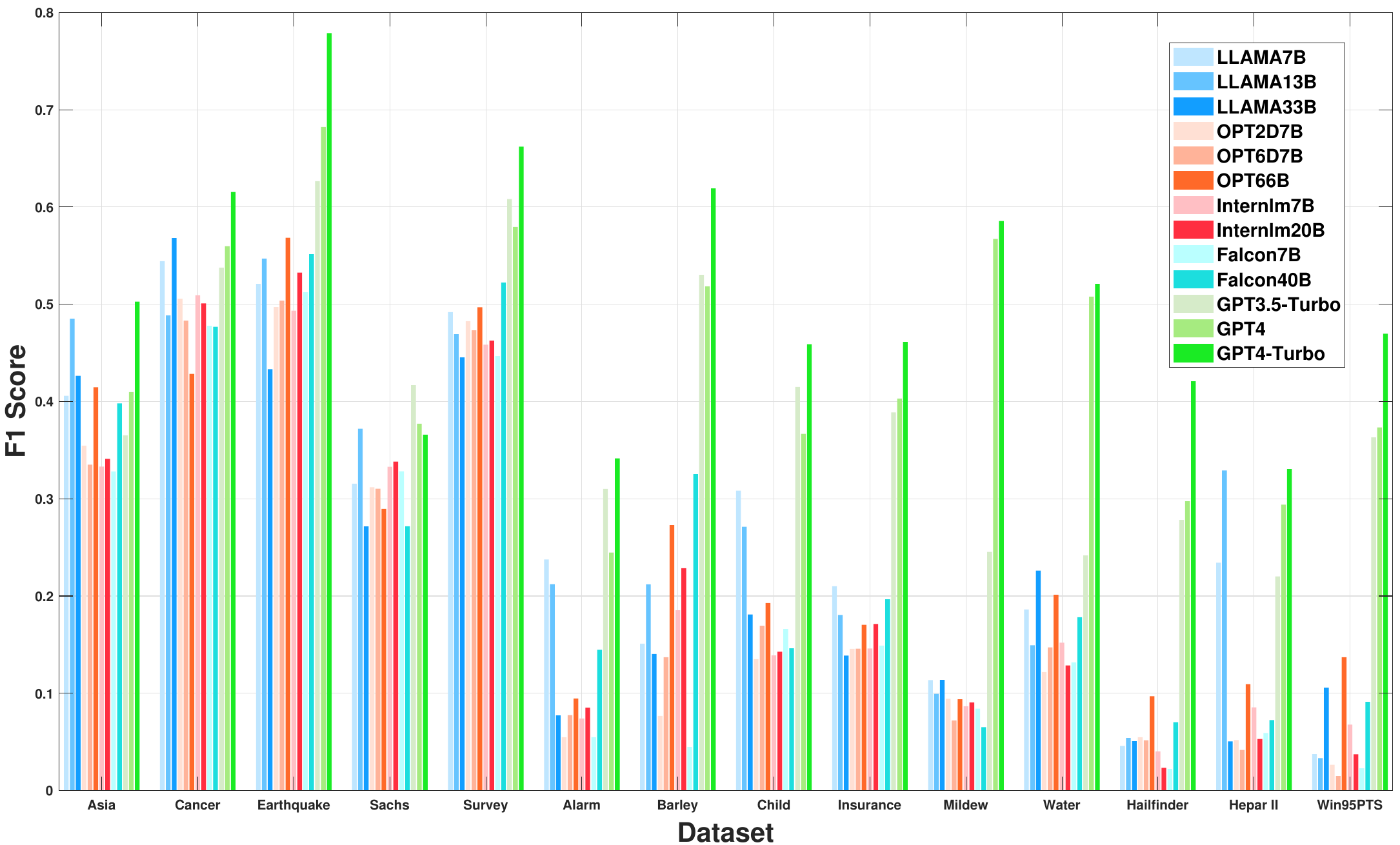}}\caption{F1 scores of LLMs on indirect correlation identification.} 
	\label{exp:indir_relat} 
\end{figure}

We also illustrate the results in Fig. \ref{exp:indir_relat} to examine the performance of the LLMs.
We find that closed-source LLMs outperform other open-source models, achieving the highest F1 scores and accuracies across all causal datasets. 
Especially, on large scale datasets, they exhibit capabilities far surpassing those of open-source models. 
As for open-source models, LLAMA series LLMs show superior performance on small and medium scale causal datasets compared to others, but on large scale datasets, OPT series LLMs exhibit the best performance.
This may be attributed to the inherently robust capabilities of LLAMA series, with the largest being LLAMA-33B, while the superior performance of OPT series, primarily OPT-66B, is likely due to its much larger parameter size compared to LLAMA series LLMs. Additionally, the performance differences between InternLM series and Falcon series LLMs are not substantial.

\newpage
\subsection{Detailed Evaluation Results for the Causal Skeleton Identification Task}\label{app:add_ske}
\begin{figure}[ht]
	\centering
	\subfigure[F1 score]{\includegraphics[width=0.45\linewidth]{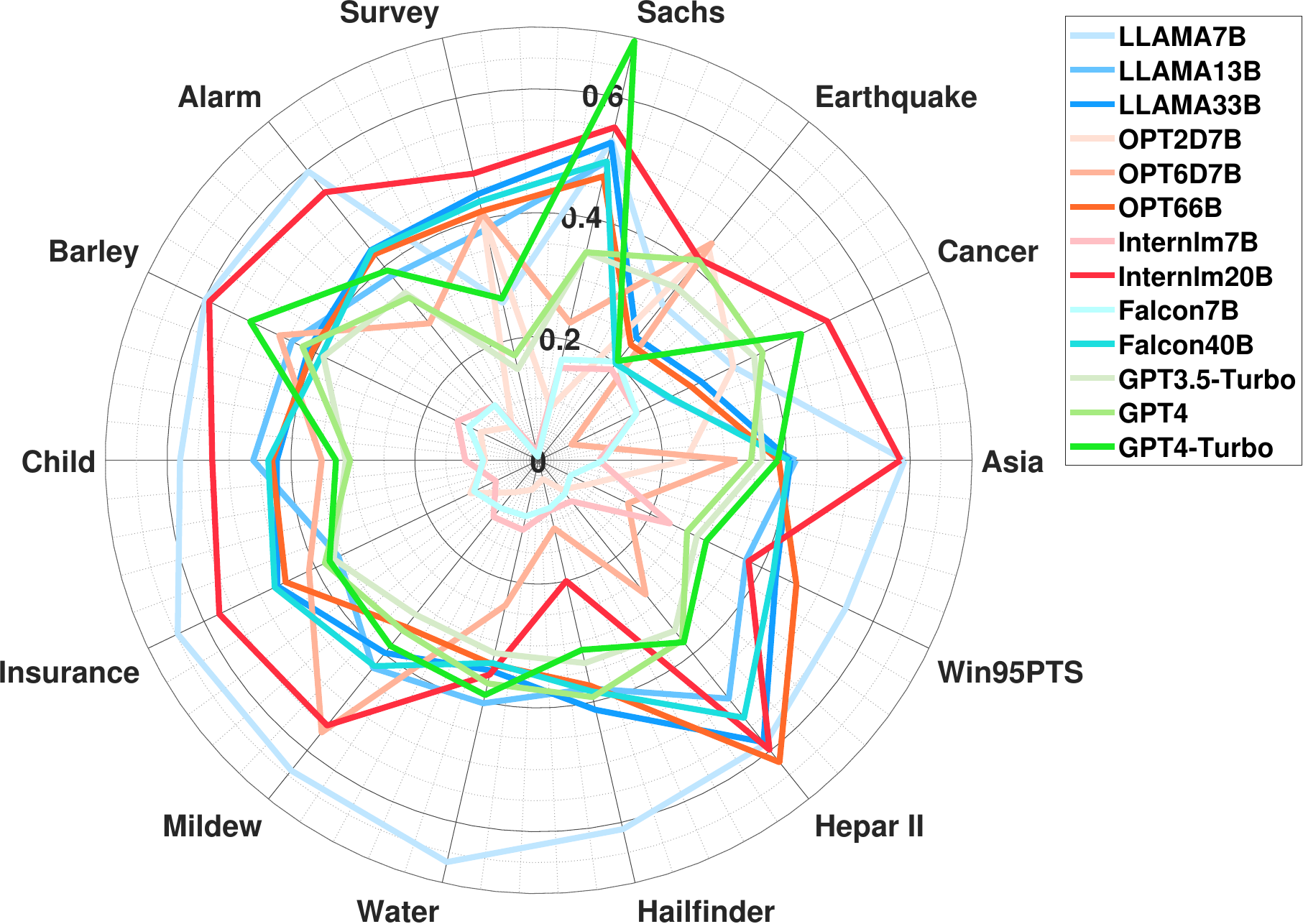}\label{App_fig:ske_reda_F1}}
	\subfigure[Accuracy]{\includegraphics[width=0.45\linewidth]{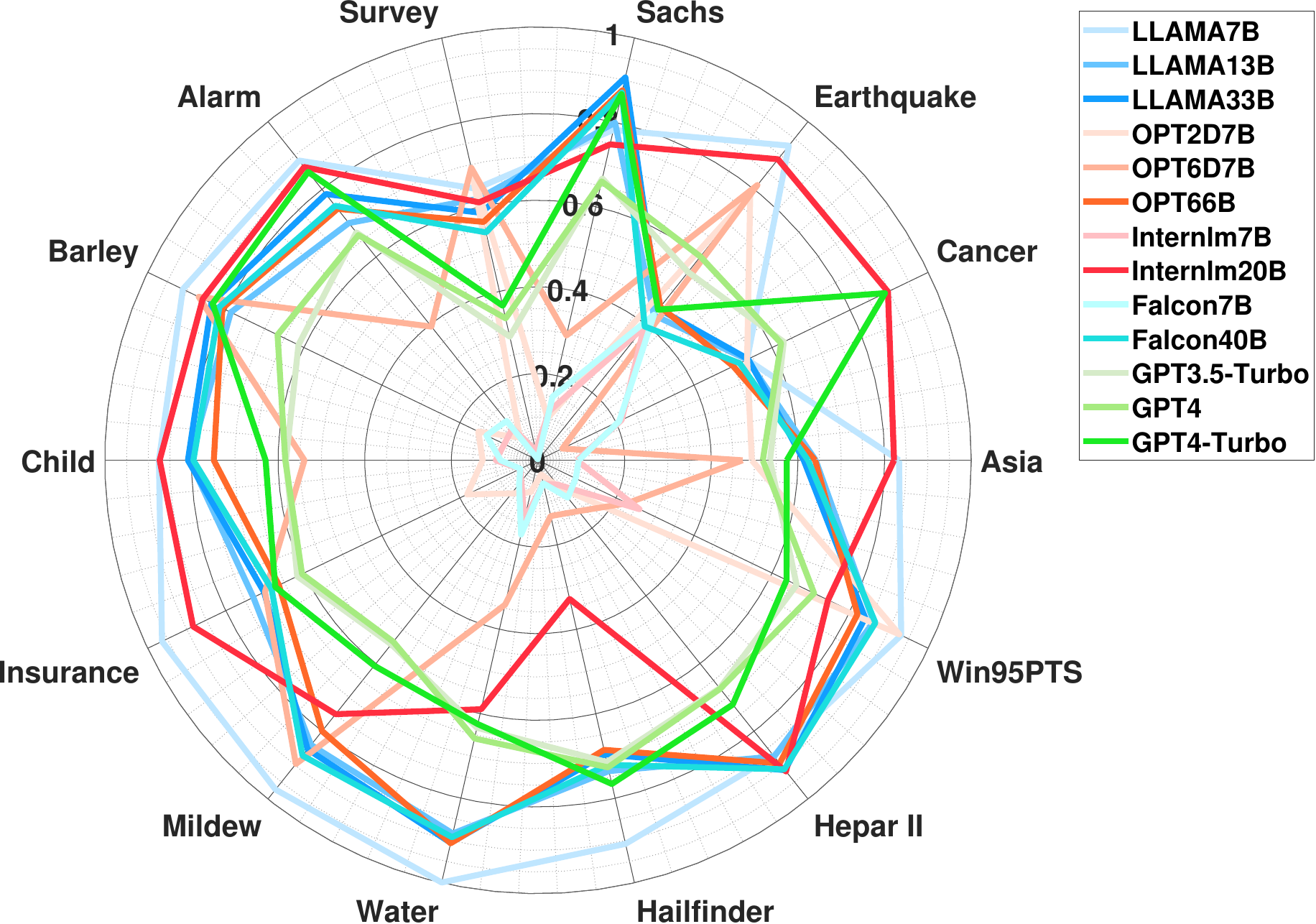}\label{App_fig:ske_reda_Acc}}\\
	\caption{Performance of causal skeleton identification for different datasets.}
	\label{App_fig:ske}
\end{figure}

In this paper, we examine the performance of LLMs in  causal skeleton identification task. The overall experimental results are presented in Tables \ref{tab:ske_F1} and \ref{tab:ske_acc}, as well as Fig. \ref{App_fig:ske}. Table \ref{tab:ske_F1} and Fig. \ref{App_fig:ske_reda_F1} record F1 scores of different LLMs on causal skeleton identification tasks across different datasets, while Table \ref{tab:ske_acc} and Fig. \ref{App_fig:ske_reda_Acc} record the accuracy of these LLMs on the causal skeleton identification task. We analyze these results in Section III-D.
\begin{table}[htbp]
	\centering
	\caption{F1 score of causal skeleton identification task for different datasets}
	\scalebox{0.85}{
		\begin{tabular}{lcccccccccccccc}
			\toprule
			LLM/F1 Score & Asia  & Cancer & Earthquake & Sachs & Survey & Alarm & Barley & Child & Insurance & Mildew & Water & Hailfinder & Hepar II & Win95PTS \\
			\midrule
			LLAMA7B & 0.5910  & 0.3488  & 0.3210  & 0.5296  & 0.2609  & 0.5966  & 0.5977  & 0.5791  & 0.6469  & 0.6409  & 0.6657  & 0.6112  & 0.5866  & 0.5504  \\
			LLAMA13B & 0.4154  & 0.2326  & 0.2000  & 0.4947  & 0.3824  & 0.3809  & 0.4424  & 0.4610  & 0.3586  & 0.4289  & 0.4026  & 0.3780  & 0.4924  & 0.3716  \\
			LLAMA33B & 0.4066  & 0.2917  & 0.2533  & 0.5263  & 0.4410  & 0.4342  & 0.4127  & 0.4228  & 0.4684  & 0.3980  & 0.3481  & 0.4135  & 0.5823  & 0.4254  \\
			OPT2D7B & 0.2377  & 0.3488  & 0.4400  & 0.0880  & 0.4085  & 0.0692  & 0.1045  & 0.0867  & 0.1225  & 0.0662  & 0.0490  & 0.0311  & 0.0605  & 0.0895  \\
			OPT6D7B & 0.3208  & 0.0588  & 0.4483  & 0.2281  & 0.4091  & 0.2830  & 0.4650  & 0.3503  & 0.4116  & 0.8621  & 0.2402  & 0.1127  & 0.2772  & 0.1602  \\
			OPT66B & 0.3877  & 0.2738  & 0.2388  & 0.4705  & 0.4127  & 0.4257  & 0.4060  & 0.4304  & 0.4535  & 0.3444  & 0.3380  & 0.3733  & 0.6238  & 0.4619  \\
			Internlm7B & 0.0977  & 0.1765  & 0.1880  & 0.1530  & 0.0155  & 0.1137  & 0.1453  & 0.1185  & 0.0768  & 0.1170  & 0.1153  & 0.0831  & 0.0846  & 0.2351  \\
			Internlm20B & 0.5845  & 0.9574  & 0.9856  & 0.5516  & 0.4750  & 0.5538  & 0.5905  & 0.9271  & 0.5724  & 0.5478  & 0.3551  & 0.2000  & 0.5980  & 0.3762  \\
			Falcon7B & 0.1046  & 0.1756  & 0.2040  & 0.1670  & 0.0059  & 0.1139  & 0.1249  & 0.0896  & 0.1157  & 0.0987  & 0.0927  & 0.0792  & 0.0685  & 0.0570  \\
			Falcon40B & 0.4041  & 0.2348  & 0.1995  & 0.4951  & 0.4288  & 0.4331  & 0.3894  & 0.4366  & 0.4731  & 0.4252  & 0.3360  & 0.3827  & 0.5319  & 0.4204  \\
			GPT3.5-Turbo & 0.3625  & 0.3864  & 0.3571  & 0.3440  & 0.1509  & 0.3460  & 0.3858  & 0.3115  & 0.3675  & 0.3188  & 0.3196  & 0.3360  & 0.3524  & 0.2836  \\
			GPT4  & 0.3432  & 0.4012  & 0.4135  & 0.3446  & 0.1736  & 0.3362  & 0.4234  & 0.3044  & 0.3830  & 0.3517  & 0.3697  & 0.3924  & 0.3717  & 0.2661  \\
			GPT4-Turbo & 0.3863  & 0.4706  & 0.2051  & 0.6946  & 0.2679  & 0.3921  & 0.5171  & 0.3273  & 0.3747  & 0.3836  & 0.3888  & 0.3141  & 0.3759  & 0.3004  \\
			\bottomrule
		\end{tabular}%
		\label{tab:ske_F1}%
	}
\end{table}%

\begin{table}[htbp]
	\centering
	\caption{Accuracy of causal skeleton identification task for different datasets}
	\scalebox{0.85}{
		\begin{tabular}{lcccccccccccccc}
			\toprule
			LLM/Accuracy & Asia  & Cancer & Earthquake & Sachs & Survey & Alarm & Barley & Child & Insurance & Mildew & Water & Hailfinder & Hepar II & Win95PTS \\
			\midrule
			LLAMA7B & 0.8333  & 0.5357  & 0.9286  & 0.7821  & 0.6429  & 0.8847  & 0.9099  & 0.8750  & 0.9632  & 0.9717  & 0.9993  & 0.9079  & 0.8719  & 0.9305  \\
			LLAMA13B & 0.6410  & 0.5357  & 0.4286  & 0.7994  & 0.6190  & 0.7003  & 0.7870  & 0.8040  & 0.7292  & 0.8401  & 0.8835  & 0.7356  & 0.8743  & 0.8597  \\
			LLAMA33B & 0.6134  & 0.5384  & 0.4469  & 0.9063  & 0.5856  & 0.7858  & 0.8366  & 0.8080  & 0.7016  & 0.8529  & 0.9054  & 0.6961  & 0.9145  & 0.8382  \\
			OPT2D7B & 0.4936  & 0.5357  & 0.7857  & 0.1003  & 0.6905  & 0.0670  & 0.1530  & 0.1278  & 0.1814  & 0.0978  & 0.0756  & 0.0334  & 0.0813  & 0.9256  \\
			OPT6D7B & 0.4722  & 0.0625  & 0.8125  & 0.2955  & 0.6923  & 0.3946  & 0.8692  & 0.5391  & 0.6994  & 0.8963  & 0.3417  & 0.1320  & 0.1500  & 0.2285  \\
			OPT66B & 0.6398  & 0.5003  & 0.4529  & 0.8758  & 0.5643  & 0.7422  & 0.8085  & 0.7483  & 0.6638  & 0.7993  & 0.9065  & 0.6854  & 0.8945  & 0.8186  \\
			Internlm7B & 0.0897  & 0.2143  & 0.3929  & 0.1160  & 0.0238  & 0.0947  & 0.0863  & 0.0976  & 0.0456  & 0.0618  & 0.1400  & 0.0553  & 0.0716  & 0.2602  \\
			Internlm20B & 0.8718  & 0.9865  & 0.9985  & 0.7476  & 0.9048  & 0.9156  & 0.9388  & 0.9573  & 0.8836  & 0.7492  & 0.5893  & 0.3282  & 0.9181  & 0.7430  \\
			Falcon7B & 0.0940  & 0.2081  & 0.4295  & 0.1504  & 0.0025  & 0.1163  & 0.1321  & 0.0864  & 0.0452  & 0.0701  & 0.1760  & 0.0545  & 0.1104  & 0.0965  \\
			Falcon40B & 0.6235  & 0.5165  & 0.3951  & 0.8702  & 0.5395  & 0.7510  & 0.8162  & 0.7954  & 0.6828  & 0.8717  & 0.8921  & 0.7211  & 0.9119  & 0.8643  \\
			GPT3.5-Turbo & 0.5355  & 0.6296  & 0.5556  & 0.6672  & 0.2927  & 0.6727  & 0.6136  & 0.5818  & 0.6169  & 0.5430  & 0.6331  & 0.7141  & 0.6727  & 0.6625  \\
			GPT4  & 0.5189  & 0.6228  & 0.6012  & 0.6601  & 0.3360  & 0.6646  & 0.6672  & 0.5833  & 0.6046  & 0.5356  & 0.6585  & 0.7267  & 0.6744  & 0.7073  \\
			GPT4-Turbo & 0.5742  & 0.8889  & 0.4444  & 0.8666  & 0.3659  & 0.8511  & 0.8336  & 0.6288  & 0.6733  & 0.6062  & 0.6245  & 0.7661  & 0.7210  & 0.6369  \\
			\bottomrule
		\end{tabular}%
		\label{tab:ske_acc}%
	}
\end{table}%

\clearpage
\subsection{Detailed Evaluation Results for the Causality Identification Task for the First Method}\label{app:add_cau1}
\begin{figure}[ht] 
	\centering
	\subfigure[F1 score]{\includegraphics[width=0.45\linewidth]{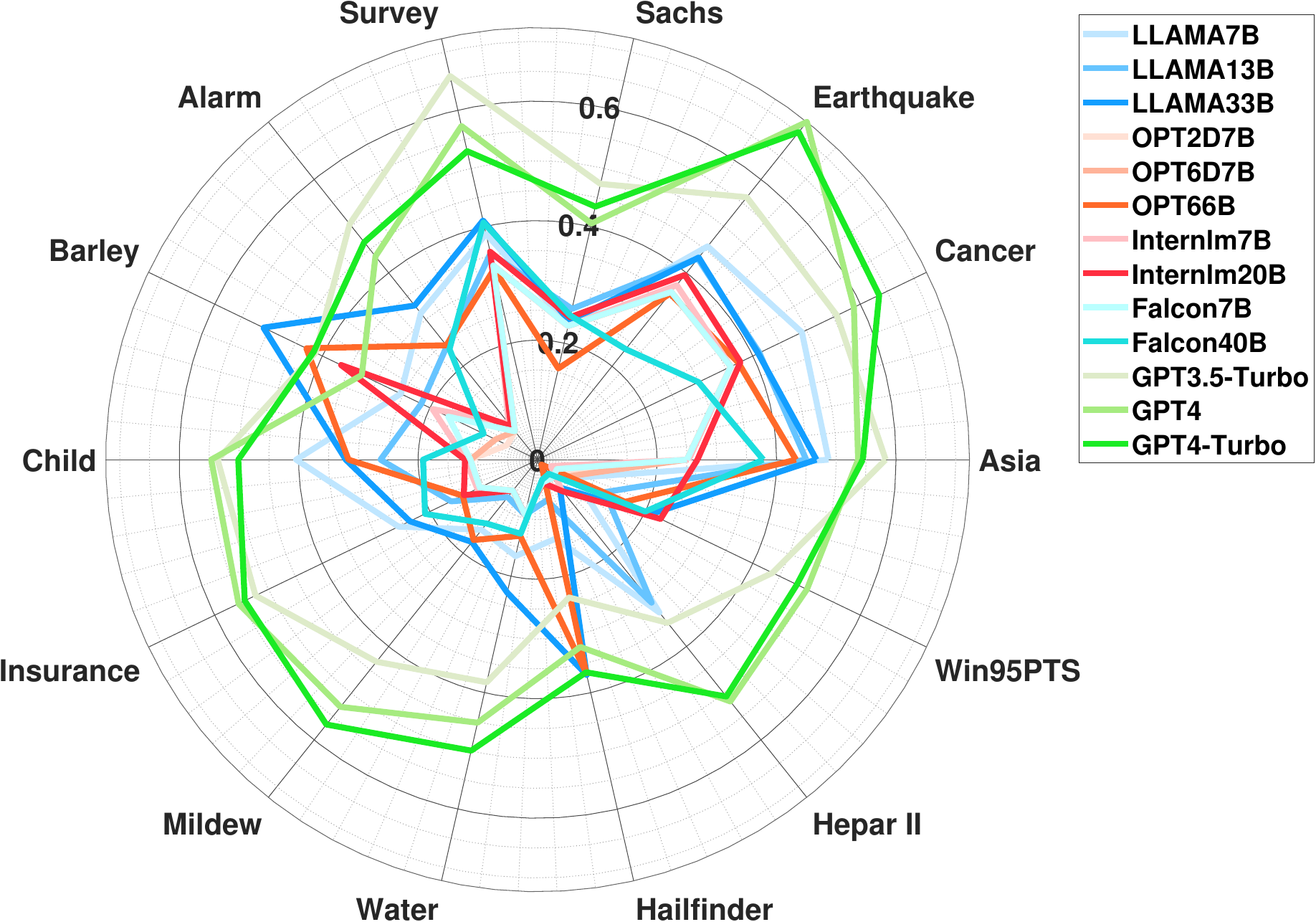}\label{App_fig:cas1_F1}}
	\subfigure[Accuracy]{\includegraphics[width=0.45\linewidth]{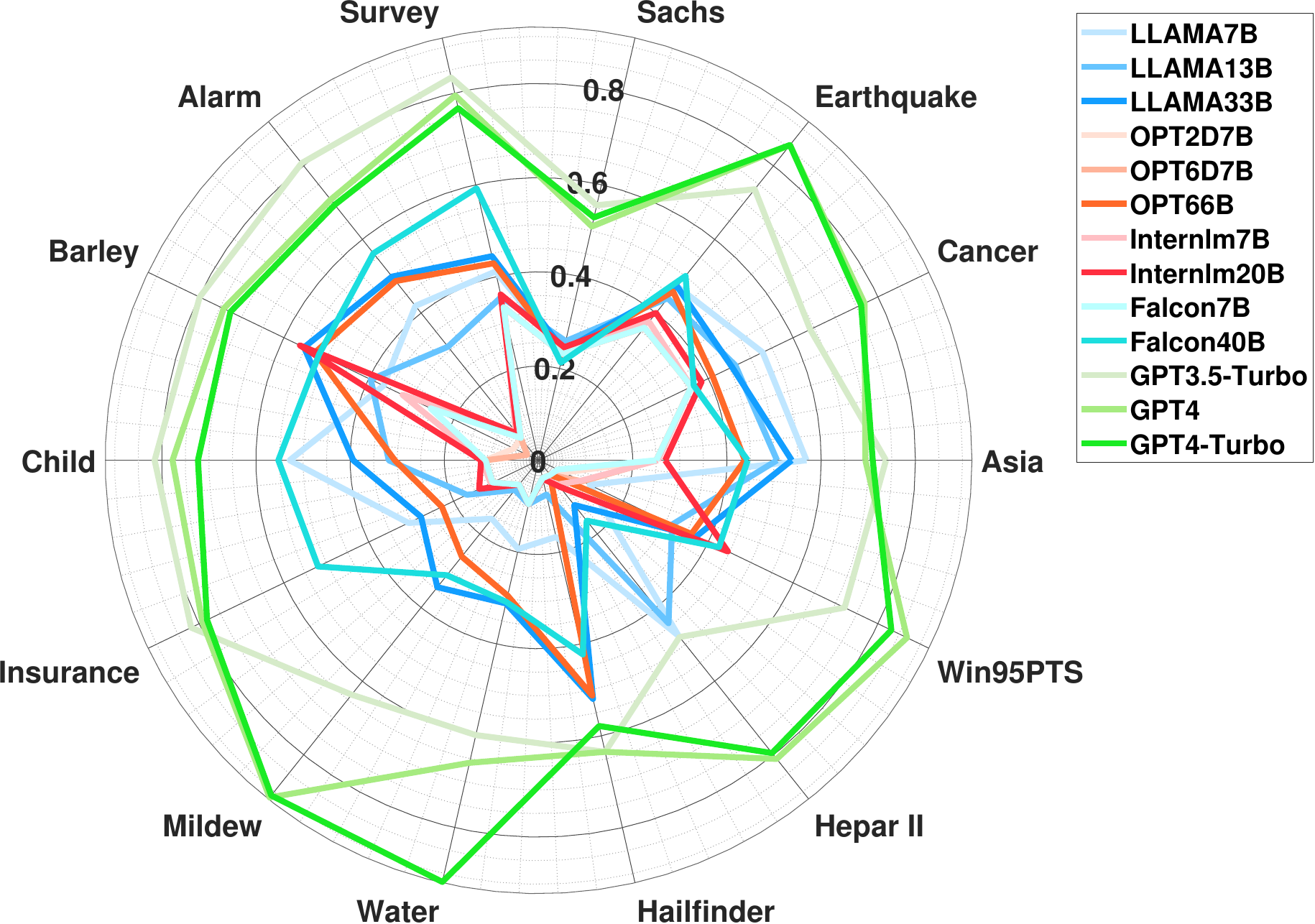}\label{App_fig:cas1_Acc}}\\
	\caption{Performance of causality identification task for the first method.}
	\label{App_fig:cas1}
\end{figure}

In this paper, we present the complete experimental results for LLMs on the causality identification task. The overall results are shown in Tables \ref{tab:cau1_F1} and \ref{tab:cau1_acc}, as well as Fig. \ref{App_fig:cas1}. Table \ref{tab:cau1_F1} and Fig. \ref{App_fig:cas1_F1} detail F1 score for the causality identification task using the first method across various datasets, {while Table \ref{App_fig:cas1} and Fig. \ref{App_fig:cas1_Acc} provide the accuracy results of these LLMs for the same task for the first method. The analysis of these results is provided in Section III-E.}
\begin{table}[htbp]
	\centering
	\caption{F1 score of causality identification task for the first method.}
	\scalebox{0.8}{
		\begin{tabular}{lcccccccccccccc}
			\toprule
			F1 Score & Asia  & Cancer & Earthquake & Sachs & Survey & Alarm & Barley & Child & Insurance & Mildew & Water & Hailfinder & Hepar II & Win95PTS \\
			\midrule
			LLAMA7B & 0.4871  & 0.4911  & 0.4566  & 0.2314  & 0.3874  & 0.3135  & 0.2541  & 0.4041  & 0.2583  & 0.1501  & 0.1655  & 0.1355  & 0.3272  & 0.0700  \\
			LLAMA13B & 0.4497  & 0.4122  & 0.4304  & 0.2589  & 0.3542  & 0.2434  & 0.2159  & 0.2629  & 0.1607  & 0.0802  & 0.0947  & 0.0697  & 0.3061  & 0.1252  \\
			LLAMA33B & 0.4668  & 0.4099  & 0.4327  & 0.2384  & 0.4105  & 0.3299  & 0.5093  & 0.3210  & 0.2374  & 0.1762  & 0.2282  & 0.3695  & 0.0597  & 0.2102  \\
			OPT2D7B & 0.2500  & 0.3590  & 0.3590  & 0.2295  & 0.3333  & 0.0606  & 0.0565  & 0.1124  & 0.1073  & 0.0660  & 0.0947  & 0.0389  & 0.0393  & 0.0427  \\
			OPT6D7B & 0.2500  & 0.3590  & 0.3590  & 0.2295  & 0.3333  & 0.0606  & 0.0782  & 0.1124  & 0.1073  & 0.0660  & 0.0947  & 0.0389  & 0.0393  & 0.0627  \\
			OPT66B & 0.4326  & 0.3733  & 0.3556  & 0.1566  & 0.3271  & 0.2449  & 0.4296  & 0.3165  & 0.1390  & 0.1719  & 0.1308  & 0.3666  & 0.0105  & 0.1627  \\
			Internlm7B & 0.2531  & 0.3590  & 0.3735  & 0.2295  & 0.3333  & 0.0776  & 0.1951  & 0.1203  & 0.1119  & 0.0662  & 0.0947  & 0.0397  & 0.0644  & 0.0248  \\
			Internlm20B & 0.2676  & 0.3768  & 0.3953  & 0.2445  & 0.3570  & 0.0714  & 0.3652  & 0.1215  & 0.1370  & 0.0660  & 0.0947  & 0.0402  & 0.0678  & 0.2285  \\
			Falcon7B & 0.2500  & 0.3590  & 0.3590  & 0.2295  & 0.3333  & 0.0606  & 0.1659  & 0.1124  & 0.1073  & 0.0660  & 0.0947  & 0.0389  & 0.0393  & 0.0362  \\
			Falcon40B & 0.3777  & 0.2991  & 0.2365  & 0.2473  & 0.4069  & 0.2364  & 0.1003  & 0.1916  & 0.2091  & 0.1367  & 0.1272  & 0.0345  & 0.0303  & 0.2001  \\
			GPT3.5-Turbo & 0.5827  & 0.5562  & 0.5617  & 0.4733  & 0.6586  & 0.5045  & 0.4126  & 0.5356  & 0.5241  & 0.4325  & 0.3821  & 0.2366  & 0.3493  & 0.4365  \\
			GPT4  & 0.5365  & 0.5877  & 0.7230  & 0.4055  & 0.5727  & 0.4352  & 0.3265  & 0.5463  & 0.5555  & 0.5292  & 0.4514  & 0.3214  & 0.5175  & 0.5004  \\
			GPT4-Turbo & 0.5444  & 0.6346  & 0.7011  & 0.4344  & 0.5295  & 0.4654  & 0.4159  & 0.5010  & 0.5441  & 0.5669  & 0.4994  & 0.3649  & 0.5069  & 0.4818  \\
			\bottomrule
		\end{tabular}%
		\label{tab:cau1_F1}%
	}
\end{table}%

\begin{table}[htbp]
	\centering
	\caption{Accuracy of causality identification task for the first method.}
	\scalebox{0.8}{
		\begin{tabular}{lcccccccccccccc}
			\toprule
			Accuracy & Asia  & Cancer & Earthquake & Sachs & Survey & Alarm & Barley & Child & Insurance & Mildew & Water & Hailfinder & Hepar II & Win95PTS \\
			\midrule
			LLAMA7B & 0.5688  & 0.5280  & 0.4720  & 0.2331  & 0.4111  & 0.4193  & 0.3652  & 0.5285  & 0.3064  & 0.1587  & 0.1932  & 0.1670  & 0.4766  & 0.1205  \\
			LLAMA13B & 0.5063  & 0.4640  & 0.4400  & 0.2595  & 0.3556  & 0.3084  & 0.3955  & 0.3180  & 0.1687  & 0.0805  & 0.0957  & 0.0746  & 0.4424  & 0.3146  \\
			LLAMA33B & 0.5375  & 0.4560  & 0.4720  & 0.2397  & 0.4444  & 0.4988  & 0.5524  & 0.3934  & 0.2768  & 0.3446  & 0.3117  & 0.5197  & 0.1214  & 0.3719  \\
			OPT2D7B & 0.2500  & 0.3600  & 0.3600  & 0.2314  & 0.3333  & 0.0606  & 0.0579  & 0.1125  & 0.1084  & 0.0661  & 0.0957  & 0.0389  & 0.0394  & 0.0598  \\
			OPT6D7B & 0.2500  & 0.3600  & 0.3600  & 0.2314  & 0.3333  & 0.0606  & 0.0265  & 0.1125  & 0.1084  & 0.0661  & 0.0957  & 0.0389  & 0.0394  & 0.1029  \\
			OPT66B & 0.4379  & 0.4103  & 0.4581  & 0.2306  & 0.4290  & 0.4867  & 0.5271  & 0.3051  & 0.2277  & 0.2615  & 0.2948  & 0.5125  & 0.0383  & 0.3603  \\
			Internlm7B & 0.2531  & 0.3600  & 0.3760  & 0.2314  & 0.3333  & 0.0782  & 0.3205  & 0.1205  & 0.1128  & 0.0663  & 0.0957  & 0.0397  & 0.0654  & 0.0998  \\
			Internlm20B & 0.2688  & 0.3840  & 0.4000  & 0.2463  & 0.3611  & 0.0719  & 0.5623  & 0.1220  & 0.1399  & 0.0661  & 0.0957  & 0.0402  & 0.0718  & 0.4461  \\
			Falcon7B & 0.2500  & 0.3600  & 0.3600  & 0.2314  & 0.3333  & 0.0606  & 0.2542  & 0.1125  & 0.1084  & 0.0661  & 0.0957  & 0.0389  & 0.0394  & 0.0452  \\
			Falcon40B & 0.4440  & 0.3656  & 0.5005  & 0.2127  & 0.5922  & 0.5621  & 0.5173  & 0.5523  & 0.5195  & 0.3121  & 0.3083  & 0.4228  & 0.1639  & 0.4252  \\
			GPT3.5-Turbo & 0.7375  & 0.6400  & 0.7360  & 0.5554  & 0.8333  & 0.8069  & 0.7985  & 0.8150  & 0.8187  & 0.6365  & 0.5984  & 0.6349  & 0.4800  & 0.7210  \\
			GPT4  & 0.6938  & 0.7680  & 0.8560  & 0.5091  & 0.7944  & 0.7078  & 0.7421  & 0.7770  & 0.7920  & 0.9156  & 0.6588  & 0.6349  & 0.8109  & 0.8681  \\
			GPT4-Turbo & 0.7094  & 0.7600  & 0.8560  & 0.5289  & 0.7667  & 0.6938  & 0.7260  & 0.7230  & 0.7811  & 0.9100  & 0.9182  & 0.5786  & 0.7949  & 0.8312  \\
			\bottomrule
		\end{tabular}%
		\label{tab:cau1_acc}%
	}
\end{table}%

\clearpage
\subsection{Detailed Evaluation Results for the Causality Identification Task for the Second Method}\label{app:add_cau2}
\begin{figure}[ht]
	\centering
	\subfigure[F1 score]{\includegraphics[width=0.45\linewidth]{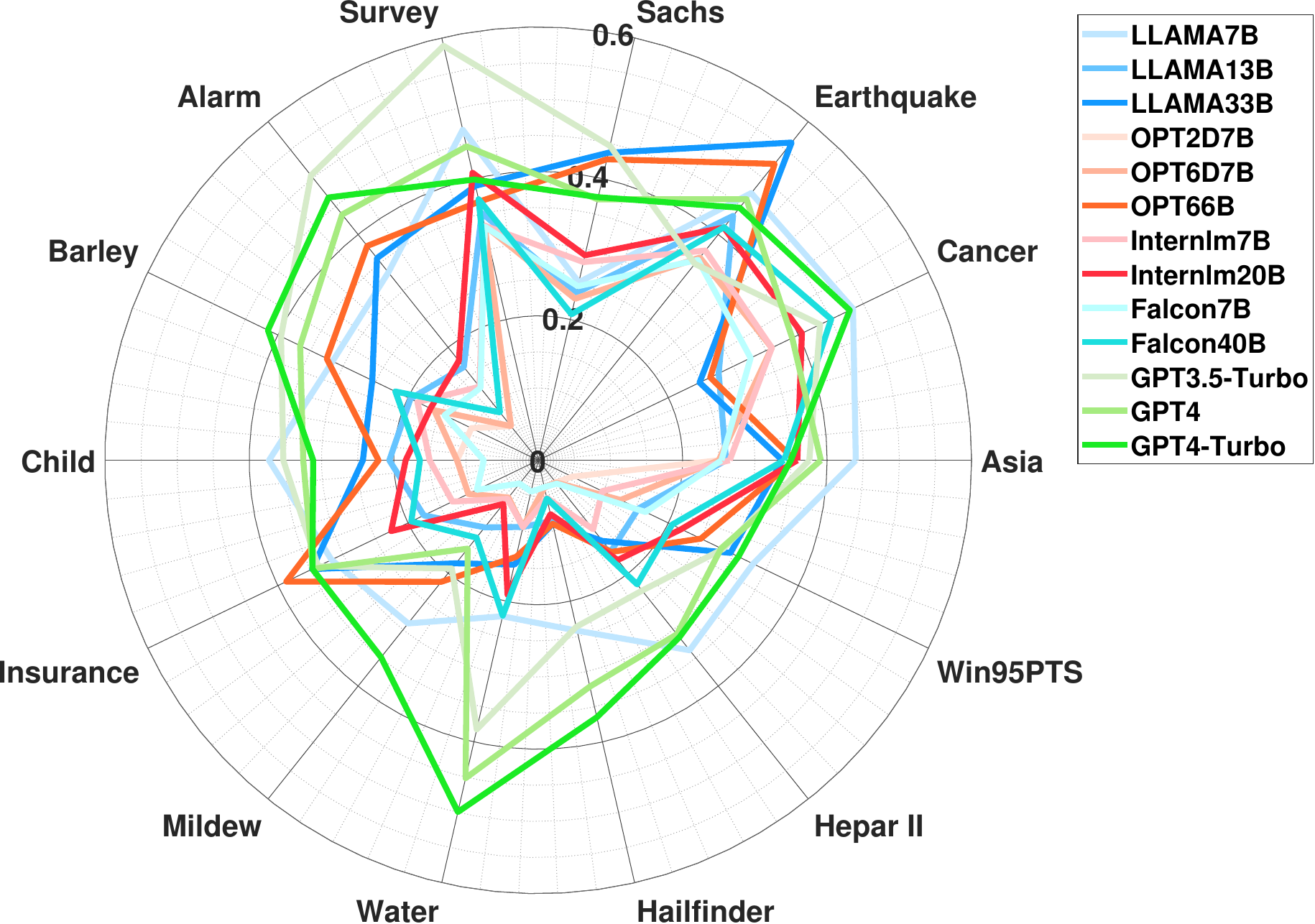}\label{App_fig:cas2_F1}}
	\subfigure[Accuracy]{\includegraphics[width=0.45\linewidth]{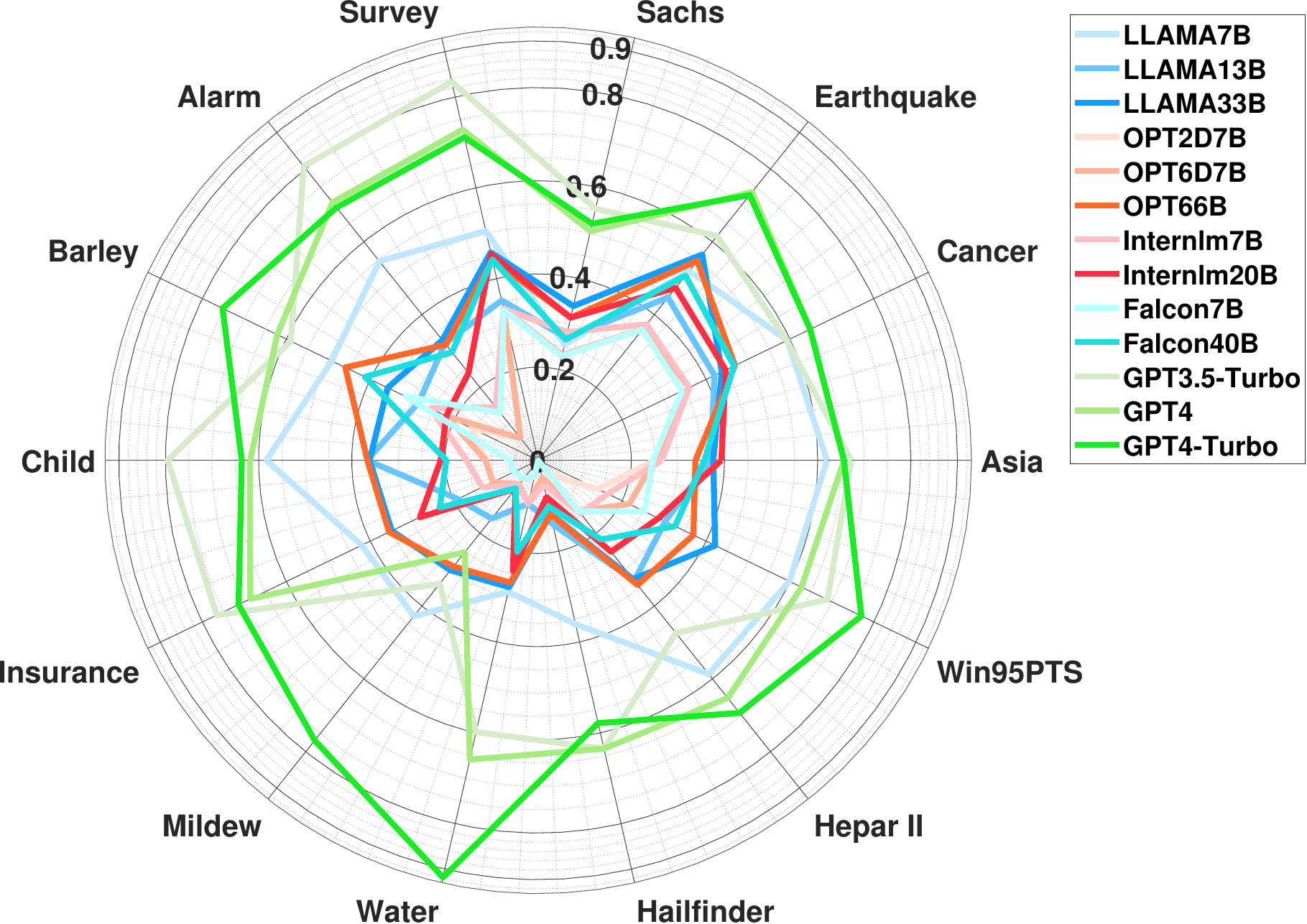}\label{App_fig:cas2_Acc}}\\
	\caption{{Performance on causality identification for the second method.}}
	\label{App_fig:cas2}
\end{figure}

In this paper, we provide the complete experimental results of LLMs on the causality identification task for the second method. The overall results are displayed in Tables \ref{tab:cau2_F1} and \ref{tab:cau2_acc}, along with Fig. \ref{App_fig:cas2}. Table \ref{tab:cau2_F1} and Fig. \ref{App_fig:cas2_F1} present the F1 scores for the causality identification task using the second method across various datasets, while Table \ref{tab:cau2_acc} and Fig. \ref{App_fig:cas2_Acc} show the accuracy results on the same task for the second method. The analysis of these results is provided in Section III-E. For the second method, we provide the results in Table \ref{tab:causal_relationship}. Overall, LLMs fail to effectively identify causal relationships under this setting. Across all datasets, F1 scores rarely exceed 0.4 and accuracies remain below 50\%, with performance deteriorating sharply on medium and large datasets where F1 scores drop to 0.1–0.2 and accuracies to 20–30\%. For SHD and SID, LLMs perform worse than classical methods. Furthermore, the generated graphs tend to be dense, with sparsity values between 0.6 and 0.7 across most datasets, indicating that LLMs produce overly connected rather than sparse causal graphs.

\begin{table}[ht]
	\centering
	\caption{Average performance of identifying causality}
	\label{tab:causal_relationship}
	\scalebox{0.9}{
		\begin{tabular}{lccccc}
			\toprule
			Dataset & F1 score* & Accuracy* & SHD* & SID* & {Network~Sparsity*}  \\
			\midrule
			Asia & 0.3252 & 0.4192 & 33.95 & 7.436 & 0.6651 \\
			Cancer & 0.3719 & 0.4725 & 12.88 & 4.819 & 0.7054 \\
			Earthquake & 0.4246 & 0.5070 & 12.10 & 4.526 & 0.7142 \\
			Survey & 0.3994 & 0.4811 & 17.88 & 5.338 & 0.6667 \\
			Sachs & 0.3101 & 0.3316 & 76.12 & 10.57 & 0.7892 \\
			Child & 0.2176 & 0.3518 & 221.6 & 18.20 & 0.6099 \\
			Insurance & 0.2400 & 0.3387 & 431.2 & 24.83 & 0.6400 \\
			Water & 0.2057 & 0.2982 & 591.6 & 33.72 & 0.6214 \\
			Mildew & 0.1506 & 0.2267 & 904.5 & 33.38 & 0.7806 \\
			Alarm & 0.2524 & 0.3581 & 778.0 & 33.57 & 0.5738 \\
			Barley & 0.2496 & 0.3953 & 75.92 & 17.22 & 0.3454 \\
			Hailfinder & 0.1332 & 0.2232 & 2144 & 51.81 & 0.6599 \\
			Hepar II & 0.1764 & 0.3337 & 2845 & 64.13 & 0.6152 \\
			Win95PTS & 0.2131 & 0.3977 & 1037 & 52.01 & 0.3532 \\
			\bottomrule
		\end{tabular}
	}
	
\end{table}
{* Averaged F1 score, accuracy, SHD, SID, and network sparsity are computed over the LLM set listed in Table~\ref{tab:LLM}. The detailed performance of each LLM is provided in Tables~\ref{tab:cau2_F1} and \ref{tab:cau2_acc}, and Figs. \ref{App_fig:cas2}, \ref{fig:cas2_SHD} and \ref{fig:cas2_SID} in Appendix~\ref{app:add_cau2}.}

\begin{table}[htbp]
	\centering
	\caption{F1 score on causality identification task for the second method.}
	\scalebox{0.85}{
		\begin{tabular}{lcccccccccccccc}
			\toprule
			LLM/F1 Score & Asia  & Cancer & Earthquake & Sachs & Survey & Alarm & Barley & Child & Insurance & Mildew & Water & Hailfinder & Hepar II & Win95PTS \\
			\midrule
			LLAMA7B & 0.4397  & 0.4838  & 0.4732  & 0.2528  & 0.4693  & 0.3327  & 0.3159  & 0.3726  & 0.3176  & 0.2889  & 0.2215  & 0.2420  & 0.3364  & 0.3303  \\
			LLAMA13B & 0.2601  & 0.2774  & 0.4328  & 0.2377  & 0.3512  & 0.1651  & 0.1963  & 0.2061  & 0.1757  & 0.1188  & 0.0947  & 0.0859  & 0.1614  & 0.1540  \\
			LLAMA33B & 0.3381  & 0.2480  & 0.5627  & 0.4368  & 0.3899  & 0.3580  & 0.2549  & 0.2427  & 0.3474  & 0.1825  & 0.1481  & 0.0904  & 0.1428  & 0.2954  \\
			OPT2D7B & 0.2500  & 0.3590  & 0.3590  & 0.2295  & 0.3333  & 0.0606  & 0.1026  & 0.1124  & 0.1073  & 0.0660  & 0.0947  & 0.0389  & 0.0393  & 0.0523  \\
			OPT6D7B & 0.2500  & 0.3590  & 0.3590  & 0.2295  & 0.3333  & 0.0606  & 0.1597  & 0.1124  & 0.1073  & 0.0660  & 0.0947  & 0.0389  & 0.0393  & 0.1266  \\
			OPT66B & 0.3554  & 0.2644  & 0.5250  & 0.4275  & 0.3674  & 0.3796  & 0.3249  & 0.2210  & 0.3870  & 0.2151  & 0.1371  & 0.0902  & 0.1623  & 0.2500  \\
			Internlm7B & 0.2651  & 0.3590  & 0.3714  & 0.2815  & 0.3333  & 0.1302  & 0.1855  & 0.1502  & 0.1317  & 0.0683  & 0.0947  & 0.0532  & 0.1222  & 0.0984  \\
			Internlm20B & 0.3588  & 0.4052  & 0.4122  & 0.2911  & 0.4082  & 0.1762  & 0.1655  & 0.1835  & 0.2259  & 0.0774  & 0.1908  & 0.0765  & 0.1766  & 0.2199  \\
			Falcon7B & 0.2537  & 0.3264  & 0.3549  & 0.2472  & 0.3317  & 0.1282  & 0.1439  & 0.0753  & 0.0942  & 0.0411  & 0.0445  & 0.0382  & 0.0413  & 0.1652  \\
			Falcon40B & 0.3404  & 0.4502  & 0.4128  & 0.2074  & 0.3708  & 0.0849  & 0.2195  & 0.1637  & 0.1950  & 0.1370  & 0.2207  & 0.0542  & 0.2197  & 0.2056  \\
			GPT3.5-Turbo & 0.3754  & 0.4331  & 0.3478  & 0.4465  & 0.6586  & 0.5045  & 0.3951  & 0.3529  & 0.3381  & 0.1917  & 0.3821  & 0.2366  & 0.2308  & 0.2836  \\
			GPT4  & 0.3913  & 0.3903  & 0.4625  & 0.3706  & 0.4458  & 0.4352  & 0.3658  & 0.3252  & 0.3462  & 0.1564  & 0.4514  & 0.3214  & 0.3081  & 0.2811  \\
			GPT4-Turbo & 0.3498  & 0.4788  & 0.4473  & 0.3736  & 0.3987  & 0.4654  & 0.4153  & 0.3115  & 0.3469  & 0.3489  & 0.4994  & 0.3649  & 0.3135  & 0.3077  \\
			\bottomrule
		\end{tabular}%
		\label{tab:cau2_F1}%
	}
\end{table}%

\begin{table}[htbp]
	\centering
	\caption{Accuracy on causality identification task for the second method.}
	\scalebox{0.85}{
		\begin{tabular}{lcccccccccccccc}
			\toprule
			LLM/Accuracy & Asia  & Cancer & Earthquake & Sachs & Survey & Alarm & Barley & Child & Insurance & Mildew & Water & Hailfinder & Hepar II & Win95PTS \\
			\midrule
			LLAMA7B & 0.6198  & 0.5933  & 0.5200  & 0.2548  & 0.5046  & 0.5464  & 0.4916  & 0.5833  & 0.4202  & 0.4282  & 0.2892  & 0.3607  & 0.5867  & 0.5984  \\
			LLAMA13B & 0.3698  & 0.4200  & 0.4467  & 0.2700  & 0.3519  & 0.3207  & 0.2848  & 0.3675  & 0.1914  & 0.1601  & 0.0957  & 0.1408  & 0.3308  & 0.3025  \\
			LLAMA33B & 0.3759  & 0.4373  & 0.5649  & 0.3393  & 0.4578  & 0.3298  & 0.3586  & 0.3618  & 0.3488  & 0.3019  & 0.2793  & 0.1178  & 0.3264  & 0.4216  \\
			OPT2D7B & 0.2500  & 0.3600  & 0.3600  & 0.2314  & 0.3333  & 0.0606  & 0.1589  & 0.1125  & 0.1084  & 0.0661  & 0.0957  & 0.0389  & 0.0394  & 0.1424  \\
			OPT6D7B & 0.2500  & 0.3600  & 0.3600  & 0.2314  & 0.3333  & 0.0606  & 0.2190  & 0.1125  & 0.1084  & 0.0661  & 0.0957  & 0.0389  & 0.1394  & 0.2165  \\
			OPT66B & 0.3376  & 0.4687  & 0.5474  & 0.3140  & 0.4421  & 0.3174  & 0.4593  & 0.3664  & 0.3550  & 0.2914  & 0.2691  & 0.1193  & 0.3421  & 0.3698  \\
			Internlm7B & 0.2656  & 0.3600  & 0.3733  & 0.2824  & 0.3333  & 0.1458  & 0.2566  & 0.1533  & 0.1335  & 0.0683  & 0.0957  & 0.0537  & 0.1452  & 0.1659  \\
			Internlm20B & 0.3932  & 0.4467  & 0.4733  & 0.3140  & 0.4537  & 0.2397  & 0.2189  & 0.2092  & 0.2814  & 0.0776  & 0.2435  & 0.0809  & 0.2510  & 0.2865  \\
			Falcon7B & 0.2460  & 0.3424  & 0.3586  & 0.2291  & 0.3304  & 0.1300  & 0.3158  & 0.0612  & 0.0573  & 0.0501  & 0.0363  & 0.0114  & 0.1404  & 0.2536  \\
			Falcon40B & 0.3570  & 0.4666  & 0.5070  & 0.2656  & 0.4421  & 0.2964  & 0.4127  & 0.1971  & 0.2343  & 0.0767  & 0.2012  & 0.1019  & 0.2172  & 0.3260  \\
			GPT3.5-Turbo & 0.6719  & 0.5920  & 0.6160  & 0.5537  & 0.8333  & 0.8069  & 0.5892  & 0.7950  & 0.7657  & 0.3390  & 0.5984  & 0.6349  & 0.4740  & 0.6892  \\
			GPT4  & 0.6563  & 0.6480  & 0.7360  & 0.5041  & 0.7278  & 0.7078  & 0.6214  & 0.6180  & 0.6856  & 0.2526  & 0.6588  & 0.6349  & 0.6525  & 0.6274  \\
			GPT4-Turbo & 0.6563  & 0.6480  & 0.7280  & 0.5207  & 0.7111  & 0.6938  & 0.7517  & 0.6360  & 0.7136  & 0.7690  & 0.9182  & 0.5786  & 0.6933  & 0.7698  \\
			\bottomrule
		\end{tabular}%
		\label{tab:cau2_acc}%
	}
\end{table}%

We visualize the performance on the causality identification task for the second method in terms of SHD in Fig.~\ref{App_fig:cas2}. 
These figures reveal that current LLMs are not yet adept at identifying causal relationships between variables despite their achieving F1 scores and accuracy rates exceeding 0.5 and 60\% on some datasets.
Among all LLMs, closed-source LLMs consistently demonstrate superior performance, achieving lower SHD and SID values across most causal datasets compared to open-source LLMs. Moreover, among open-source LLMs, LLAMA series LLMs are second only to closed-source LLMs in their ability to understand causal relationships on small and medium scale datasets, while Falcon series excel in identifying causality over large scale datasets. Specifically, GPT4-Turbo exhibits the best performance among closed-source LLMs, and for open-source LLMs, LLAMA-7B, Falcon-7B, and Falcon-40B outperform other LLMs, closely following GPT-based LLMs. 

We also visualize the performance in terms of SID in Fig \ref{fig:cas2_SID}. 
This figure reveals that current LLMs are not yet adept at identifying causal relationships between variables despite their achieving F1 scores and accuracy rates exceeding 0.5 and 60\% on some datasets. Additionally, it is supported by SHD metrics presented in Fig. \ref{fig:cas2_SHD}.
Among all LLMs, closed-source LLMs consistently demonstrate superior performance, achieving lower SHD and SID values across most causal datasets compared to open-source LLMs. Moreover, among open-source LLMs, LLAMA series are second only to closed-source LLMs in their ability to understand causal relationships on small and medium scale datasets, while Falcon series excel in identifying causality over large scale datasets. Specifically, GPT4-Turbo exhibits the best performance among closed-source LLMs, and for open-source LLMs, LLAMA-7B, Falcon-7B, and Falcon-40B outperform other LLMs, closely following GPT-based LLMs.

\begin{figure}[htb] 
	\center
	{\includegraphics[width=0.9\linewidth]  {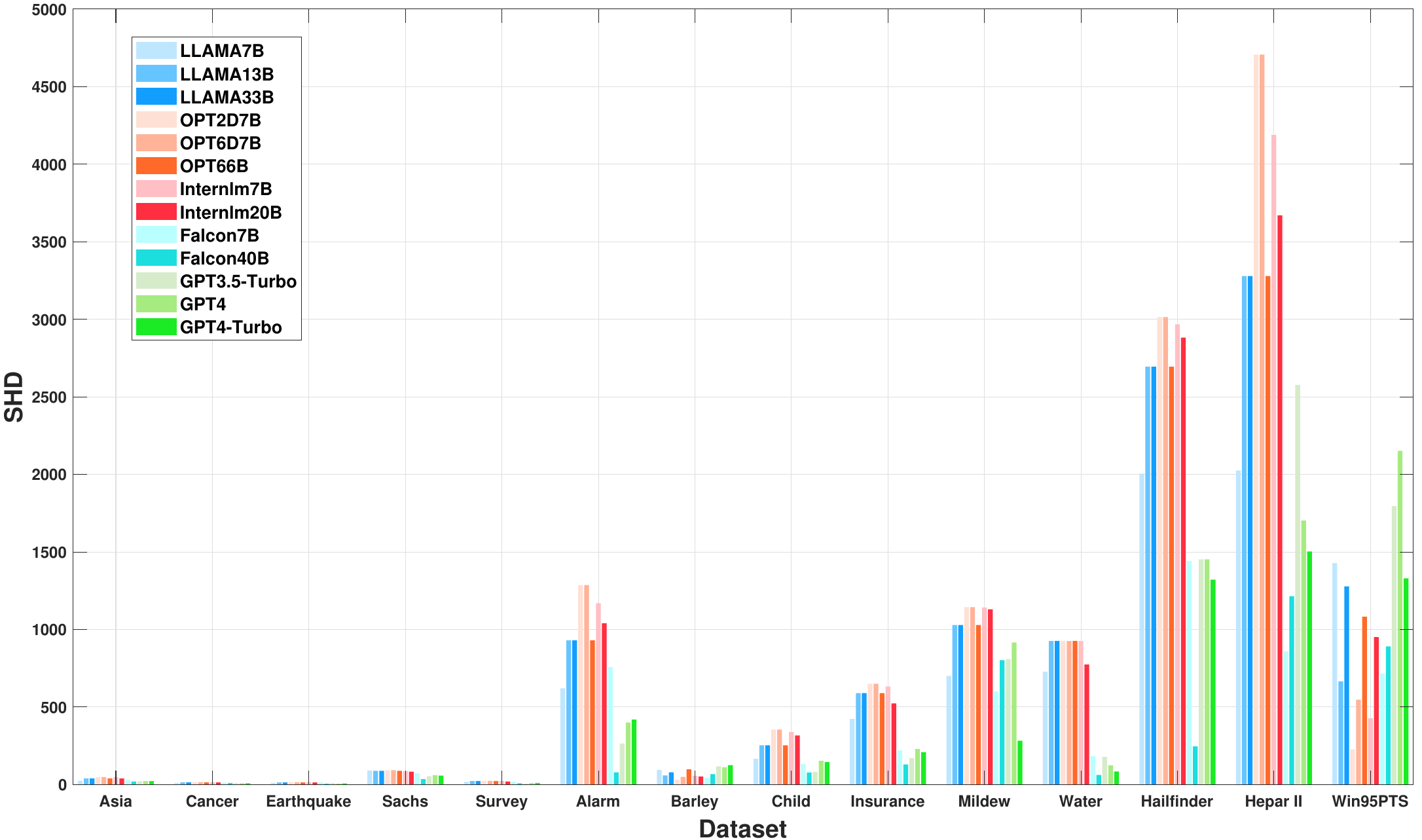}}\caption{SHD of identifying causality for the second method.} 
	\label{fig:cas2_SHD} 
\end{figure}

\begin{figure}[htb] 	
	\center{\includegraphics[width=0.8\linewidth]  {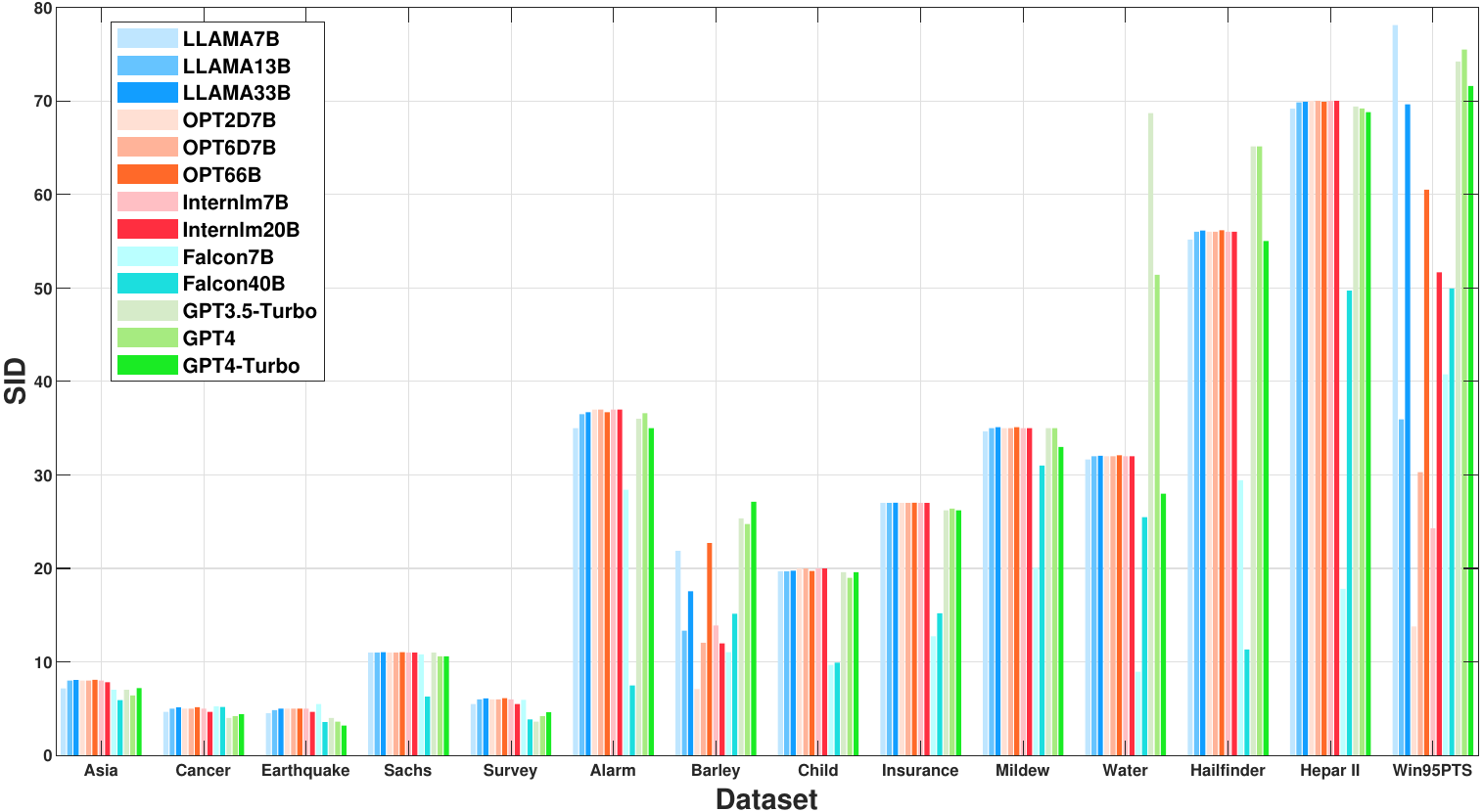}}\caption{SID of causality identification for the second method.} 
	\label{fig:cas2_SID} 
\end{figure}

\begin{figure}[htb] 
	\center{\includegraphics[width=1.0\linewidth]  {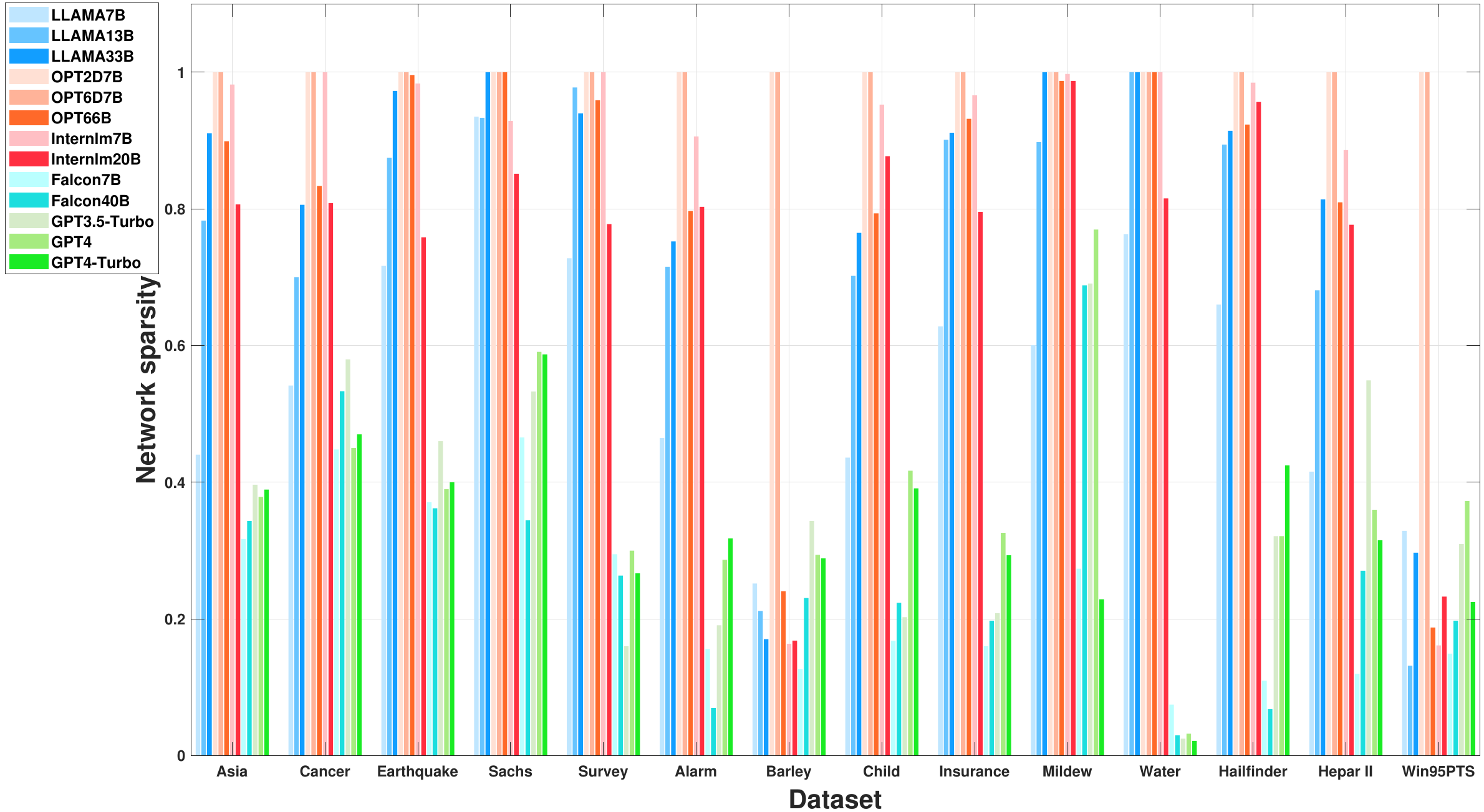}}\caption{Network sparsity of causality identification for the second method.} 
	\label{fig:cas2_sparsity} 
\end{figure}

We provide the network sparsity of causality identification for the second method in Fig. \ref{fig:cas2_sparsity}. Regarding network sparsity, LLMs with lower SHD and SID typically maintain sparser network structures. Closed-source LLMs have an overall network sparsity below 0.5, while Falcon series, which perform well on large scale causal datasets, do not exceed a sparsity level of 0.4. Other open-source models exhibit higher network sparsity.

\clearpage
\subsection{Detailed Evaluation Results for the CoT-Analogous Causal Structure Identification Task}\label{app:cot_table}
We evaluate LLMs with CoT-analogous causal structure identification task. 
Firstly, we design CoT prompting, which presents the LLM with a sequence of question-and-answer pairs that echo the questioning style posed to the model. These question prompts, along with comprehensive CoT findings, are documented in Appendix \ref{app:cot}. The evaluation results are provided in Table \ref{tab:CoT}, which demonstrate notable performance enhancements across all models on both datasets when utilizing CoT prompts. Meanwhile, we find that nearly all LLMs with parameter sizes greater than 6 billion are capable of effectively completing CoT-analogous causal structure identification experiments.
This indicates that LLMs exhibit strong reasoning capabilities in causal learning, which aids in identifying causality. The results also indicate that CausalBN-Bench supports popular prompt techniques and provides guidance for further improving LLMs' causal learning abilities.

\begin{table}[ht]
	\centering
	\caption{Performance for LLMs on CoT-analogous causal structure identification Task.}
	\label{tab:CoT}
	\scalebox{1.0}{
		\begin{tabular}{lcc}
			\toprule
			{LLMs} & {Accuracy} & {Maximal inference length} \\
			\midrule
			BERT-large & -     & - \\
			RoBERTa-large & -     & - \\
			DeBERTa-v3-large  & -     & - \\
			DistilBERT-mnli & -     & - \\
			LLAMA-7B & 100\% & 24 \\
			LLAMA-13B & 100\% & 24 \\
			LLAMA-33B & 60\%  & 21 \\
			OPT-1D3B & -     & - \\
			OPT-2D7B & -     & - \\
			OPT-6D7B & 96\% & 24 \\
			OPT-66B & 89\% & 24 \\
			InternLM-7B & 95\% & 24 \\
			InternLM-20B & 100\% & 24 \\
			Falcon-7B & 93\% & 21 \\
			Falcon-40B & 100\% & 24 \\
			GPT3.5-Turbo & 100\% & 24 \\
			GPT4  & 100\% & 24 \\
			GPT4-Turbo & 100\% & 24 \\
			\bottomrule
	\end{tabular}}
\end{table}

\clearpage
\section{Detailed Evaluation Results for Different Prompt Formats}\label{app:add_exp_prompt}
\subsection{Detailed Evaluation Results for the Causality Identification Task with the Prompt Format ``Variable Name + Background Knowledge''}\label{app:add_kno_cau}

\subsubsection{F1 Score and Accuracy of Causality Identification Task for Prompt Format ``Variable Name + Background Knowledge''}
\begin{figure}[ht]
	\centering
	\subfigure[F1 score]{\includegraphics[width=0.45\linewidth]{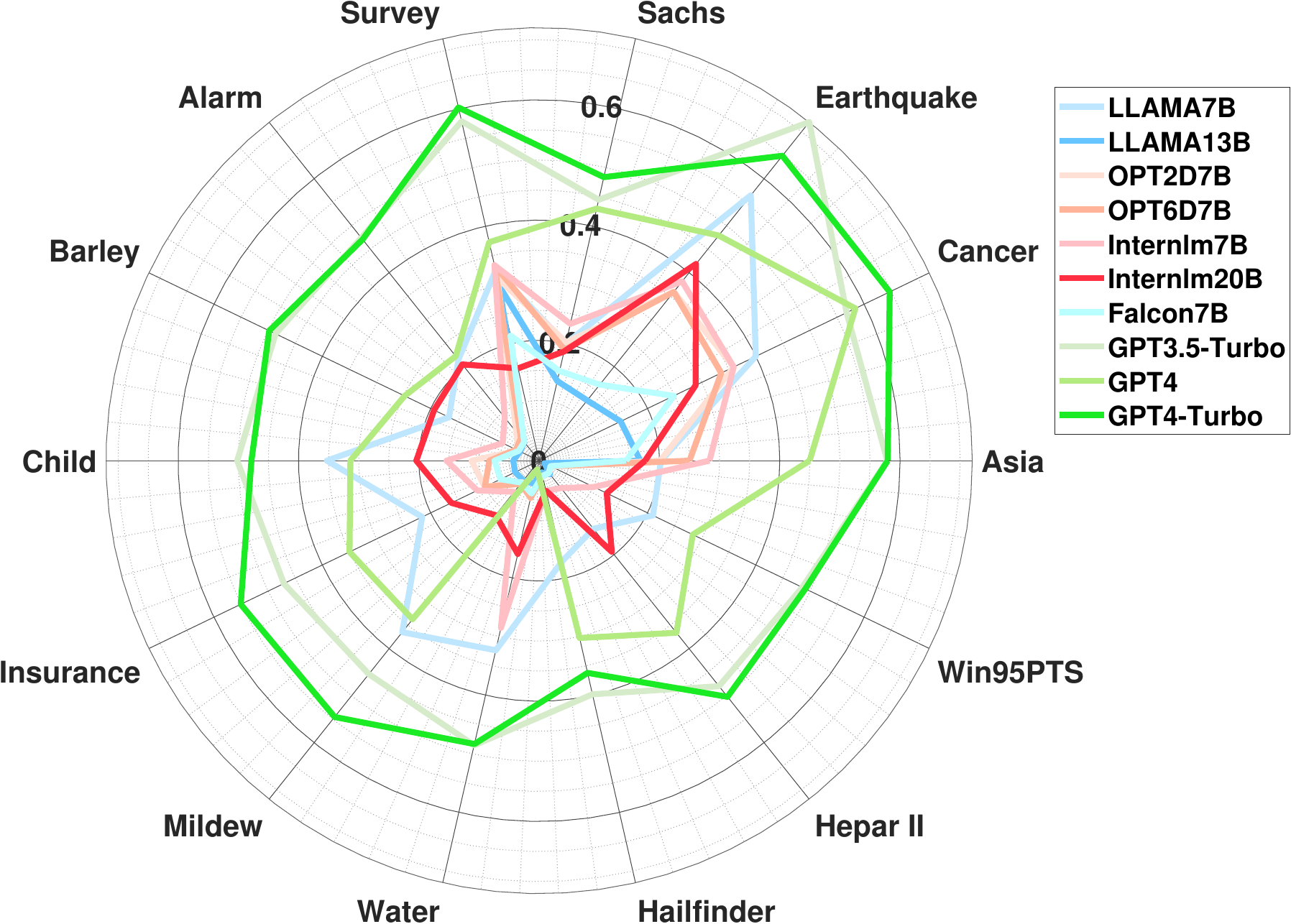}\label{App_fig:kon_cau_F1}}
	\subfigure[Accuracy]{\includegraphics[width=0.45\linewidth]{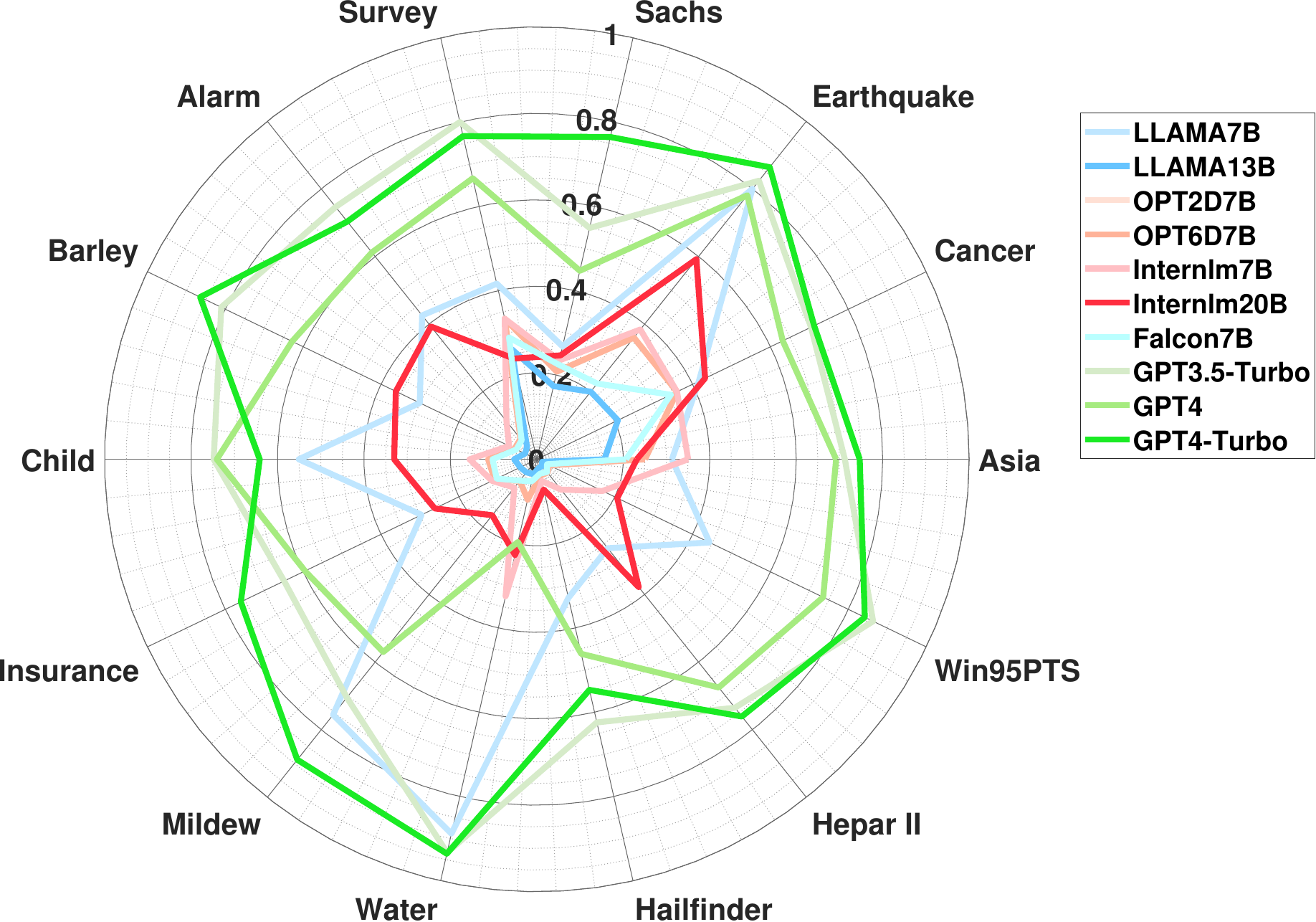}\label{App_fig:kno_cau_Acc}}\\
	\caption{Performance of causality identification for prompt format ``variable name + background knowledge''.}
	\label{App_fig:kno_cau}
\end{figure}

In this paper, we present the comprehensive experimental outcomes of LLMs on the causality identification task for prompt format ``Variable Name + Background Knowledge''. The overall results are illustrated in Tables \ref{tab:cau_kno_F1} and \ref{tab:cau_kno_acc}, as well as Fig. \ref{App_fig:kno_cau}. Specifically, Table \ref{tab:cau_kno_F1} and Fig. \ref{App_fig:kon_cau_F1} display the F1 scores for the causality identification task across various datasets using the first method, while Table \ref{tab:cau_kno_acc} and Fig. \ref{App_fig:kno_cau_Acc} present the accuracy results for the same task using the second method. The analysis of these findings is provided in Section IV-A.
\begin{table}[htbp]
	\centering
	\caption{F1 score of causality identification for prompt format ``variable name + background knowledge''}
	\scalebox{0.85}{
		\begin{tabular}{lcccccccccccccc}
			\toprule
			LLM / F1 Score & Asia  & Cancer & Earthquake & Sachs & Survey & Alarm & Barley & Child & Insurance & Mildew & Water & Hailfinder & Hepar II & Win95PTS \\
			\midrule
			LLAMA7B & 0.2025  & 0.3997  & 0.5643  & 0.1998  & 0.3246  & 0.2150  & 0.1662  & 0.3528  & 0.2157  & 0.3649  & 0.3234  & 0.1717  & 0.1448  & 0.2100  \\
			LLAMA13B & 0.1681  & 0.1504  & 0.1282  & 0.1358  & 0.2979  & 0.0396  & 0.0354  & 0.0430  & 0.0443  & 0.0419  & 0.0394  & 0.0200  & 0.0290  & 0.0099  \\
			OPT2D7B & 0.2013  & 0.3590  & 0.3590  & 0.2019  & 0.3117  & 0.0606  & 0.0388  & 0.1124  & 0.1002  & 0.0455  & 0.0638  & 0.0259  & 0.0262  & 0.0217  \\
			OPT6D7B & 0.2500  & 0.3360  & 0.3590  & 0.1893  & 0.3333  & 0.0485  & 0.0426  & 0.0826  & 0.1002  & 0.0494  & 0.0633  & 0.0258  & 0.0262  & 0.0215  \\
			Internlm7B & 0.2832  & 0.3590  & 0.3813  & 0.2334  & 0.3333  & 0.0917  & 0.0672  & 0.1538  & 0.1142  & 0.0663  & 0.2852  & 0.0515  & 0.0583  & 0.1008  \\
			Internlm20B & 0.1761  & 0.2887  & 0.4184  & 0.1862  & 0.1574  & 0.2047  & 0.1938  & 0.2041  & 0.1618  & 0.1159  & 0.1591  & 0.0511  & 0.1937  & 0.1248  \\
			Falcon7B & 0.1446  & 0.2484  & 0.1617  & 0.1529  & 0.2132  & 0.0397  & 0.0376  & 0.0741  & 0.0712  & 0.0467  & 0.0560  & 0.0261  & 0.0262  & 0.0211  \\
			GPT3.5-Turbo & 0.5787  & 0.5673  & 0.7202  & 0.4450  & 0.5788  & 0.4740  & 0.4862  & 0.5018  & 0.4709  & 0.4540  & 0.4855  & 0.3985  & 0.4797  & 0.4856  \\
			GPT4  & 0.4497  & 0.5834  & 0.4789  & 0.4299  & 0.3727  & 0.2222  & 0.2482  & 0.3137  & 0.3497  & 0.3374  & 0.0148  & 0.3025  & 0.3660  & 0.2835  \\
			GPT4-Turbo & 0.5794  & 0.6468  & 0.6485  & 0.4836  & 0.6024  & 0.4705  & 0.4979  & 0.4782  & 0.5505  & 0.5452  & 0.4834  & 0.3620  & 0.5023  & 0.4912  \\
			\bottomrule
		\end{tabular}%
		\label{tab:cau_kno_F1}%
	}
\end{table}%

\begin{table}[htbp]
	\centering
	\caption{Accuracy of causality identification for prompt format ``variable name + background knowledge''}
	\scalebox{0.85}{
		\begin{tabular}{lcccccccccccccc}
			\toprule
			LLM / Accuracy & Asia  & Cancer & Earthquake & Sachs & Survey & Alarm & Barley & Child & Insurance & Mildew & Water & Hailfinder & Hepar II & Win95PTS \\
			\midrule
			LLAMA7B & 0.3125  & 0.4160  & 0.8000  & 0.2678  & 0.4167  & 0.4256  & 0.3005  & 0.5505  & 0.2960  & 0.7558  & 0.8881  & 0.3285  & 0.2629  & 0.4434  \\
			LLAMA13B & 0.1563  & 0.2080  & 0.2000  & 0.1736  & 0.2611  & 0.0364  & 0.0303  & 0.0525  & 0.0425  & 0.0387  & 0.0350  & 0.0205  & 0.0198  & 0.0132  \\
			OPT2D7B & 0.2500  & 0.3600  & 0.3600  & 0.2080  & 0.3333  & 0.0606  & 0.0567  & 0.1125  & 0.1084  & 0.0640  & 0.0943  & 0.0387  & 0.0393  & 0.0323  \\
			OPT6D7B & 0.2500  & 0.3600  & 0.3600  & 0.2080  & 0.3333  & 0.0606  & 0.0567  & 0.1125  & 0.1084  & 0.0633  & 0.0957  & 0.0385  & 0.0394  & 0.0319  \\
			Internlm7B & 0.3500  & 0.3600  & 0.3840  & 0.2347  & 0.3333  & 0.1126  & 0.0714  & 0.1555  & 0.1150  & 0.0836  & 0.3250  & 0.0517  & 0.0896  & 0.1687  \\
			Internlm20B & 0.2313  & 0.4320  & 0.5920  & 0.2466  & 0.2389  & 0.3927  & 0.3620  & 0.3300  & 0.2620  & 0.1654  & 0.2270  & 0.0716  & 0.3780  & 0.2064  \\
			Falcon7B & 0.2063  & 0.3440  & 0.2240  & 0.2248  & 0.2889  & 0.0564  & 0.0524  & 0.1025  & 0.1029  & 0.0632  & 0.0521  & 0.0341  & 0.0378  & 0.0261  \\
			GPT3.5-Turbo & 0.7125  & 0.7040  & 0.8240  & 0.5488  & 0.8000  & 0.7465  & 0.8102  & 0.7475  & 0.6488  & 0.7048  & 0.9330  & 0.6241  & 0.7349  & 0.8633  \\
			GPT4  & 0.6924  & 0.6314  & 0.7810  & 0.4472  & 0.6665  & 0.6118  & 0.6283  & 0.7403  & 0.5953  & 0.5700  & 0.1962  & 0.4614  & 0.6749  & 0.7351  \\
			GPT4-Turbo & 0.7469  & 0.7120  & 0.8640  & 0.7645  & 0.7667  & 0.7031  & 0.8644  & 0.6410  & 0.7599  & 0.8887  & 0.9355  & 0.5470  & 0.7608  & 0.8420  \\
			\bottomrule
		\end{tabular}%
		\label{tab:cau_kno_acc}%
	}
\end{table}%

\clearpage
\subsubsection{SHD of causality identification for prompt format ``variable name + background knowledge"}\label{app:kno_cas_SHD}
\begin{figure}[htb] 
	\center
	{\includegraphics[width=0.9\linewidth]  {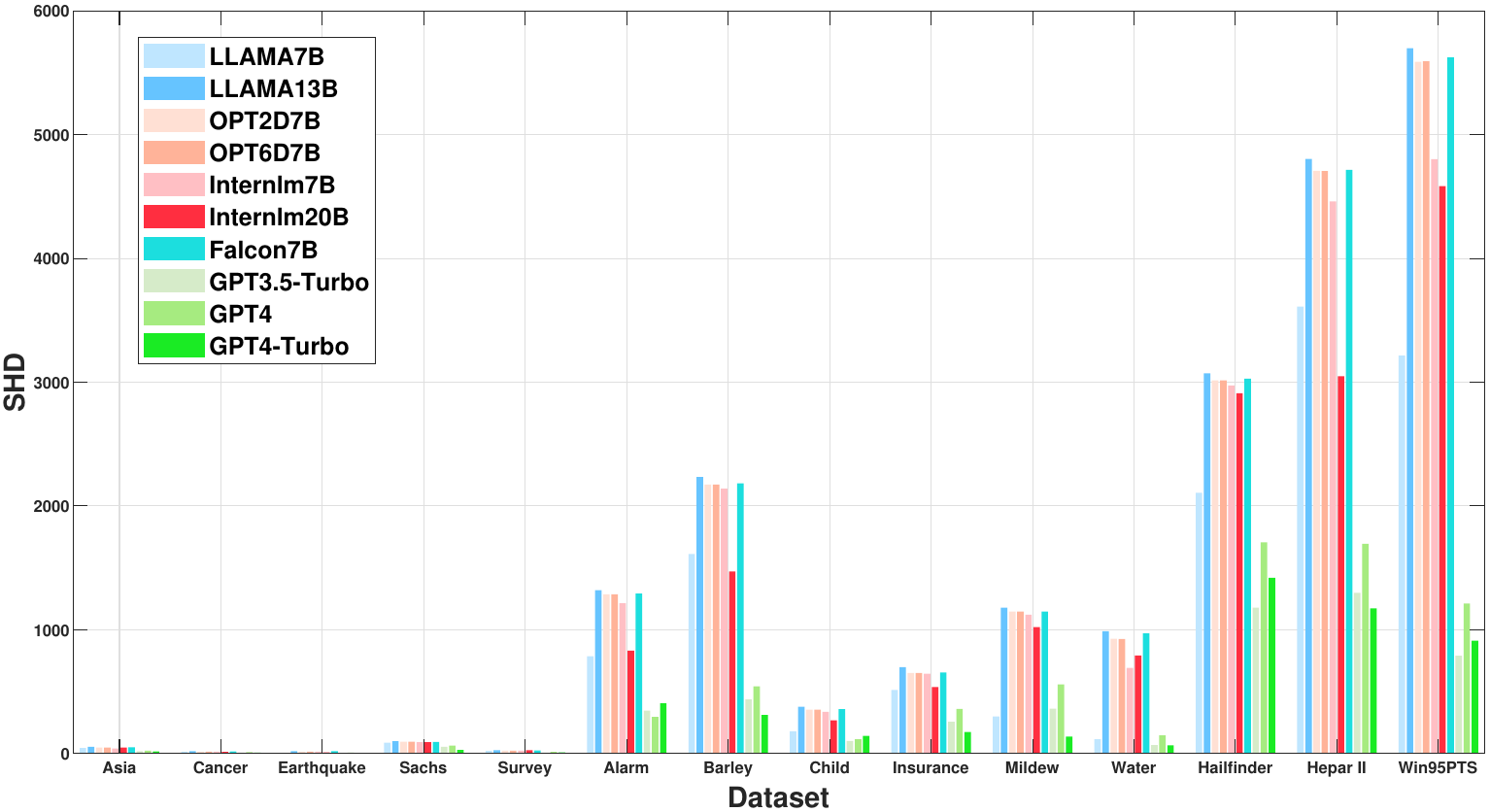}}\caption{SHD of causality identification for prompt format ``variable {name} + background knowledge".} 
	\label{fig:kno_cas_SHD} 
\end{figure}

We provide SHD of causality identification for prompt format ``variable name + background knowledge" in Fig. \ref{fig:kno_cas_SHD}.

\subsubsection{SID of causality identification for prompt format ``variable name + background knowledge"}\label{app:kno_cas_SID}
\begin{figure}[htb] 	
	\center{\includegraphics[width=0.8\linewidth]  {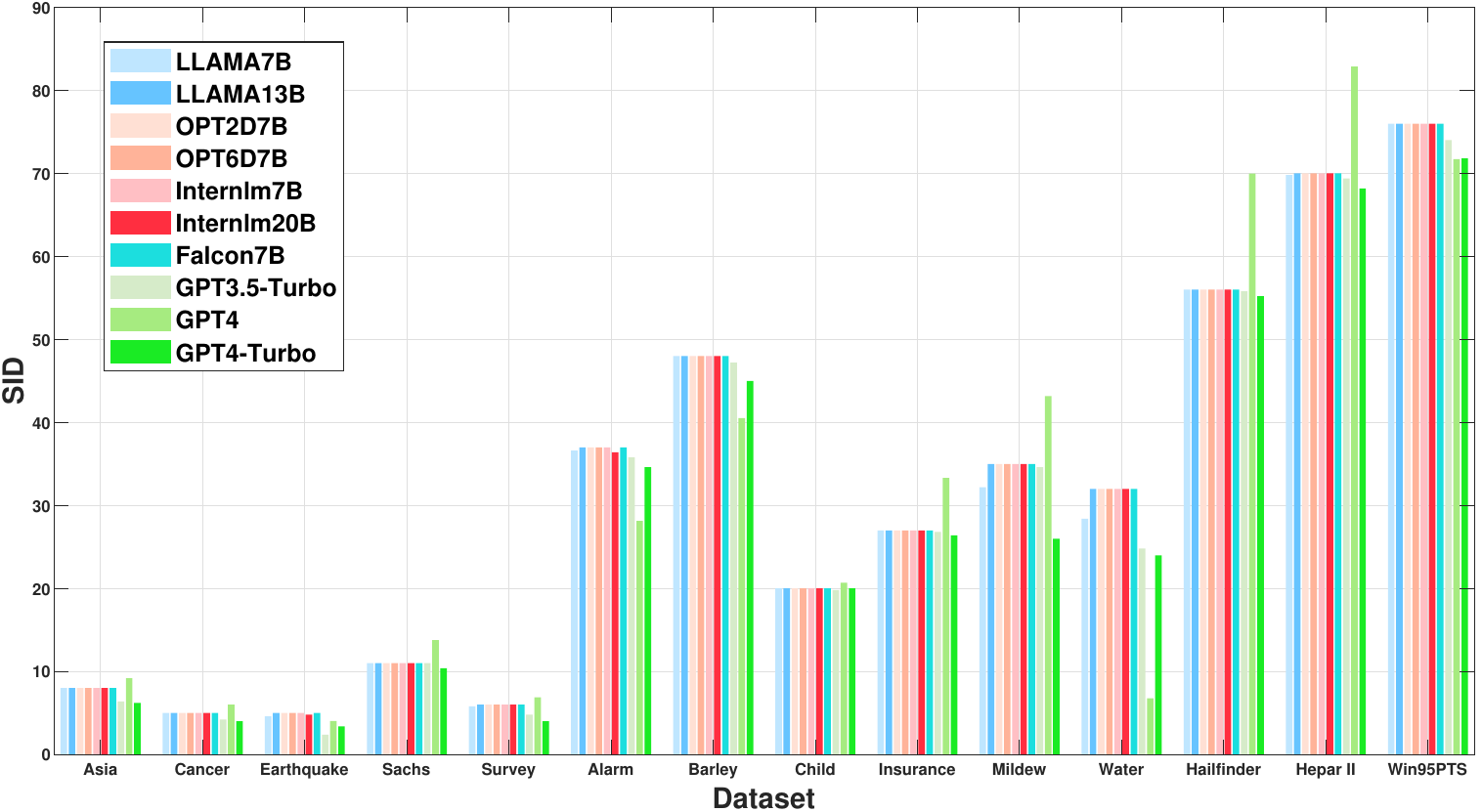}}\caption{SID of causality identification for prompt format ``variable {name} + background knowledge".} 
	\label{fig:kno_cas_SID} 
\end{figure}

We provide SID of causality identification for prompt format ``variable name + background knowledge" in Fig. \ref{fig:kno_cas_SID}. Among all LLMs, closed-source LLMs continue to exhibit better performance than open-source ones, achieving lower SID across most causal datasets, as shown in Fig. \ref{fig:kno_cas_SID}. 
Moreover, among open-source LLMs, LLAMA series rank second to closed-source LLMs in terms of causality identification on small and medium scale datasets, whereas Falcon series demonstrate superior ability to recognize causal relationships on large scale datasets. Specifically, GPT4-Turbo achieves the best performance among closed-source LLMs, and for open-source LLMs, LLAMA-7B and Falcon-7B perform better than other LLMs, trailing closely behind closed-source LLMs. 

\subsubsection{Network sparsity of causality identification for prompt format ``variable name + background knowledge"}\label{app:kno_cas_sparsity}
\begin{figure}[htb] 
	\center{\includegraphics[width=0.9\linewidth]  {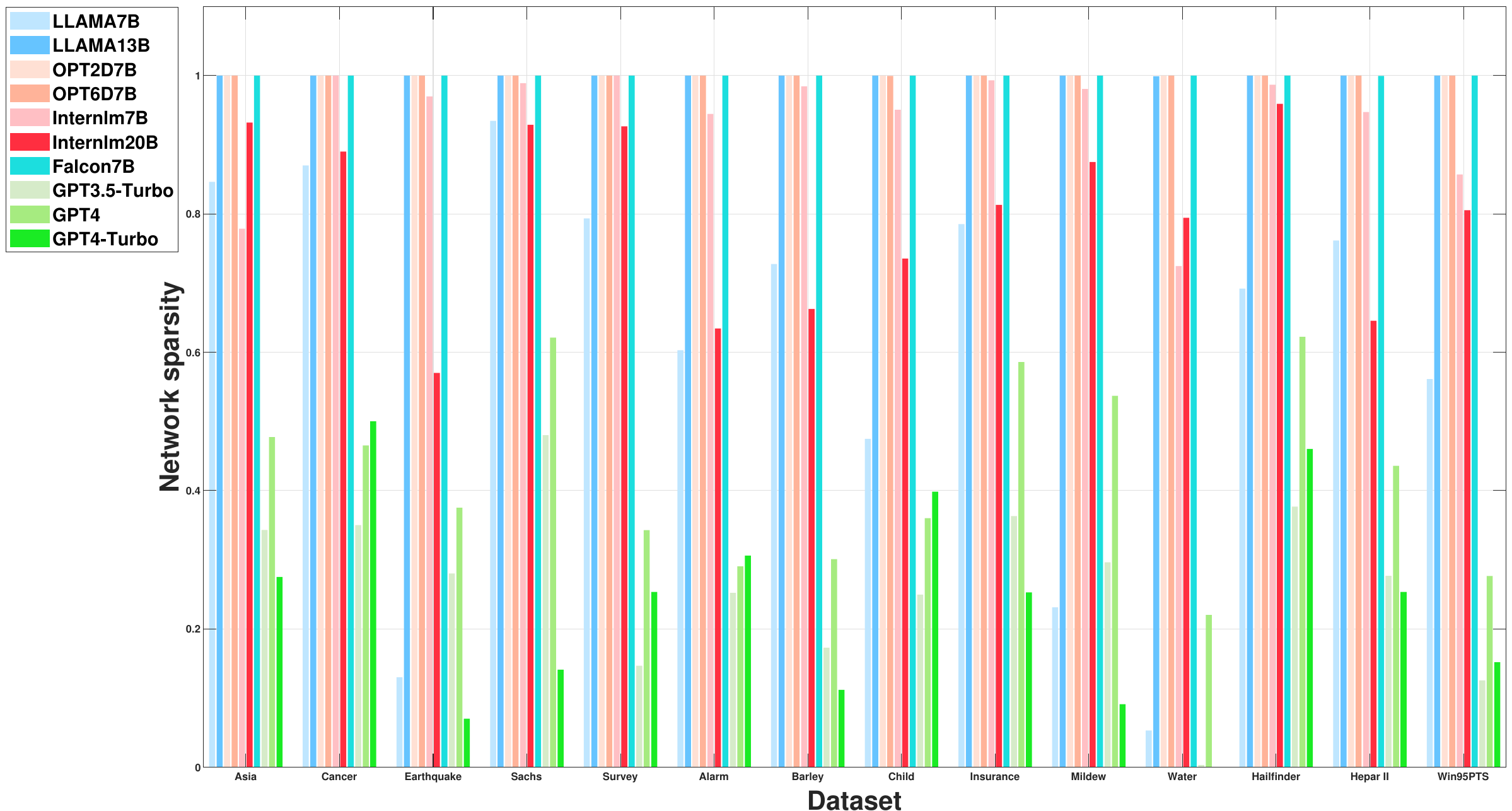}}\caption{Network sparsity of causality identification for prompt format ``variable {name} + background knowledge".} 
	\label{fig:Kon_cas_sparsity} 
\end{figure}

We provide network sparsity of causality identification for prompt format ``variable name + background knowledge" in Fig.~\ref{fig:Kon_cas_sparsity}. From Fig.~\ref{fig:Kon_cas_sparsity}, we find that LLMs with lower SHD and SID typically maintain sparser network structures. Closed-source LLMs' overall network sparsity remains below 0.5, while Falcon series LLMs, which show better performance on large scale causal datasets, also exhibit overall network sparsity that does not exceed 0.4. 

\clearpage
\subsection{Performance Comparison Between the Prompt Formats ``variable names" and ``variable names + background knowledge"}\label{app:kno_cas_comparion}
\subsubsection{Comparison of F1 score between prompt format ``variable names" and ``variable names with background knowledge"}\label{app:kno_cas_f1_v1}
In this subsubsection, we provide F1 score for all LLMs for the prompt formats ``variable name" and ``variable name + background knowledge" on each dataset.
\begin{figure}[ht]
	\centering
	\subfigure{
		\includegraphics[width=0.8\linewidth]{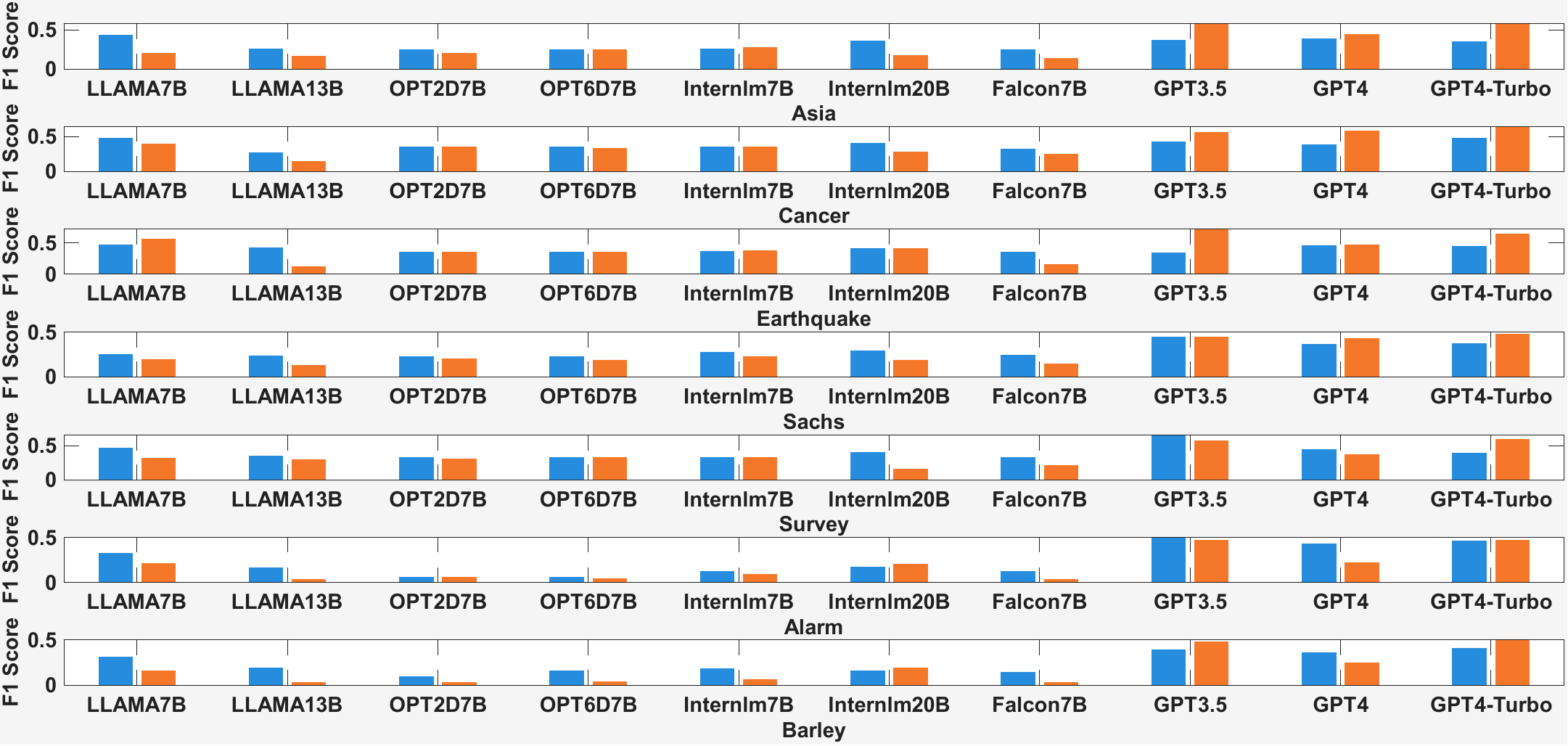}
		\label{exp:kno_cas_f1_v1}
	}
	\subfigure{
		\includegraphics[width=0.8\linewidth]{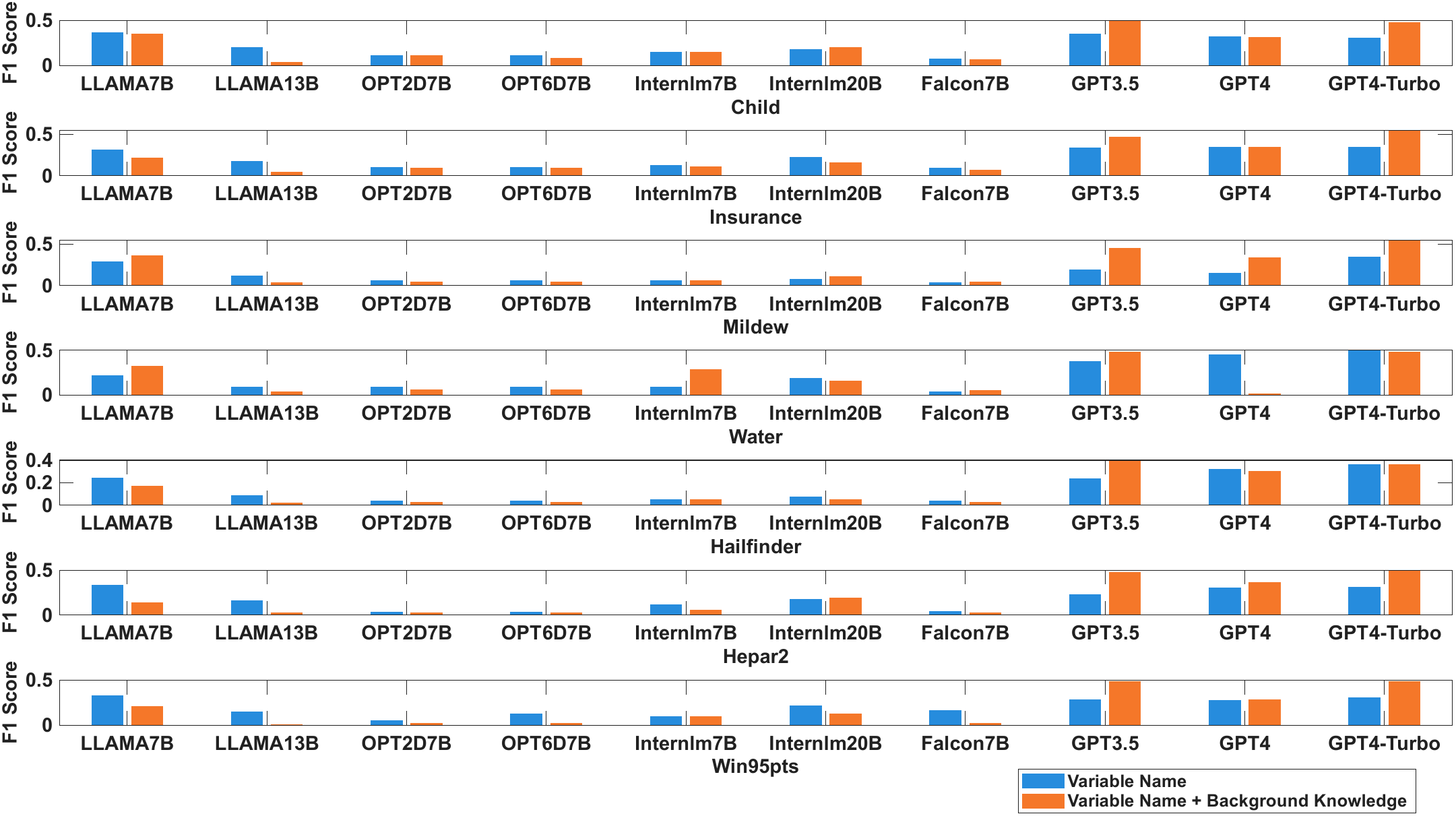}
		\label{exp:kno_cas_f1_v2}
	}
	\caption{{F1 score for prompt formats ``variable name'' and ``variable name + background knowledge".}}
	\label{exp:kno_cas_f1}
\end{figure}

\clearpage
\subsubsection{Comparison of SHD between prompt format ``variable names" and ``variable names with background knowledge"}\label{app:kno_cas_SHD_comparison}
\begin{figure}[ht]
	\centering
	\subfigure{
		\includegraphics[width=0.8 \linewidth]{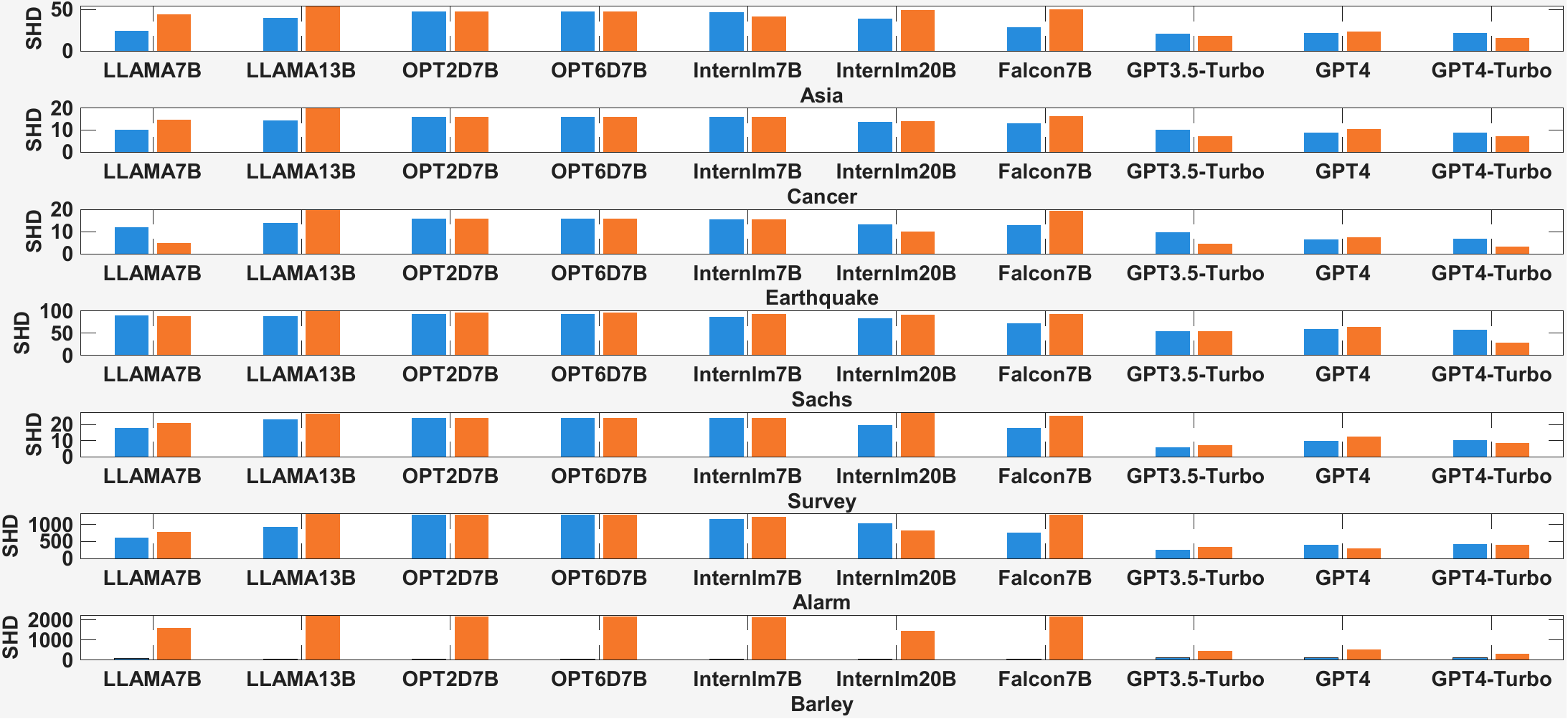}
		\label{exp:kno_cas_SHD_v1}
	}
	
	\subfigure{
		\includegraphics[width=0.8\linewidth]{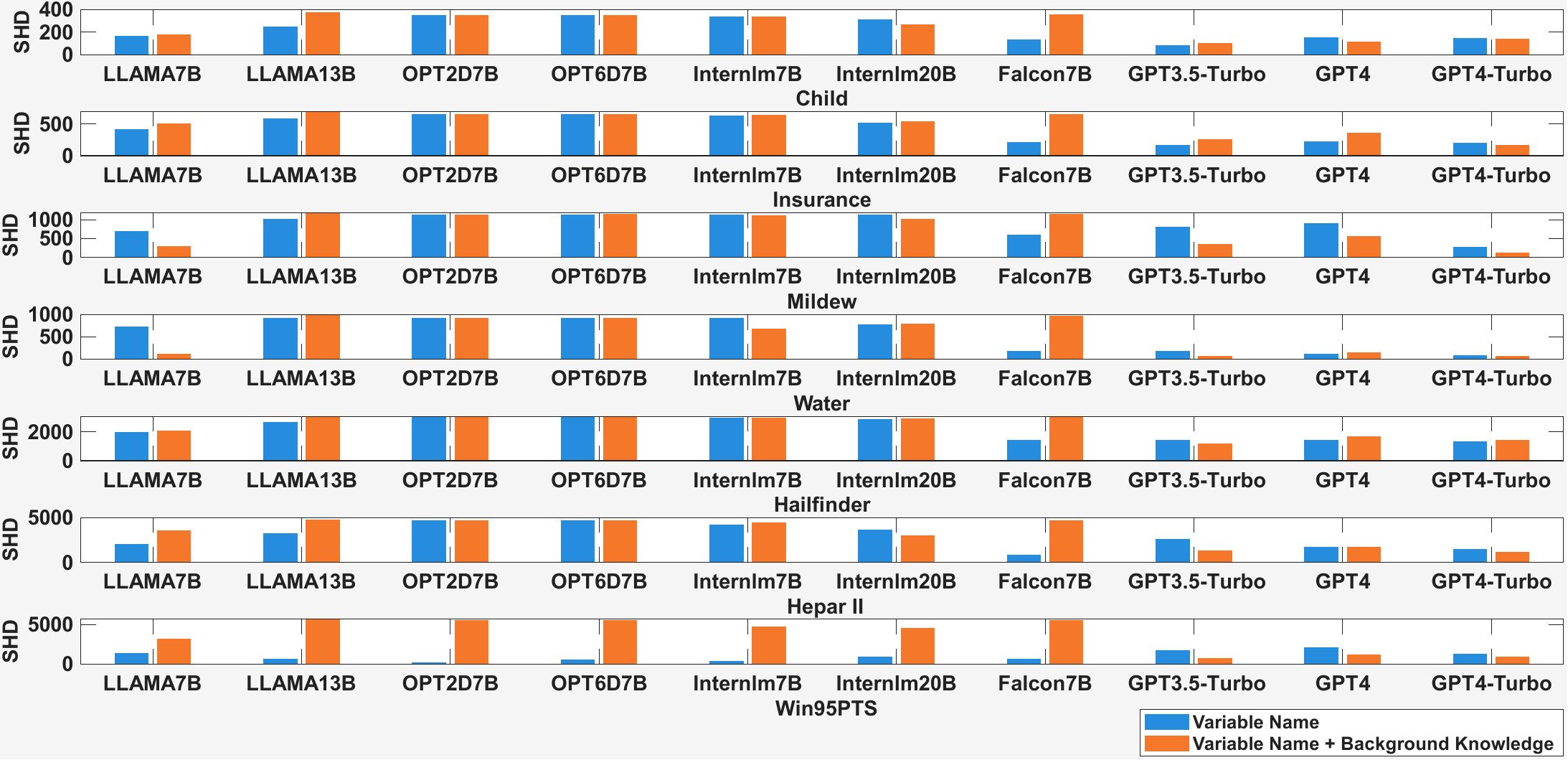}
		\label{exp:kno_cas_SHD_v2}
	}
	\caption{{SHD for variable names and variable names with background knowledge.}}
	\label{exp:kno_cas_SHD}
\end{figure}

We provide SHD for all LLMs for the prompt formats ``variable name" and ``variable name + background knowledge" on each dataset, which are depicted in Figs. \ref{exp:kno_cas_f1} and \ref{exp:kno_cas_SHD}. 
From Fig.~\ref{exp:kno_cas_SHD}, it can be seen that although most open-source LLMs enjoy enhancements on some datasets, such as on Earthquake and Water,  their SHD is generally lower
when using the prompt format ``variable name + background knowledge".
For closed-source LLMs, there is a significant improvement in SHD across all causal datasets, particularly on medium and large scale datasets.

\clearpage
\subsection{Detailed Evaluation Results and Analysis of All Four Prompt Formats}\label{app:diff_prompt}

We evaluate the performance of LLMs using variable names combined with training data as input prompts, and the detailed prompts are provided in Appendix \ref{app:var_sd}. Similar to traditional causal learning algorithms, we utilize training data with 500, 1000 or 1500 observed samples for each node. To conduct this experiment, we first attempt to input training data directly into LLMs. Unfortunately, all open-source LLMs fail to understand the training data. Additionally, closed-source LLMs could not process all training data simultaneously due to token count limitations. To address this issue, we focus on 500 observed samples for each node and divide them into five segments, each with 100 observed samples. Specifically, we first inform the LLM: ``There are a total of five parts matrices, together forming an input sample of the causality learning network named Alarm''. The LLM would prompt us to enter the corresponding part number and sample data. After we input each part's number and its sample data, the LLM would reassemble the split sample data according to the part numbers and sample data, thereby recognizing the training data. By interacting with the LLM, we could input training data into it. Although this method enables GPT series LLMs to understand training data, all open-source LLMs fail. Therefore, we only discuss the performance of GPT series LLMs with variable names and training data, and the experimental results are shown in Table \ref{tab:Causal+data+know}. From it, we observe that using variable names with both background knowledge and training data as prompts enhances the LLM's capability to identify causality. Its performance is significantly better than the other three prompt formats. 
This means, embedding both background knowledge and training data into variable names can promote the LLM's ability to identify causality, but this requires clear variable names, indicating that LLMs understand causal relationships through semantic associations of clear entities rather than from contextual information or numerical distributions. Furthermore, we find that training data slightly outperforms background knowledge in enhancing the performance of LLMs, and this method brings the best performance among all LLMs. 
\begin{table}[htbp]
	\centering
	\caption{Performance for different prompt formats}
	\label{tab:Causal+data+know}%
	\scalebox{0.60}{\small
		\begin{tabular}{clcccccccc}
			\toprule
			\multirow{1.5}[4]{*}{Causal Network} & \multirow{1.5}[4]{*}{LLMs} & \multicolumn{2}{c}{Var.} & \multicolumn{2}{c}{Var. + B. K.} & \multicolumn{2}{c}{Var. + T. D.} & \multicolumn{2}{c}{Var. + B. K. + T. D.} \\
			\cmidrule{3-10}          &       & F1 score & SHD   & F1 score & SHD   & F1 score & SHD   & F1 score & SHD \\
			\midrule
			\multirow{2}[6]{*}{Asia} & GPT3.5-Turbo & 0.3754  & 21.00  & 0.5787  & 18.40  & 0.5148  & 24.20  & 0.5385  & 18.00  \\
			\cmidrule{2-10}          & GPT4  & 0.3913  & 20.37  & 0.4497  & 23.80  & 0.5280  & 27.20  & 0.6265  & 15.30  \\
			\cmidrule{2-10}          & GPT4-Turbo & 0.3498  & 22.00  & 0.5794  & 18.20  & 0.5988  & 21.80  & 0.7804  & 8.00  \\
			\midrule
			\multirow{2}[6]{*}{Cancer} & GPT3.5-Turbo & 0.4331  & 10.20  & 0.5673  & 7.40  & 0.5449  & 8.60  & 0.6250  & 6.00  \\
			\cmidrule{2-10}          & GPT4  & 0.3903  & 11.51  & 0.5834  & 7.30  & 0.6752  & 5.40  & 0.6845  & 5.30  \\
			\cmidrule{2-10}          & GPT4-Turbo & 0.4788  & 8.80  & 0.6468  & 6.80  & 0.6712  & 6.40  & 0.7024  & 4.00  \\
			\midrule
			\multirow{2}[6]{*}{Earthquake} & {GPT3.5-Turbo} & 0.3478  & 9.60  & 0.7202  & 4.40  & 0.7813  & 3.60  & 0.5921  & 7.00  \\
			\cmidrule{2-10}          & GPT4  & 0.4625  & 6.37  & 0.4789  & 7.37  & 0.7141  & 6.00  & 0.7855  & 3.50  \\
			\cmidrule{2-10}          & GPT4-Turbo & 0.4473  & 6.80  & 0.6485  & 3.40  & 0.7319  & 5.60  & 0.9169  & 1.00  \\
			\midrule
			\multirow{2}[6]{*}{Sachs} & GPT3.5-Turbo & 0.4465  & 54.00  & 0.4450  & 54.60  & -     & -     & -     & - \\
			\cmidrule{2-10}          & GPT4  & 0.3706  & 58.17  & 0.4299  & 64.46  & 0.4763  & 52.36  & 0.4648  & 52.99  \\
			\cmidrule{2-10}          & GPT4-Turbo & 0.3736  & 58.00  & 0.4836  & 28.50  & 0.4663  & 52.92  & 0.4543  & 53.58  \\
			\midrule
			\multirow{2}[6]{*}{Survey} & GPT3.5-Turbo & 0.6586  & 6.00  & 0.5788  & 7.20  & -     & -     & -     & - \\
			\cmidrule{2-10}          & GPT4  & 0.4458  & 9.60  & 0.3727  & 12.43  & 0.4126  & 10.16  & 0.5125  & 8.47  \\
			\cmidrule{2-10}          & GPT4-Turbo & 0.3987  & 10.40  & 0.6024  & 6.50  & 0.5422  & 7.97  & 0.6538  & 6.08  \\
			\midrule
			\multirow{3}[6]{*}{Alarm} & GPT3.5-Turbo & 0.5045  & 264.40  & 0.4740  & 347.00  & -     & -     & -     & - \\
			\cmidrule{2-10}          & GPT4  & 0.4352  & 539.03  & 0.2222  & 295.64  & 0.4951  & 301.41  & 0.5126  & 232.10  \\
			\cmidrule{2-10}          & GPT4-Turbo & 0.4654  & 419.20  & 0.4705  & 406.40  & 0.5126  & 232.10  & 0.5489  & 87.97  \\
			\midrule
			\multirow{2}[6]{*}{Barley} & GPT3.5-Turbo & 0.3951  & 118.50  & 0.4862  & 437.20  & -     & -     & -     & - \\
			\cmidrule{2-10}          & GPT4  & 0.3658  & 108.80  & 0.2482  & 543.35  & 0.4790  & 146.22  & 0.5792  & 179.33  \\
			\cmidrule{2-10}          & GPT4-Turbo & 0.4153  & 125.16  & 0.4979  & 312.40  & 0.4765  & 145.41  & 0.5630  & 173.98  \\
			\midrule
			\multirow{2}[6]{*}{Child} & GPT3.5-Turbo & 0.3529  & 82.00  & 0.5018  & 62.10  & -     & -     & -     & - \\
			\cmidrule{2-10}          & GPT4  & 0.3252  & 124.47  & 0.3137  & 140.50  & 0.4895  & 79.50  & 0.5760  & 45.40  \\
			\cmidrule{2-10}          & GPT4-Turbo & 0.3115  & 145.60  & 0.4782  & 83.60  & 0.4766  & 84.10  & 0.6029  & 42.10  \\
			\midrule
			\multirow{2}[6]{*}{Insurance} & GPT3.5-Turbo & 0.3381  & 170.80  & 0.4709  & 256.00  & -     & -     & -     & - \\
			\cmidrule{2-10}          & GPT4  & 0.3462  & 209.10  & 0.3497  & 207.60  & 0.3798  & 201.60  & 0.5592  & 174.20  \\
			\cmidrule{2-10}          & GPT4-Turbo & 0.3469  & 208.80  & 0.5505  & 175.00  & 0.5849  & 167.40  & 0.6892  & 152.40  \\
			\midrule
			\multirow{2}[6]{*}{Mildew} & GPT3.5-Turbo & 0.1917  & 589.60  & 0.4540  & 361.60  & -     & -     & -     & - \\
			\cmidrule{2-10}          & GPT4  & 0.1564  & 627.70  & 0.3374  & 293.50  & 0.5165  & 163.70  & 0.5989  & 114.30  \\
			\cmidrule{2-10}          & GPT4-Turbo & 0.3489  & 283.00  & 0.5452  & 136.40  & 0.5255  & 149.50  & 0.6159  & 102.90  \\
			\midrule
			\multirow{2}[6]{*}{Water} & GPT3.5-Turbo & 0.3821  & 178.60  & 0.4855  & 68.00  & -     & -     & -     & - \\
			\cmidrule{2-10}          & GPT4  & 0.4514  & 122.61  & 0.4648  & 117.54  & 0.4325  & 147.60  & 0.4889  & 67.10  \\
			\cmidrule{2-10}          & GPT4-Turbo & 0.4994  & 63.80  & 0.4834  & 68.60  & 0.4119  & 159.80  & 0.4749  & 70.40  \\
			\midrule
			\multirow{2}[6]{*}{Hailfinder} & GPT3.5-Turbo & 0.2366  & 2152.30  & 0.3985  & 1178.80  & -     & -     & -     & - \\
			\cmidrule{2-10}          & GPT4  & 0.3214  & 1465.90  & 0.3025  & 1706.50  & 0.3059  & 1712.40  & 0.3148  & 1484.90  \\
			\cmidrule{2-10}          & GPT4-Turbo & 0.3649  & 1321.60  & 0.3620  & 1420.60  & 0.2959  & 1865.20  & 0.3750  & 1224.50  \\
			\midrule
			\multirow{2}[6]{*}{Hepar II} & GPT3.5-Turbo & 0.2308  & 2577.20  & 0.4797  & 1299.00  & -     & -     & -     & - \\
			\cmidrule{2-10}          & GPT4  & 0.3081  & 1572.50  & 0.3660  & 1394.90  & 0.3755  & 1347.50  & 0.3594  & 1412.80  \\
			\cmidrule{2-10}          & GPT4-Turbo & 0.3135  & 1502.60  & 0.5023  & 1172.20  & 0.4219  & 1275.40  & 0.4196  & 1286.40  \\
			\midrule
			\multirow{2}[6]{*}{Win95PTS} & GPT3.5-Turbo & 0.2836  & 1795.40  & 0.4856  & 932.60  & -     & -     & -     & - \\
			\cmidrule{2-10}          & GPT4  & 0.2811  & 1844.71  & 0.2835  & 1797.00  & 0.2495  & 2145.60  & 0.3252  & 1211.20  \\
			\cmidrule{2-10}          & GPT4-Turbo & 0.3077  & 1329.40  & 0.4912  & 912.40  & 0.3250  & 994.07  & 0.4986  & 909.30  \\
			\bottomrule
	\end{tabular}}
	
\end{table}

\clearpage
\section{Detailed Evaluation Results on Other Influencing Factors}\label{app:add_exp_other}

\subsection{Detailed Evaluation Results on Data Identification for the Correlation Identification Task} \label{app:fine-grained}
We examine the impact of data granularity on the performance of LLMs, with results in Table~\ref{tab:fg}. In this context, ``understanding" refers not to producing correct answers, but to correctly parsing the question, recognizing the input, and subsequently engaging in causal learning.
Overall, most LLMs can process data from coarse to fine granularity, except for BERT series and OPT-1D3B, which fail to recognize prompts and generate invalid outputs. When dealing with training data, substantial limitations of LLMs are observed. Open-source LLMs with fewer than 100 billion parameters fail to interpret such data, while GPT-3.5-Turbo can only handle inputs up to $7 \times 500$. GPT-4-Turbo can process table-like or matrix inputs up to $78 \times 500$, but its performance still declines as data granularity increases. Notably, none of the evaluated models can handle training data extended to $1,000$ dimensions.
These findings indicate that existing LLMs are adequate for recognizing coarse and moderately fine data, yet their ability to process highly fine-grained inputs remains limited.

\begin{table}[htbp]
	\centering
	\caption{Fine-grained Analysis}
	
	\label{tab:fg}
	\scalebox{0.65}{
		\begin{tabular}{lcccccc}
			\toprule
			\multicolumn{1}{c}{\multirow{1.5}[4]{*}{LLMs}} & \multirow{1.5}[4]{*}{Correlation} & \multirow{1.5}[4]{*}{Causal skeleton} & \multicolumn{4}{c}{Causality} \\
			\cmidrule{4-7}          &       &       & Var.  & Var. + B. K. & Var. + T. D. & Var. + B. K. + T. D. \\
			\midrule
			BERT-large & $\times$ & $\times$ & $\times$ & ×     & $\times$ & $\times$ \\
			RoBERTa-large & $\times$ & $\times$ & $\times$ & ×     & $\times$ & $\times$ \\
			DeBERTa-v3-large & $\times$ & $\times$ & $\times$ & ×     & $\times$ & $\times$ \\
			DistilBERT-mnli & $\times$ & $\times$ & $\times$ & ×     & $\times$ & $\times$ \\
			
			LLAMA7B & $\surd$      & $\surd$      & $\surd$      & $\surd$      & $\times$ & $\times$ \\
			LLAMA13B & $\surd$      & $\surd$      & $\surd$      & $\surd$      & $\times$ & $\times$ \\
			LLAMA30B & $\surd$      & $\surd$      & $\surd$      & $\surd$      & $\times$ & $\times$ \\
			OPT1D3B & $\times$ & $\times$ & $\times$ & ×     & $\times$ & $\times$ \\
			OPT2D7B & $\times$ & $\times$ & $\times$ & ×     & $\times$ & $\times$ \\
			OPT6D7B & $\surd$      & $\surd$      & $\surd$      & $\surd$      & $\times$ & $\times$ \\
			OPT60B & $\surd$      & $\surd$      & $\surd$      & $\surd$      & $\times$ & $\times$ \\
			Internlm7B & $\surd$      & $\surd$      & $\surd$      & $\surd$      & $\times$ & $\times$ \\
			Internlm20B & $\surd$      & $\surd$      & $\surd$      & $\surd$      & $\times$ & $\times$ \\
			Falcon7B & $\surd$      & $\surd$      & $\surd$      & $\surd$      & $\times$ & $\times$ \\
			Falcon40B & $\surd$      & $\surd$      & $\surd$      & $\surd$      & $\times$ & $\times$ \\
			GPT3.5-Turbo & $\surd$      & $\surd$      & $\surd$      & $\surd$      & $\surd$      & $\times$ \\
			GPT4  & $\surd$      & $\surd$      & $\surd$      & $\surd$      & $\surd$      & $\surd$  \\
			GPT4-Turbo & $\surd$      & $\surd$      & $\surd$      & $\surd$      & $\surd$      & $\surd$  \\
			\bottomrule
	\end{tabular}}
\end{table}

\subsection{Detailed Evaluation Results on Causal Network Analysis}\label{app:causal_network}
In this subsection, we analyze two structural metrics in causal learning: out-degree and in-degree \cite{cooper1990computational}.
	Out-degree denotes the number of directed edges emanating from a node, indicating how many other nodes it influences, while in-degree denotes the number of directed edges pointing to a node, reflecting how many other nodes directly influence it.
	These metrics are critical for identifying key nodes of influence and dependency, thereby revealing the structure and dynamics of causal relationships~\cite{Guo2024Causal}.
	As shown in Table~\ref{tab:degree}, the DAGs generated by LLMs remain inferior to those by traditional causal learning methods.
	Across multiple datasets, the DAGs generated by LLMs exhibit larger average out-degrees and in-degrees, suggesting a tendency to produce overly dense structures.
	Among LLMs, closed-source models yield degree values closer to those of classical methods (see the last column ``Average degree" in Table~\ref{tab:degree}) than open-source models, yet both categories still fall short of matching the performance of traditional approaches.

	\begin{table}[ht]
		\centering
		\caption{Result about average degree for all LLMs}
		\scalebox{0.7}{
			\begin{tabular}{lccccccc}
				\toprule
				\multirow{1.5}[4]{*}{Dataset} & \multicolumn{2}{c}{Open-source LLMs} & \multicolumn{2}{c}{Closed-source LLMs} & \multicolumn{2}{c}{All LLMs} & \multirow{1.5}[4]{*}{Average degree} \\
				\cmidrule{2-7}          & Degree & Accuracy & Degree & Accuracy & Degree & Accuracy &  \\
				\midrule
				Asia  & 5.10  & 34.65  & 1.90  & 66.15  & 3.82  & 41.92  & 2.00  \\
				\midrule
				Cancer & 3.27  & 42.55  & 1.14  & 62.93  & 2.42  & 47.25  & 1.60  \\
				\midrule
				Earthquake & 3.61  & 45.11  & 0.28  & 69.33  & 2.28  & 50.70  & 1.60  \\
				\midrule
				Sachs & 9.75  & 27.32  & 5.49  & 52.62  & 8.05  & 33.16  & 3.09  \\
				\midrule
				Survey & 4.66  & 39.82  & 1.83  & 75.74  & 3.13  & 48.11  & 2.00  \\
				\midrule
				Alarm & 28.02  & 24.47  & 7.99  & 73.62  & 20.01  & 35.81  & 2.49  \\
				\midrule
				Child & 14.43  & 31.76  & 3.69  & 65.41  & 10.13  & 39.53  & 2.50  \\
				\midrule
				Barley & 25.52  & 25.25  & 4.46  & 68.30  & 18.65  & 35.18  & 3.50  \\
				\midrule
				Insurance & 23.03  & 22.39  & 3.90  & 72.16  & 15.38  & 33.87  & 3.85  \\
				\midrule
				Mildew & 32.74  & 15.86  & 1.43  & 45.35  & 20.21  & 22.67  & 2.63  \\
				\midrule
				Water & 29.75  & 17.01  & 4.94  & 72.51  & 19.83  & 29.82  & 4.12  \\
				\midrule
				Hailfinder & 51.82  & 10.53  & 22.53  & 61.61  & 40.10  & 22.32  & 2.36  \\
				\midrule
				Hepar II & 47.65  & 25.19  & 25.01  & 60.66  & 38.59  & 33.37  & 3.51  \\
				\midrule
				Win95PTS & 23.62  & 30.83  & 15.30  & 69.55  & 21.12  & 39.77  & 2.95  \\
				\bottomrule
		\end{tabular}}%
	\label{tab:degree}%
\end{table}%

\clearpage
\subsection{Detailed Accuracy Comparison for Prompts by Variable Names and Modified Variable Names}\label{app:acc_modified_var}
\begin{figure}[ht]
	\centering
	\subfigure{
		\includegraphics[width=0.7\linewidth]{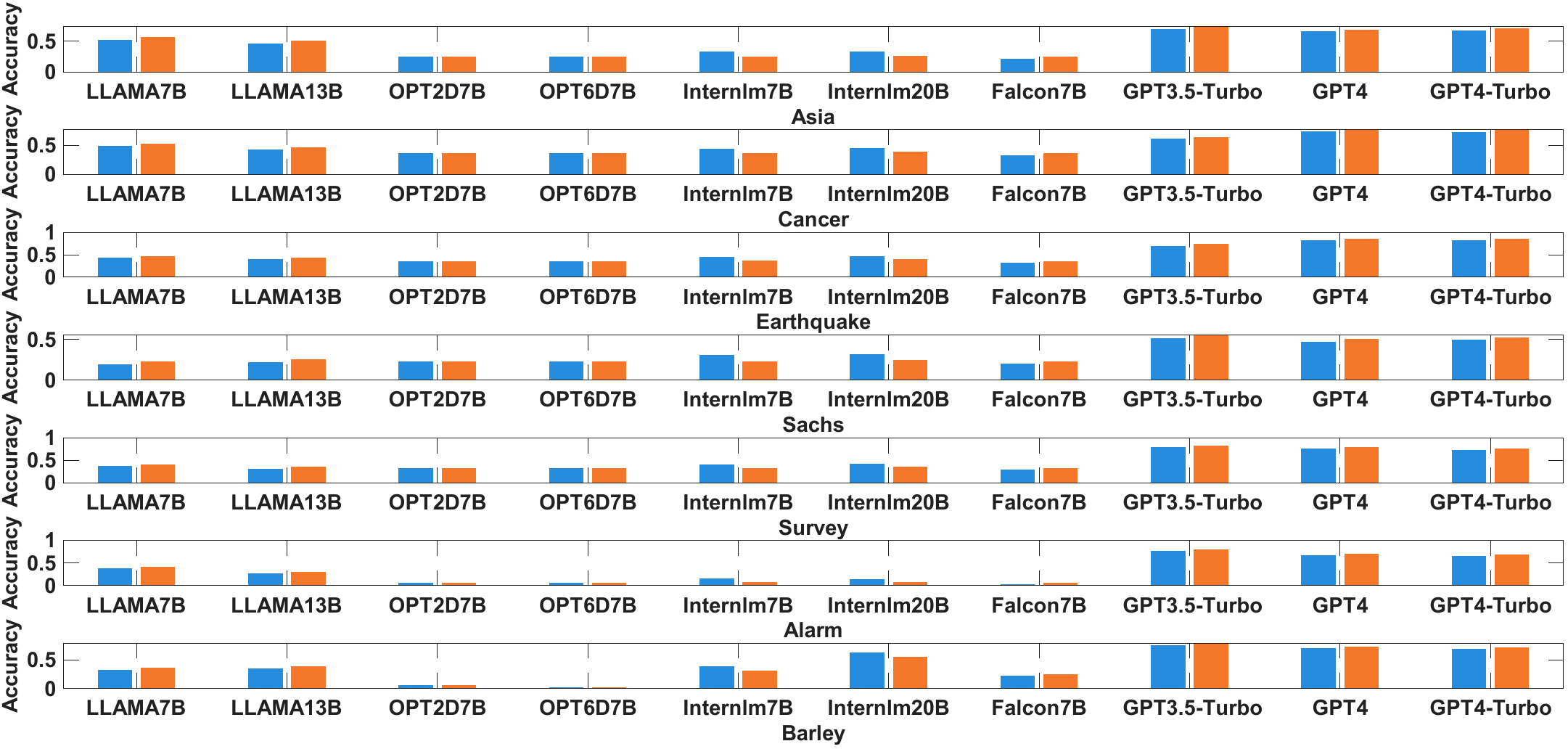}
		\label{fig:cas_var_acc_v1}
	}
	
	\subfigure{
		\includegraphics[width=0.7\linewidth]{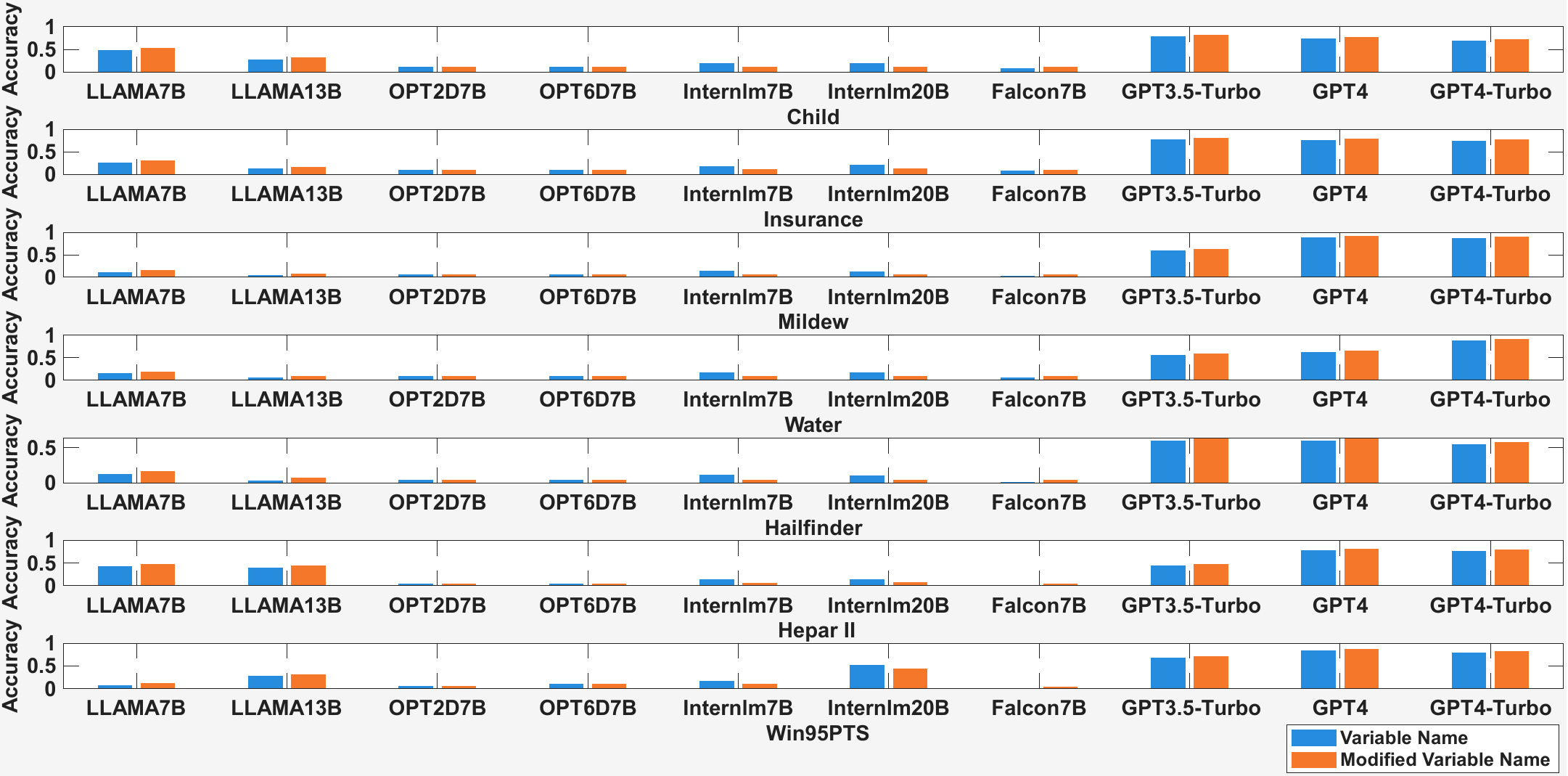}
		\label{fig:cas_var_acc_v2}
	}
	\caption{{Accuracy for prompts using variable names and modified ones.}}
	\label{fig:cas_var_acc}
\end{figure}

We first search Wikipedia and the websites about causal learning community such as Bnlearn for the actual meanings of each variable name. The experimental results of the accuracy metric, as shown in Fig. \ref{fig:cas_var_acc}, indicate that after modifying variable names, there is an improvement in the accuracy across almost all LLMs. This suggests that more detailed variable names can enhance LLMs' ability to recognize causality. However, we also notice that on some datasets, such as Water and Mildew, the performance difference between more detailed variable names and the original ones is negligible. This interesting phenomenon indicates that while more detailed variable names provide more extensive background information, which is expected to improve performance, this is not always the case. 

\clearpage
\subsection{Detailed Evaluation Results on Sentence Paraphrasing for the Correlation Identification Task} \label{app:sen_par_cor}
We check the impact of different expressions of the prompts upon the performance of LLMs.
We perform experiments with two evaluation tasks, i.e. correlation and causality identification tasks respectively by utilizing five prompt sentences.
{In particular, combined with previous work, we designed five different prompts, including questions starting with ``Are",  with ``Do" and with ``Is there" structure.}
The experimental results about causal identification task are provided in Table \ref{tab:re_relation}.
From Table \ref{tab:re_relation},  we can see that across the {five types of prompt sentences}, the performance variation among all LLMs is minimal, with an average F1 score ranging between 0.28 and 0.30. Type 3 achieves the highest F1 score (i.e., 0.3089) and the highest accuracy (i.e., 0.3308), followed by Type 1 with an F1 score of 0.2907, while the F1 scores for the other prompts hover around 0.28. Looking at specific LLMs, we find that the performance difference across the five prompt types is also minimal for open-source LLMs. For closed-source LLMs, however, there is a more significant performance variation across the five prompts, with the highest F1 score for Type 3 reaching 0.49 and the lowest F1 score (i.e., 0.41) for Type 5. Overall, the capability of open-source LLMs to recognize correlations in every prompt type is significantly inferior to that of closed-source LLMs.
\begin{table}[h]
	\centering
	\caption{Sentence paraphrase for correlation identification task}
	\label{tab:re_relation}	\scalebox{1.0}{
		\begin{tabular}{llrr}
			\toprule
			\multicolumn{1}{l}{Correlation identification task} & \multicolumn{1}{c}{LLMs} & \multicolumn{1}{c}{F1} & \multicolumn{1}{c}{Accuracy} \\
			\midrule
			\multirow{2}[6]{*}{Type 1: Are \textit{Var. A} and \textit{Var. B} related?} & Open-Source LLMs & 0.2562  & 0.2646  \\
			\cmidrule{2-4}          & Closed-Source LLMs & 0.4338  & 0.4760  \\
			\cmidrule{2-4}          & All  LLMs & 0.2907  & 0.3057  \\
			\midrule
			\multirow{2}[6]{*}{Type 2: Are \textit{Var. A} and \textit{Var. B} correlated?} & Open-Source LLMs & 0.2525  & 0.2599  \\
			\cmidrule{2-4}          & Closed-Source LLMs & 0.4169  & 0.4654  \\
			\cmidrule{2-4}          & All  LLMs & 0.2845  & 0.2999  \\
			\midrule
			\multirow{2}[6]{*}{Type 3: Is there a correlation between \textit{Var. A} and \textit{Var. B}? } & Open-Source LLMs & 0.2647  & 0.2813  \\
			\cmidrule{2-4}          & Closed-Source LLMs& 0.4915  & 0.5729  \\
			\cmidrule{2-4}          & All  LLMs & 0.3089  & 0.3380  \\
			\midrule
			\multirow{2}[6]{*}{Type 4: Is there a relation between \textit{Var. A} and \textit{Var. B}?} & Open-Source LLMs & 0.2535  & 0.2602  \\
			\cmidrule{2-4}          & Closed-Source LLMs & 0.4336  & 0.4758  \\
			\cmidrule{2-4}          & All  LLMs & 0.2885  & 0.3022  \\
			\midrule
			\multirow{2}[6]{*}{Type 5: Do \textit{Var. A} and \textit{Var. B} have a connection? } & Open-Source LLMs & 0.2574  & 0.2673  \\
			\cmidrule{2-4}          & Closed-Source LLMs & 0.4140  & 0.4442  \\
			\cmidrule{2-4}          & All  LLMs & 0.2879  & 0.3017  \\
			\bottomrule
	\end{tabular}}%
	
\end{table}%

\subsection{Detailed Results for Few-Shot In-Context Learning Evaluation}\label{app:fewshot}
\begin{figure}[htb] 
	\center{\includegraphics[width=0.5\linewidth]  {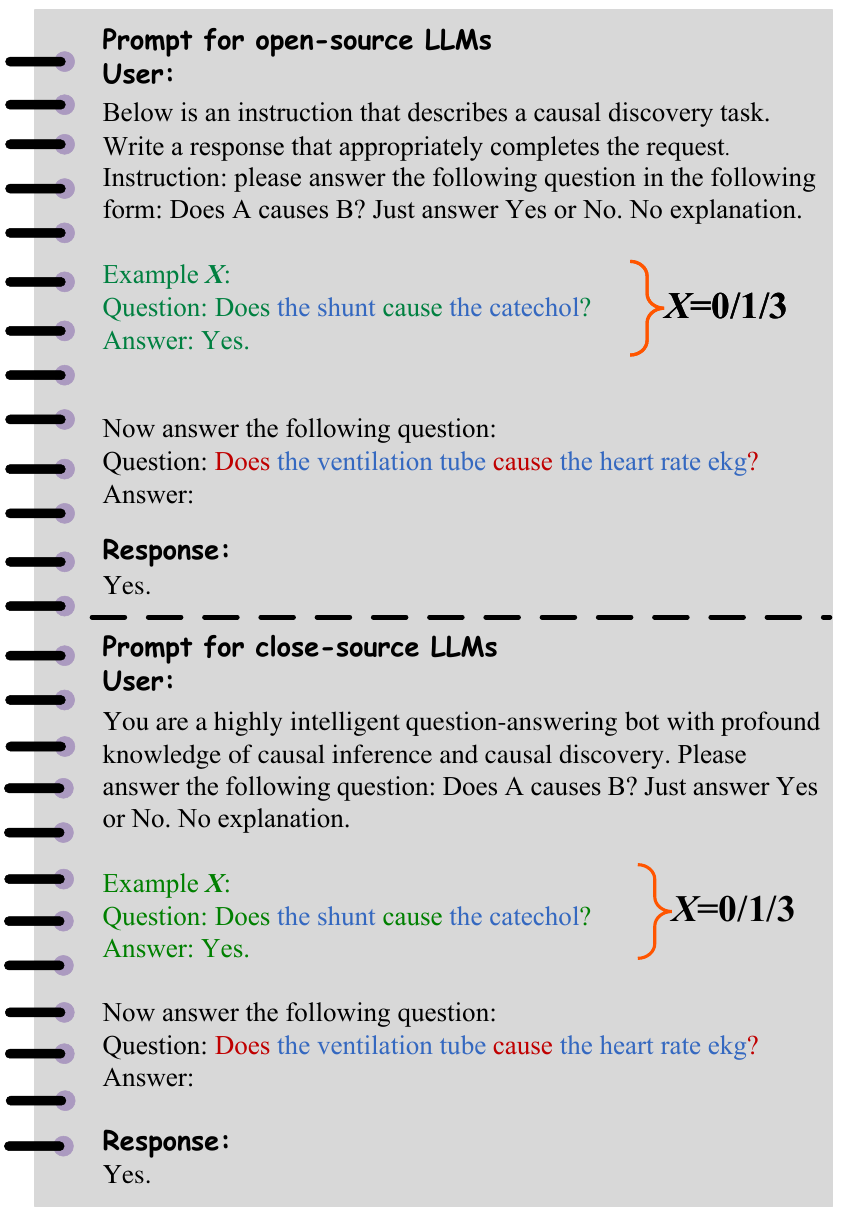}}\caption{Prompt template for causality identification tasks under 0/1/3-shot in-context learning, where $X=0,1,3$ denotes the number of examples used in the prompt.} 
	\label{fig:prompt_few_shot} 
\end{figure}

{To further examine whether a small number of task demonstrations can improve the causal learning ability of LLMs, we conduct additional few-shot in-context learning (ICL) experiments on the correlation identification, causal skeleton identification, and causality identification tasks. 
	\subsubsection{Experimental Setup}
	The design of this experiment follows the setting in \cite{chen2024causal}. The corresponding prompts for causality identification tasks are shown in Fig.~\ref{fig:prompt_few_shot}, while those for correlation identification and causal skeleton construction tasks are designed based on the same principle.
	Unlike the 0-shot setting, where the model answers the target question directly without any demonstrations, the few-shot setting adds a small number of labeled examples before the target question to provide task guidance. In this work, we consider two few-shot settings: 1-shot and 3-shot.
	To avoid answer leakage, the in-context examples are not taken from the same test graph as the target sample. Instead, all demonstrations are constructed from variable pairs sampled from different causal networks. This design prevents the model from accessing the target graph structure through the demonstrations, while still providing useful guidance on the question format and the required binary output. As in the zero-shot setting, the model is required to answer each question using only ``Yes'' or ``No,'' and F1 score and accuracy are used as the evaluation metrics. }

\subsubsection{Experimental analysis}
{Table~\ref{tab:ICL_F1} and Table~\ref{tab:accuracy_ICL} report the few-shot results in terms of F1 score and Accuracy, respectively.
	Overall, few-shot ICL improves LLMs' causal learning performance across all three evaluation tasks, though the magnitude and stability of the gains vary by task and dataset. Among the three evaluation tasks, the largest and most consistent improvements are observed on the causal skeleton identification task. For example, under the 3-shot setting, the F1 score increases from 0.3445 to 0.4477 on Alarm, from 0.3727 to 0.4806 on Child, from 0.3850 to 0.4916 on Barley, and from 0.3711 to 0.4757 on Insurance. These results suggest that in-context examples are especially helpful for aligning the model with the notion of direct structural dependence and for reducing systematic bias that may arise in the zero-shot setting.}

{The causality identification task also benefits from few-shot prompting, although the improvement is generally smaller than that on the causal skeleton identification task. On relatively large graphs, the gains are still clear. For example, the F1 score increases from 0.2176 to 0.2861 on Alarm, from 0.2057 to 0.2769 on Child, and from 0.2400 to 0.3188 on Barley. This indicates that in-context examples can help the model better understand the meaning of direct causality. However, because this task requires more accurate directional and structural reasoning, the gain from few-shot prompting is more limited.}

{In comparison, the correlation identification task shows the smallest and least stable improvement. On some datasets, the 1-shot setting even leads to slight performance fluctuations, while the 3-shot setting usually recovers and surpasses the zero-shot baseline. This suggests that when only one demonstration is provided, the model may be more sensitive to example-specific noise or a mismatch between the demonstration and the target question. Increasing the number of demonstrations to three generally provides more reliable guidance and improves output calibration.}

{Overall, these results show that few-shot ICL can consistently improve LLM performance on causal question answering based on causal networks, but its effect depends on the task type. The gain is most obvious on the causal skeleton identification task, moderate on the causality identification task, and relatively limited on the correlation identification task. In addition, the improvements are usually more noticeable on larger and more complex graphs, where variable meanings are less clear and task boundaries are harder to infer from the zero-shot prompt.}

\begin{table}[htbp]  
	\centering  
	\caption{F1 score for few-shot in-context learning}    
	\small  
	{  
		\begin{tabular}{l|ccc|ccc|ccc}    
			\toprule    \multicolumn{1}{c|}{\multirow{2}[4]{*}{Datasets}} & \multicolumn{3}{c|}{Correlation} & \multicolumn{3}{c|}{Causal Skeleton} & \multicolumn{3}{c}{Causality} \\
			\cmidrule{2-10}          & 0-shot & 1-shot & 3-shot & 0-shot & 1-shot & 3-shot & 0-shot & 1-shot & 3-shot \\    
			\midrule    {Asia}  & 0.4126  & 0.4108  & 0.4351  & 0.3571  & 0.3654  & 0.4022  & 0.3252  & 0.3229  & 0.3627  \\    
			Cancer & 0.5279  & 0.5316  & 0.5538  & 0.3352  & 0.3451  & 0.3796  & 0.3719  & 0.3742  & 0.4121  \\    
			Earthquake & 0.5673  & 0.5609  & 0.5927  & 0.3426  & 0.3567  & 0.3924  & 0.4246  & 0.4201  & 0.4683  \\    
			Sachs & 0.5181  & 0.5229  & 0.5487  & 0.3913  & 0.4102  & 0.4635  & 0.3994  & 0.4088  & 0.4554  \\    
			Survey & 0.3479  & 0.3521  & 0.3708  & 0.2948  & 0.2929  & 0.3366  & 0.3101  & 0.3158  & 0.3467  \\    Alarm & 0.1791  & 0.1716  & 0.2118  & 0.3445  & 0.3659  & 0.4477  & 0.2176  & 0.2334  & 0.2861  \\    
			Barley & 0.2262  & 0.2357  & 0.2599  & 0.3850  & 0.4063  & 0.4916  & 0.2400  & 0.2551  & 0.3188  \\    
			Child & 0.1687  & 0.1784  & 0.2027  & 0.3727  & 0.3981  & 0.4806  & 0.2057  & 0.2196  & 0.2769  \\    
			Insurance & 0.2601  & 0.2528  & 0.2924  & 0.3711  & 0.3935  & 0.4757  & 0.1506  & 0.1659  & 0.2133  \\    Mildew & 0.2379  & 0.2460  & 0.2707  & 0.3833  & 0.4059  & 0.4892  & 0.2524  & 0.2687  & 0.3255  \\    Water & 0.1791  & 0.1742  & 0.2067  & 0.3093  & 0.3188  & 0.3826  & 0.2496  & 0.2469  & 0.3092  \\    
			Hailfinder & 0.1257  & 0.1318  & 0.1562  & 0.2852  & 0.3007  & 0.3794  & 0.1332  & 0.1440  & 0.1876  \\    
			Hepar II & 0.1592  & 0.1665  & 0.1901  & 0.3851  & 0.4087  & 0.4941  & 0.1764  & 0.1889  & 0.2429  \\    
			Win95PTS & 0.1541  & 0.1489  & 0.1841  & 0.3075  & 0.3242  & 0.3973  & 0.2131  & 0.2250  & 0.2804  \\    \bottomrule    \end{tabular}}%
	\label{tab:ICL_F1}%
\end{table}%

\begin{table}[htbp]  
	\centering  
	\caption{Accuracy for few-shot in-context learning}
	\small
	{    
		\begin{tabular}{l|ccc|ccc|ccc}    
			\toprule    \multicolumn{1}{c|}{\multirow{2}[4]{*}{Datasets}} & \multicolumn{3}{c|}{Correlation} & \multicolumn{3}{c|}{Causal Skeleton} & \multicolumn{3}{c}{Causality} \\\cmidrule{2-10}          & 0-shot & 1-shot & 3-shot & 0-shot & 1-shot & 3-shot & 0-shot & 1-shot & 3-shot \\    
			\midrule    
			{Asia}  & 0.4059  & 0.4038  & 0.4302  & 0.5385  & 0.5496  & 0.5918  & 0.4192  & 0.4167  & 0.4596  \\    
			Cancer & 0.5333  & 0.5362  & 0.5594  & 0.5212  & 0.5327  & 0.5698  & 0.4725  & 0.4749  & 0.5151  \\    
			Earthquake & 0.5775  & 0.5714  & 0.6041  & 0.5902  & 0.6058  & 0.6436  & 0.5070  & 0.5024  & 0.5512  \\    
			Sachs & 0.3675  & 0.3728  & 0.3992  & 0.6029  & 0.6241  & 0.6785  & 0.4811  & 0.4910  & 0.5392  \\    
			Survey & 0.5242  & 0.5286  & 0.5481  & 0.4815  & 0.4783  & 0.5256  & 0.3316  & 0.3373  & 0.3701  \\    
			Alarm & 0.1879  & 0.1805  & 0.2231  & 0.5877  & 0.6121  & 0.6998  & 0.3518  & 0.3682  & 0.4254  \\    
			Barley & 0.3705  & 0.3812  & 0.4078  & 0.6502  & 0.6740  & 0.7631  & 0.3387  & 0.3536  & 0.4185  \\    
			Child & 0.2551  & 0.2663  & 0.2926  & 0.5871  & 0.6142  & 0.7024  & 0.2982  & 0.3124  & 0.3721  \\    
			Insurance & 0.2466  & 0.2389  & 0.2811  & 0.5762  & 0.6014  & 0.6880  & 0.2267  & 0.2417  & 0.2928  \\    
			Mildew & 0.2425  & 0.2510  & 0.2768  & 0.6074  & 0.6318  & 0.7175  & 0.3581  & 0.3749  & 0.4337  \\    
			Water & 0.2644  & 0.2588  & 0.2936  & 0.6020  & 0.6132  & 0.6809  & 0.3953  & 0.3921  & 0.4568  \\    Hailfinder & 0.1243  & 0.1309  & 0.1570  & 0.5043  & 0.5211  & 0.6038  & 0.2232  & 0.2344  & 0.2805  \\    Hepar II & 0.1669  & 0.1746  & 0.1994  & 0.6051  & 0.6310  & 0.7186  & 0.3337  & 0.3461  & 0.4052  \\    Win95PTS & 0.1416  & 0.1364  & 0.1771  & 0.6594  & 0.6812  & 0.7557  & 0.3977  & 0.4104  & 0.4689  \\    
			\bottomrule    
	\end{tabular}}%
	\label{tab:accuracy_ICL}%
\end{table}%

\subsection{Detailed Evaluation Results for Parameter-Efficient Fine-Tuning Evaluation}\label{app:fine-tuned}

{Besides in-context prompting, we further study whether parameter-level adaptation can improve LLM performance on causal learning tasks. To this end, we introduce fine-tuned baselines for the causality identification task, which directly evaluate whether an LLM can correctly determine the existence of a direct causal relation between two variables. This task is also the core causal discrimination setting in CausalBN-Bench.}

\subsubsection{Experimental Setting}~{Following the original evaluation protocol, each data sample is formulated as a binary question over variables in a causal network, and the model is required to output only ``Yes'' or ``No.'' We use F1 score and accuracy as the evaluation metrics. To construct the supervision data, we adopt a subgraph-driven strategy based on the ground-truth graph of each causal network. Specifically, for each causal network, we first sample multiple subgraphs by selecting a subset of nodes and keeping the induced edges. We then sample variable pairs within each subgraph and generate causality question-answer pairs with binary labels according to the ground-truth causal structure. This design provides effective supervision while avoiding direct exposure to the complete network structure during training.}

{To further reduce the risk of information leakage, we impose a non-overlap constraint between the variable pairs used for fine-tuning and those used for evaluation. If a variable pair $(X_i, X_j)$ appears in the fine-tuning set of a given causal network, then neither $(X_i, X_j)$ nor its reversed order $(X_j, X_i)$ is allowed to appear in the evaluation set of the same network. Under this setting, the gains from fine-tuning are more likely to come from better task alignment and decision-boundary calibration, rather than from memorizing specific graph edges.}

{We consider two fine-tuning settings. In the \textbf{in-distribution} setting, each causal network is fine-tuned separately using only the training samples generated from its own sampled subgraphs, and the resulting model is evaluated on unseen variable pairs from the same network. This setting simulates a practical scenario in which a user adapts an LLM to a specific domain graph using limited supervised data. In the \textbf{out-of-distribution} setting, we select a subset of causal networks as the training source, fine-tune a unified model on the corresponding subgraph-based supervision data, and then evaluate it on all causal networks, including those not seen during training. This setting is used to test whether fine-tuning can learn transferable causal discrimination patterns across different graph structures.}

{For parameter-efficient adaptation, we adopt low-rank adaptation (LoRA) \cite{hu2022lora} for all fine-tuned LLMs to reduce computational cost and improve reproducibility. Each model is trained for 2 epochs with a learning rate of $1.5\times10^{-4}$, and early stopping is applied based on the validation F1 score or accuracy.}

\subsubsection{Experimental Result Analysis}~{Table~\ref{tab:sft} summarizes the comparison among zero-shot, in-distribution fine-tuning, and out-of-distribution fine-tuning.The results show that parameter-efficient fine-tuning further improves model performance on the causality identification task. In general, the in-distribution setting brings the most stable and largest gains. For example, for LLaMA-7B, the F1 score increases from 0.3483 to 0.3926, and the accuracy increases from 0.4855 to 0.5334. For OPT-66B, the F1 score improves from 0.2934 to 0.3341, and the accuracy improves from 0.3571 to 0.4017. These results indicate that when the fine-tuning and evaluation data come from the same causal-network distribution, parameter-level adaptation can effectively improve the model's causal decision boundary and output calibration.
	The out-of-distribution setting also improves performance for many models, but the gains are usually smaller and less stable. For instance, LLaMA-7B still reaches an F1 score of 0.3678 and an Accuracy of 0.5071, both higher than its zero-shot baseline. However, some smaller models show only limited gains under out-of-distribution fine-tuning, which suggests that cross-network transfer remains challenging. This also indicates that although fine-tuning helps models learn useful task-specific cues, the learned decision patterns remain sensitive to graph distribution shifts and may not generalize as well across different causal structures.}

{Different from the zero-shot and few-shot ICL settings, where model parameters remain frozen, and adaptation is achieved only through prompt examples, the parameter-efficient fine-tuning baselines in Table XXV provide an additional supervised-adaptation perspective by examining whether LoRA-based fine-tuning can improve causal discrimination under in-distribution and out-of-distribution settings.} The results show that when a limited amount of supervised task data is available, the causal discrimination ability of LLMs can be further improved, especially under in-distribution adaptation. Meanwhile, the smaller gains in the out-of-distribution setting highlight that cross-network generalization remains an important challenge for future research.

\begin{table}[htbp]  
	\centering  
	\caption{F1 score and accuracy in fine-tuned models}
	\scalebox{1}[1]{{ 
			\begin{tabular}{l|ccc|ccc}    
				\toprule    \multicolumn{1}{c|}{\multirow{2}[4]{*}{Models}} & \multicolumn{3}{c|}{F1 Score} & \multicolumn{3}{c}{Accuracy} \\\cmidrule{2-7}          & \multicolumn{1}{c}{Zero-shot} & \multicolumn{1}{c}{In-distribution} & \multicolumn{1}{c|}{Out-of-distribution} & \multicolumn{1}{c}{Zero-shot} & \multicolumn{1}{c}{In-distribution} & \multicolumn{1}{c}{Out-of-distribution} \\
				\midrule    
				LLAMA7B & 0.3483 & 0.3926 & 0.3678 & 0.4855 & 0.5334 & 0.5071 \\    
				LLAMA13B & 0.2084 & 0.2469 & 0.2241 & 0.2895 & 0.3352 & 0.3096 \\    
				LLAMA33B & 0.2884 & 0.3298 & 0.3072 & 0.3587 & 0.4059 & 0.3821 \\    
				OPT2D7B & 0.1575 & 0.1971 & 0.1648 & 0.1684 & 0.2146 & 0.1825 \\    
				OPT6D7B & 0.1669 & 0.2135 & 0.1796 & 0.1851 & 0.2359 & 0.2004 \\    
				OPT66B & 0.2934 & 0.3341 & 0.3135 & 0.3571 & 0.4017 & 0.3792 \\    
				Internlm7B & 0.1889 & 0.2416 & 0.1972 & 0.2023 & 0.2587 & 0.2095 \\    
				Internlm20B & 0.2406 & 0.2768 & 0.2551 & 0.2835 & 0.3264 & 0.3012 \\    
				Falcon7B & 0.1633 & 0.2082 & 0.1610 & 0.1830 & 0.2365 & 0.1784 \\    
				Falcon40B & 0.2344 & 0.2719 & 0.2438 & 0.2937& 0.3367 & 0.3061 \\    
				GPT3.5-Turbo & 0.3698 & 0.4012 & 0.3817 & 0.6399 & 0.6664 & 0.6487 \\    
				GPT4  & 0.3608 & 0.3885 & 0.3689 & 0.6237 & 0.6488 & 0.6286 \\    
				GPT4-Turbo & 0.3873 & 0.4119 & 0.3968 & 0.6992 & 0.7206 & 0.7048 \\    
				\bottomrule    
	\end{tabular}}}  
	\label{tab:sft}%
\end{table}%

\clearpage
\section{Detailed Discussions Based on Results and Findings from CausalBN-Bench}\label{app:discussion}
\subsection{Discussion on Evaluation Task Design}\label{app:task_design}
{CausalBN-Bench, including correlation, causal skeleton, causality identification tasks, is designed to evaluate whether LLMs can interpret and reason about causal structures under directed acyclic graph semantics, rather than to replicate or favor any specific causal discovery pipeline. }
{
	\begin{itemize}
		\item From the perspective of data and paradigm compatibility, the evaluation datasets of CausalBN-Bench are primarily drawn from the widely used bnlearn Bayesian network repository, including Asia, Cancer, and Alarm. These datasets provide explicit ground-truth DAG structures and are naturally suitable for defining structured evaluation tasks under Bayesian network semantics. By contrast, LiNGAM typically relies on stronger assumptions such as linearity, non-Gaussianity, and continuous variables, whereas many datasets in CausalBN-Bench do not satisfy these assumptions. GES is centered on score-based global graph search, which is not aligned with the current question-answering-based and task-decomposed evaluation protocol designed for LLMs. Therefore, the aim of CausalBN-Bench is not intended to cover all paradigms of statistical causal discovery; instead, it prioritizes task settings that are more consistent with the current data foundation, graph-structured semantics, and the evaluation format of LLMs.
		\item From the benchmark design perspective, we intentionally organize the evaluation tasks into correlation identification, causal skeleton identification, and causality identification to build a progressive, interpretable, and modular evaluation framework. This hierarchy corresponds to an increasing level of reasoning difficulty: from recognizing statistical association, to identifying structural connection, and finally to inferring causal direction. We believe that this layered design for causality identification is suitable for LLM benchmarking because it allows different levels of causal reasoning ability to be evaluated separately, rather than focusing solely on whether the final full graph is correctly recovered. In addition, this task formulation is more expressible in natural language, better aligned with graph-level ground truth, and more amenable to unified, standardized comparison across different LLMs, prompts, and knowledge conditions. By contrast, functional causal models and score-based approaches usually depend on specific statistical assumptions, functional-form assumptions, or graph search mechanisms, which are not the primary focus of CausalBN-Bench.
		\item From a theoretical positioning perspective: we would like to emphasize that CausalBN-Bench is not intended to reproduce or favor any particular algorithmic pipeline, such as PC algorithm. Rather, CausalBN-Bench aims to evaluate whether LLMs can understand and reason about causal structures under Bayesian networks (DAG semantics) from causal communities. Specifically, the ground-truth answers for evaluation tasks, such as causal skeleton identification and causality identification, are derived from the structural semantics implied by the true directed acyclic graphs, rather than from executing a specific discovery algorithm. In other words, our focus is on whether LLMs can interpret structural relations and causal directions in causal graphs, rather than on whether they can imitate PC's search procedure. Although these evaluation tasks share some conceptual foundations with constraint-based causal discovery methods, such as graph structure, dependency relations, and direction discrimination, this does not mean that the benchmark is tied to or equivalent to the PC algorithm.
	\end{itemize}
}

\subsection{Discussion of the Choice of Traditional Causal Learning Methods}\label{app:tranditional_cl}
{This subsection explains the rationale for selecting the traditional causal discovery baselines in CausalBN-Bench. Our goal is not to exhaust all existing causal discovery algorithms, but to provide a representative, methodologically diverse, and reproducible reference set for comparison with LLM-based approaches.}

{The selected baselines are intended to cover several major paradigms in causal discovery. Specifically, PC is included as a representative constraint-based method; HC and GES are included as representative score-based methods; MMHC is included as a representative hybrid method; DAG-GNN represents deep-learning-based approaches; ENSAOBS represents recent continuous-optimization-based methods; MIGA represents evolutionary search methods; and LiNGAM represents functional-causal-model-based methods. Together, these methods cover a broad range of commonly studied causal discovery paradigms and provide a more comprehensive comparison setting.}

{CausalBN-Bench includes both HC and GES because, although both belong to the score-based family, they differ in an important way: HC performs greedy search directly in the DAG space, whereas GES performs greedy search over equivalence classes. Therefore, including GES and HC  improves the coverage of score-based baselines. CausalBN-Bench also includes LiNGAM to extend the comparison to functional causal models. However, LiNGAM relies on assumptions such as linearity, non-Gaussianity, and continuous-valued variables, which are not fully aligned with many causal networks in CausalBN-Bench, since most of them are discrete Bayesian networks from the bnlearn repository. Therefore, LiNGAM is included mainly to broaden methodological coverage and to illustrate how such an assumption mismatch affects performance under our benchmark setting.}

{Overall, this baseline suite is designed to balance methodological diversity, dataset compatibility, and reproducibility, and to provide a broad set of traditional references for evaluating LLM-based methods under the same datasets and evaluation metrics.}

\subsection{Discussion on Error Analysis}\label{app:error_an}
{
	Although the main manuscript reports the overall quantitative results, the errors made by LLMs also reveal several recurring failure patterns. For clarity, we summarize these patterns from four aspects.
	\begin{itemize}
		\item \textbf{Collider structure failures:} LLMs show clear difficulty in identifying collider structures of the form $A \to C \gets B$. Compared with simple chains or direct links, collider motifs typically achieve lower precision and F1 scores in our fine-grained evaluation. One possible reason is that collider reasoning is less intuitive: two otherwise independent causes can become conditionally associated after conditioning on their common effect. This reversal is easy to miss, especially when the LLM relies more on surface associations than on explicit conditional-independence reasoning. As a result, colliders may be confused with chains or confounders, leading to false-positive edges between variables that share only a common child.
		\item \textbf{Confounder and fork errors with spurious-edge proliferation:} LLM-generated graphs are often denser than the ground-truth graphs. In many cases, the predicted graph contains substantially more edges than the true causal structure. This tendency may preserve some true relations, but it also introduces many false positives, reducing precision and worsening graph-level distance metrics. Our average-degree analysis shows a similar pattern: once the predicted degree becomes much larger than the ground-truth level, the overall structural quality drops noticeably. This suggests that LLMs often adopt an overly inclusive strategy, treating many variables as potentially connected.
		\item \textbf{Semantic misalignment driven by variable names:} The predictions of LLMs appear to be influenced by variable names and prior semantic knowledge. When variable names are informative and align with common-world intuition, LLMs may benefit from these semantic cues. However, when variable names are ambiguous, index-based, or weak in semantic meaning, the LLM is more likely to miss edges or assign incorrect directions. Conversely, when names suggest strong semantic associations, the LLM may overuse such cues and produce predictions that are linguistically plausible but structurally incorrect. This indicates that semantic information is both a strength and a source of bias in LLM-based causal learning.
		\item \textbf{Training data token limitations:} LLMs remain limited in handling large structured numerical inputs such as adjacency matrices or raw data tables. Since they are primarily trained on natural language, their ability to recover global graph structure directly from large matrix-like inputs is still limited. When the graph becomes large, chunk-based prompting is often needed, but this can fragment the reasoning process and weaken the LLM's ability to integrate information across the whole graph. In such settings, performance appears to rely more on semantic priors than on full distributional understanding from structured data.
	\end{itemize}
	Overall, these error patterns suggest that current LLMs have not yet formed a stable causal reasoning mechanism comparable to symbolic or statistical causal discovery methods. Their behavior is influenced by a combination of semantic priors, local pattern matching, and limited structural reasoning. This also explains why LLMs can sometimes give plausible answers on small or semantically meaningful tasks, yet still perform unreliably on larger graphs, collider-heavy structures, or structured-data settings.
}

\subsection{Discussion on LLM Performance in Large-Scale Causal Networks}\label{app:large_network}

{Our experiments show that LLM performance significantly degrades once the number of variables exceeds approximately 50. This motivates us to classify networks into small, medium, large, and very-large categories. We analyze the degradation in large-scale causal networks from three perspectives:}

{
	\begin{itemize}
		\item \textbf{Complexity and sparsity of causal networks:} More complex networks with many hub nodes or a higher average degree often result in overly dense graphs generated by LLMs. Prior work on LLM-based graph generation reports similar issues, where the model produced nearly twice as many edges as the ground truth, resulting in a fully connected graph. In our causal discovery results, predicted graphs have higher in-degrees, out-degrees, and lower sparsity, negatively impacting accuracy. When the structure is dense or hub-centered, the model must consider many relations at once, increasing the chance of incorrect connections, leading to many false-positive edges.
		\item \textbf{Prompt length and information integration limits:} As causal network size grows, the prompt length increases with more variables, relations, and background descriptions. Long and complex prompts degrade LLM reasoning quality. Studies show that performance drops sharply with increasing input length, even within the context window. Open-source models, with smaller context windows, often struggle with long prompts, reducing their ability to retain and integrate information, especially when describing large networks with rich text or large tables.
		\item \textbf{Limitations in handling training data:} Adding numerical or high-dimensional data introduces bottlenecks. In our experiments, many open-source LLMs with fewer than 100 billion parameters struggle to process fine-grained matrices and large training data blocks. Even stronger models require careful prompt design to use data effectively. We observed that adding data often reduces performance; the setting with both variable names and data sometimes yields a lower F1 score than the setting with only variable names. The model struggles to split, read, and combine large numeric tables across long inputs and multiple steps, which worsens performance on large networks.
\end{itemize}}

{These factors become more pronounced as network size increases, explaining why LLM accuracy drops when the graph exceeds 50 nodes. In contrast, experiments on simpler graphs, such as chain-like structures or low-degree graphs, show much better performance. This suggests that breaking the task into smaller parts can help the model. Based on these insights, we propose several strategies to mitigate performance issues on large-scale causal networks:}

{
	\begin{itemize}
		\item \textbf{Decompose the causal learning task:} Instead of one end-to-end causal network structure prediction, break the task into smaller pieces, akin to multi-step reasoning. Partition variables into communities or use local Markov blanket approximations. Infer each subnetwork separately and merge them with conflict resolution. This divide-and-conquer approach, similar to chain-of-thought prompting, improves reasoning on complex tasks. Our ablation studies show higher stability on chain-like structures when solved step by step, suggesting that intermediate supervision can alleviate overload.
		\item \textbf{Multi-path reasoning and self-consistency:} To reduce error propagation in multi-step inferences, we encourage reasoning augmentation. Sampling multiple reasoning chains for the same structure and taking a majority vote, known as self-consistency, improves accuracy. Running several independent subnetwork-inference paths and merging them with consistency checks can remove spurious edges and increase robustness, as multiple plausible reasoning trajectories often converge on the correct structure.
		\item \textbf{Tool-augmented and hybrid LLMs:} To compensate for arithmetic and statistical weaknesses, models can call external tools during reasoning. Recent work shows that LLMs invoking tools like calculators or search engines improve performance. In our setting, the model can call a conditional independence test oracle or a score-based solver. The LLM proposes hypotheses, and tools verify them. This hybrid paradigm combines the scalability of traditional methods with the flexibility of LLMs.
		\item \textbf{Structural priors and sparsification:} To counteract over-dense tendencies, we can bias outputs toward sparsity. For example, limit the number of candidate parents per node, threshold edges by confidence score, or prune low-probability edges as a post-processing step. This mirrors standard structure learning strategies. Lowering the statistical significance threshold can cut false positives and improve precision. A fast constraint-based or score-based method can obtain a skeleton, then the LLM can refine those edges, adding sparsity priors to reduce false positives in large-scale causal networks.
\end{itemize}}

\subsection{Discussion on Practical Implications for Real-World Applications}\label{app:practical}
{{
		A discussion of the practical implications is crucial for linking the evaluation results of CausalBN-Bench to real-world deployment considerations. Therefore, in this subsection, we discuss the practical implications of CausalBN-Bench. From the manuscript, CausalBN-Bench reveals three practical findings that are directly relevant to deployment. First, LLMs perform reasonably well on small problems but fall behind classical causal discovery methods as graph size increases, which indicates that they are better suited for assisting causal learning rather than serving as stand-alone solvers. Second, LLM performance depends strongly on variable semantics, and unclear names or weak metadata can lead to unreliable and even contradictory causal judgments. Third, LLMs tend to output overly dense graphs with many false-positive edges, which increases the risk of misleading downstream decisions. Based on these insights, we summarize their implications for real-world applications as follows:	}
	{
		\begin{itemize} \item \textbf{LLMs should serve as auxiliary tools for causal learning, not as direct solvers for causal discovery.} The results of CausalBN-Bench show that LLMs perform well on small datasets but lag behind traditional methods on larger ones. In practice, LLMs should be used to generate and evaluate causal hypotheses, rather than replace traditional algorithms. They can propose candidate edges with natural language explanations and suggest plausible intervention hypotheses, but final validation of the causal graph should rely on statistically grounded discovery algorithms. This approach is consistent with previous studies, which emphasize that LLMs should not directly determine the existence or direction of causal relationships, but can assist in the search process for causal graphs. In practical applications, LLMs are best used as components that provide semantic priors and search guidance, rather than as stand-alone solvers producing a final causal graph. A recommended workflow is to use LLMs to propose candidate edges, intervention hypotheses, and domain constraints, and then delegate final decisions to traditional causal discovery algorithms or experimental validation. This division of labor reduces the risk of LLMs producing confident but incorrect structures, while still benefiting from their ability to summarize domain knowledge and generate plausible hypotheses. 	
			\item \textbf{Variable semantics and metadata quality.} The results from CausalBN-Bench highlight that clear variable names and descriptions are critical for LLM performance. When variable labels are meaningful, LLMs can leverage background knowledge to propose correct causal links. However, anonymized or poorly described variables lead to misleading inferences. For example, in our experiments, the variable names ``Asia" and ``Visiting to Asia" led to opposite causal assessments. This underscores the importance of high-quality metadata, such as data dictionaries and ontologies. In practice, applications must provide rich semantic context for each variable; otherwise, LLM-generated edges will require careful human review. The strong dependence on variable semantics also suggests that successful deployment requires explicit attention to metadata quality. When variable names and descriptions are clear, LLMs can effectively assist in hypothesis generation. However, when variables are anonymized or ambiguously defined, LLMs become unreliable. Therefore, practitioners should treat metadata construction as part of the causal learning pipeline, maintaining data dictionaries, consistent naming conventions, and concise variable descriptions. In safety-critical domains, adding a human verification step to check LLM-proposed edges against documented semantics and domain rationale is recommended. 	
			\item \textbf{Risk of spurious edges in dense graphs.} CausalBN-Bench shows that LLMs often generate overly dense graphs, leading to many false-positive edges. In fields like healthcare, finance, and public policy, such spurious edges can mislead interventions or resource allocation, with serious consequences. Acting on incorrect causal relationships can waste resources or cause harm. Our metrics, such as SHD and SID, confirm that even when accuracy appears acceptable, inferred graphs can still be far from the ground truth. In practice, this implies a high risk of false causality errors, as a single extra edge can distort the graph. Therefore, LLM-derived structures should be treated as hypotheses and interpreted with caution. This tendency to output overly dense graphs suggests that post-processing and calibration are necessary if LLM outputs are used. In domains like healthcare and finance, even a small number of spurious edges can lead to incorrect intervention recommendations and misallocation of resources. To mitigate this risk, LLM-derived structures should be filtered through conservative criteria before downstream use. This can be implemented by combining LLM proposals with sparsity constraints, statistical significance thresholds, stability checks across resampled datasets, or agreement checks across multiple prompts and models. When intervention decisions are involved, independent validation using established causal identification logic and, when feasible, controlled experiments should be required. 
	\end{itemize} }
	
	{Overall, CausalBN-Bench provides an evidence-based foundation for selecting deployment strategies. It indicates that LLM assistance is most useful for small to moderate networks with high-quality semantics, where LLMs can efficiently generate interpretable candidate hypotheses. For larger networks or environments with weak semantic grounding, traditional causal discovery methods should remain the primary tool, and LLMs should be used only for early-stage exploration under strict validation and governance. These practical implications provide concrete guidance for the safe and effective application of LLMs in real-world causal learning tasks.}

}

\subsection{Discussion on Causal Reasoning and Effect Estimation in LLMs}\label{app:future_reasoning}
{Causal structure discovery and causal effect reasoning are closely related, but they are not equivalent. Recovering a directed acyclic graph provides a compact representation of qualitative causal mechanisms, and many effect-oriented questions rely on such structural information. For example, deciding whether a total causal effect from $X$ to $Y$ can exist, or whether a set of variables can block spurious associations, depends on directed paths, colliders, confounders, and d-separation reasoning. However, accurate causal effect estimation typically requires additional ingredients beyond structure, including identification assumptions, estimators, and often numerical computation. Therefore, strong performance on structural recovery does not necessarily imply strong performance on effect estimation.}

{In this work, the focus is placed on graph-level structural understanding and recovery, and standard structure metrics such as F1, structural Hamming distance, and structural intervention distance are reported. These metrics are partially related to intervention semantics because certain structural errors can lead to large changes in implied interventional behavior. At the same time, they do not directly measure effect magnitude, causal strength, or counterfactual correctness. Structure discovery is therefore regarded as a foundational capability, whereas effect estimation is better treated as a complementary evaluation dimension that should be assessed separately.}

{An important future direction is to extend CausalBN-Bench toward causal effect evaluation in a way that preserves clarity, reproducibility, and interpretability. One natural extension is to introduce graph-based causal effect reasoning tasks that remain grounded in causal network semantics without requiring heavy numerical computation. Such tasks could examine whether a model can use, or infer, structural properties to answer effect-related questions, such as whether a total causal effect from $X$ to $Y$ is possible, whether an $X$-to-$Y$ relation is mediated or confounded, and whether the effect is identifiable through conceptual adjustment-set reasoning without requiring a numeric estimate. Standardized prompt templates and controlled evaluation principles would be important in this setting to reduce confounds, prevent answer leakage, and support diagnosis across different graph sizes and structures.}

{A further extension would be to consider end-to-end causal effect estimation as a separate evaluation track. In such a setting, a model could be given a dataset description or sampled data, together with an intervention query and stated assumptions, and then evaluated on its ability to select an estimation strategy, produce an effect estimate, and provide an interpretation. This direction is naturally different from structure discovery because it also involves estimator selection, numerical computation, and reporting quality, all of which may strongly affect performance. For this reason, it is more appropriate to treat effect estimation as a separate benchmark dimension, while viewing structure discovery as a potential intermediate component that can support downstream decisions such as adjustment selection and effect-query generation.}

\clearpage
\section{Exploratory Experiments for Future Work}\label{app:future}

\subsection{Causal Strength}\label{app:cau_str}
\begin{figure}[htb] 
	\center{\includegraphics[width=0.43\linewidth]  {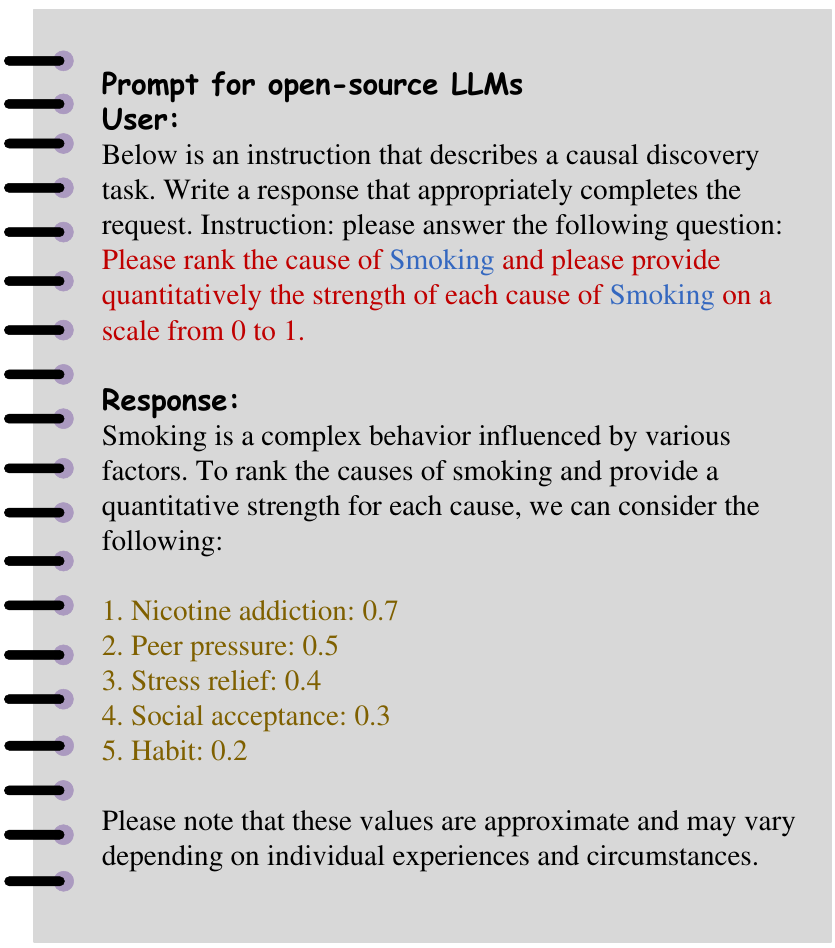}}\caption{Prompt for open-source LLMs in the causal strength evaluation task.} 
	\label{fig:causal_strength} 
\end{figure}

\begin{figure}[htb] 
	\center{\includegraphics[width=1\linewidth]  {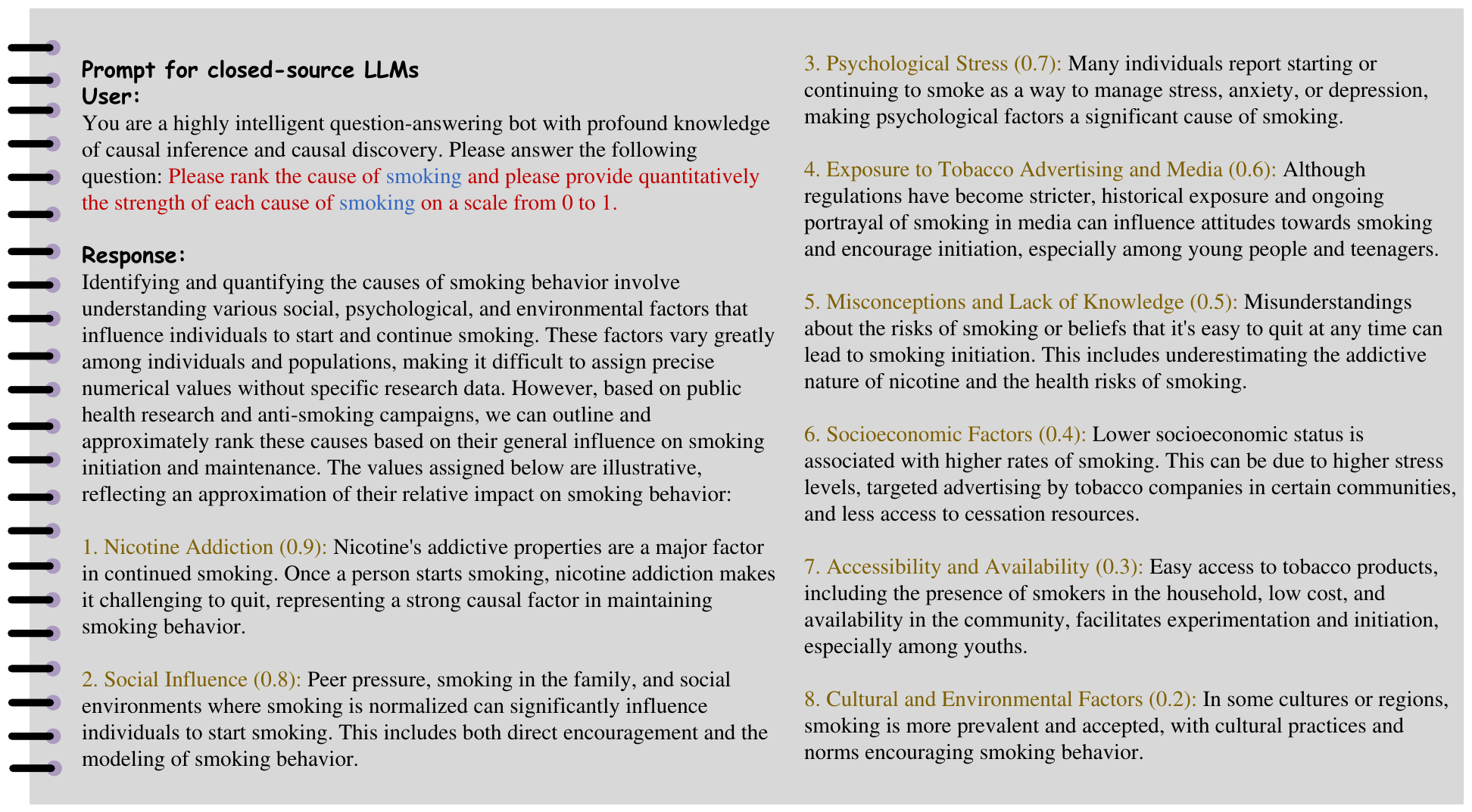}}\caption{Prompt for closed-source LLMs in the causal strength evaluation task.} 
	\label{fig:causal_strength2} 
\end{figure}

The prompts for causal strength are shown in Fig. \ref{fig:causal_strength} and \ref{fig:causal_strength2}, where the red part is the prompt, the blue part is the variable name, and the yellow part represents the causes related to that variable name obtained by LLMs, along with their causal strength ranging between [0,1].

\subsection{Token-Generation Probability Evaluation}\label{app:future_probality}
{To provide a preliminary exploration of continuous causal strength assessment, we additionally conduct a token-generation probability evaluation for causal strength. Inspired by \cite{takayamaintegrating}, this supplementary experiment aims to quantify the model's confidence in a queried causal statement on a continuous scale within $[0,1]$.}

\subsubsection{Evaluation Setup}~{For each queried variable pair $(X_i, X_j)$, we construct a natural-language prompt that asks whether $X_j$ has a causal effect on $X_i$. The prompt format follows the causality-identification setting described in Appendix~\ref{app:future}, except that, instead of using only the final discrete answer, we additionally extract the token-generation probability of the target answer token ``yes".}

{To obtain non-degenerate probability estimates, we adopt a stochastic decoding setting rather than the deterministic setting used in the main experiments. Specifically, we set temperature = 0.7, top-p = 1, and the maximum output length to 512 tokens. For each query, we repeatedly sample the same prompt $M$ times. In the $m$-th run, we record the log-probability assigned by the model to the target token "yes" at the decision position, denoted by $L_{X_i, X_j}^m$, where $m \in \{1,\dots,M\}$. Here, $i$ and $j$ index the queried variable pair $(X_i, X_j)$, and $M$ denotes the total number of repeated samplings.
	Based on the repeated samplings, we compute the averaged ``yes'' probability for the queried pair $(X_i, X_j)$ as
	\begin{equation}
		p_{X_i, X_j}=\frac{1}{M}\sum_{m=1}^{M}\exp \bigl(L_{X_i, X_j}^m\bigr),
	\end{equation}
	where $p_{X_i, X_j} \in [0,1]$ is interpreted as a soft confidence score for the directional causal statement $X_j \rightarrow X_i$. A larger value of $p_{ij}$ indicates that the model assigns higher confidence to the existence of a causal effect from $X_j$ to $X_i$.}

\subsubsection{Evaluation Analysis}
{Based on the above probability-based setting, we conduct experiments on the causality identification task and compare the results with those obtained under the original single-shot discrete evaluation setting. Table~\ref{app:causal_strength_prob} reports the comparison between the original single-shot discrete evaluation and the probability-based evaluation using token-generation probability $p_{X_i, X_j}$. Overall, the probability-based setting achieves comparable or improved performance for most models, especially the stronger LLMs. For example, GPT-4-Turbo improves from 0.3873 to 0.4330 in F1 score and from 0.6992 to 0.7242 in accuracy, while GPT-4 improves from 0.3608 to 0.4079 in F1 score and from 0.6237 to 0.6454 in accuracy. Similar gains can also be observed for some relatively strong open-source models, such as Falcon40B and OPT66B.}

{These results suggest that token-level probability aggregation provides a more fine-grained confidence signal than a single discrete Yes/No answer, which can help improve causal judgment, particularly for relatively capable models. However, the gains are not uniform across all models. For weaker models, the improvements are marginal, and in some cases, performance may slightly decrease, indicating that probability-based evaluation is more effective when the model already has a certain level of causal discrimination.
}
{
	\begin{table}[htbp]
		\centering
		\caption{Comparison of single-shot and multi-shot token-generation probability evaluation on the causality identification task}
		{
			\begin{tabular}{l|cc|cc}
				\toprule
				\multicolumn{1}{c|}{\multirow{1.5}[4]{*}{Models}} & \multicolumn{2}{c|}{F1 score} & \multicolumn{2}{c}{Accuracy} \\
				\cmidrule{2-5}          & \multicolumn{1}{l}{Single-shot} & \multicolumn{1}{l|}{Multi-shot Token-generation Probability} & \multicolumn{1}{l}{Single-shot} & \multicolumn{1}{l}{Multi-shot Token-generation Probability} \\
				\midrule
				LLAMA7B & 0.3483  & 0.3698  & 0.4855  & 0.5098  \\
				LLAMA13B & 0.2084  & 0.2384  & 0.2895  & 0.2787  \\
				LLAMA33B & 0.2884  & 0.2993  & 0.3587  & 0.3742  \\
				OPT2D7B & 0.1575  & 0.1497  & 0.1684  & 0.1428  \\
				OPT6D7B & 0.1669  & 0.1688  & 0.1851  & 0.1855  \\
				OPT66B & 0.2934  & 0.3226  & 0.3571  & 0.3703  \\
				Internlm7B & 0.1889  & 0.1868  & 0.2023  & 0.2057  \\
				Internlm20B & 0.2406  & 0.2561  & 0.2835  & 0.2882  \\
				Falcon7B & 0.1633  & 0.1691  & 0.1830  & 0.1800  \\
				Falcon40B & 0.2344  & 0.2809  & 0.2930  & 0.3253  \\
				GPT3.5-Turbo & 0.3698  & 0.3787  & 0.6399  & 0.6594  \\
				GPT4  & 0.3608  & 0.4079  & 0.6237  & 0.6454  \\
				GPT4-Turbo & 0.3873  & 0.4330  & 0.6992  & 0.7242  \\
				\bottomrule
		\end{tabular}}%
		\label{app:causal_strength_prob}%
	\end{table}%
}
\end{document}